\pdfoutput=1
\documentclass[11pt]{article}

\usepackage[final]{acl}

\usepackage{times}
\usepackage{latexsym}

\usepackage[T1]{fontenc}

\usepackage[utf8]{inputenc}

\usepackage{microtype}

\usepackage{inconsolata}

\usepackage{colortbl}
\usepackage{graphicx}
\usepackage{booktabs}
\usepackage{amsmath} 
\usepackage{amssymb}
\usepackage{mathtools}
\usepackage{amsthm}
\usepackage{multirow}
\usepackage{multicol}
\usepackage{mathrsfs}
\usepackage{diagbox}
\usepackage{dsfont}
\usepackage{hyperref}

\definecolor{darkgreen}{RGB}{60, 160, 60}

\title{Rethinking the Role of Prompting Strategies in LLM Test-Time Scaling:\\ A Perspective of Probability Theory}

\author{
 Yexiang Liu$^{1,2}$,
 Zekun Li$^{3}$,
  Zhi Fang$^{1,2}$,
  Nan Xu$^{1,4}$,
 Ran He$^{1,2}$\thanks{Corresponding author.},
 Tieniu Tan$^{1,2,5}$ \\
  $^1$MAIS, Institute of Automation, Chinese Academy of Sciences \\
  $^2$School of Artificial Intelligence, University of Chinese Academy of Sciences \\
  $^3$University of California, Santa Barbara \\
  $^4$Beijing Wenge Technology Co., Ltd ~
  $^5$Nanjing University\\ 
  \texttt{\{liuyexiang2023, fangzhi2023, xunan2015, ran.he, tieniu.tan\}@ia.ac.cn}\\
  \texttt{zekunli@cs.ucsb.edu} 
}

\begin{document}
\maketitle
\begin{abstract}
Recently, scaling test-time compute on Large Language Models (LLM) has garnered wide attention. However, there has been limited investigation of how various reasoning prompting strategies perform as scaling. In this paper, we focus on a standard and realistic scaling setting: majority voting. We systematically conduct experiments on 6 LLMs $\times$ 8 prompting strategies $\times$ 6 benchmarks. 
Experiment results consistently show that as the sampling time and computational overhead increase, complicated prompting strategies with superior initial performance gradually fall behind simple Chain-of-Thought.
We analyze this phenomenon and provide theoretical proofs. 
Additionally, we propose a probabilistic method to efficiently predict scaling performance and identify the best prompting strategy under large sampling times, eliminating the need for resource-intensive inference processes in practical applications.
Furthermore, we introduce two ways derived from our theoretical analysis to significantly improve the scaling performance. We hope that our research can promote to re-examine the role of complicated prompting, unleash the potential of simple prompting strategies, and provide new insights for enhancing test-time scaling performance. Code is available at \url{https://github.com/MraDonkey/rethinking_prompting}.
\end{abstract}

\section{Introduction}

Over the past few years, how to enhance the reasoning abilities of large language models (LLMs) has been a topic of widespread interest \citep{llama3,anil2023palm2,touvron2023llama2,gpt-4o-mini,gemini1.5}. Researchers have introduced various prompting strategies to improve the reasoning capacity of LLMs, such as Chain of Thought (CoT) \citep{cot} and so on \citep{SBP,AnP,self-refine}. Recently, many studies have shown that scaling LLM test-time compute can also effectively improve reasoning \citep{snell2024scaling, o1, ji2025test, bi2024forest}.

However, how different prompting strategies behave when scaling test-time compute is less explored. In this paper, we focus on a standard and effective scaling setting: majority voting.
We comprehensively evaluate the performance of 8 mainstream prompting strategies under equivalent sampling time or computation overhead. We test 4 open-sourced and 2 closed-sourced LLMs on 6 reasoning benchmarks, finding that simple CoT consistently performs best on all LLMs across benchmarks with given budgets as scaling increases, even if it falls behind at the beginning. 
This indicates that current LLMs can achieve remarkable reasoning capabilities by only relying on simple CoT without other complicated prompting strategies.
It also reminds us to reflect on the necessity of complicated prompting for scaling and fairly compare different strategies under the same budget.

We systematically analyze this phenomenon and provide theoretical and experimental proofs. We conclude that this is caused by two reasons. 
One is that there are more easy questions and fewer hard questions for CoT compared to other strategies. Easy questions are more likely to get right solutions, and the error possibility decreases until 0\% as scaling. In comparison, hard questions are the opposite. 
The other is that CoT is less likely to be affected by wrong answers. Although CoT sometimes has lower pass@1 accuracy, its probability of obtaining the correct answer is more prominent in the result distribution. In contrast, other strategies have higher disturbed peaks in the distribution of incorrect answers. These two reasons enable CoT to improve reasoning performance more rapidly and gradually dominate as scaling.

What's more, we propose a method with the complexity $O(1)$ according to probability theory to quickly predict the scaling performance, which can serve as the test-time scaling law for majority voting.
Experiments show that our method can accurately estimate the scaling performance and select the best strategy with arbitrary sampling time.

Furthermore, we explore two ways to significantly improve scaling performance with our theories. (1) Adaptively scaling according to the question difficulty. (2) Dynamically selecting the optimal prompting strategy. Extensive experiments verify their general effectiveness and superiority, \textit{e.g.}, improving Majority@10 accuracy from 86.0\% to 97.4\% and 15.2\% to 61.0\% for LLaMA-3-8B-Instruct \cite{llama3} on GSM8K \citep{GSM8K} and MATH-500 \cite{math} by combining (1) and (2), respectively.

Our contributions can be summarized as follows:
\begin{itemize}
\vspace{-5pt}
   \item We comprehensively study the test-time scaling performance on 6 LLMs $\times$ 8 prompting strategies $\times$ 6 benchmarks. (Section \ref{scaling system designs})
\vspace{-5pt}
   \item We find that CoT consistently performs best under the equivalent sampling time and computation overhead. (Section \ref{cot dominates})
\vspace{-5pt}
   \item We analyze this phenomenon and provide theoretical and experimental proofs. (Section \ref{section analyze})
\vspace{-18pt}
   \item We propose a method to quickly predict the scaling performance and the best strategy under given sampling times. (Section \ref{predict scaling performance section})
\vspace{-5pt}
   \item Based on the above analysis, we introduce two ways to significantly improve the scaling performance. (Section \ref{improving scaling performance})
\end{itemize}

\newcommand{\M}{\mathcal{M}}
\newcommand{\numberprompt}{n}
\newcommand{\numbersample}{N}
\newcommand{\overhead}{O}
\newcommand{\promptgroup}[1]{\mathcal{P}_{#1}}
\newcommand{\prompt}[1]{\mathbf{P}_{#1}}
\newcommand{\dataset}[1]{\mathfrak{D}_{#1}}
\newcommand{\promptbest}{\mathbf{P}^{*}}
\newcommand{\Fo}[2]{\mathfrak{R}_{#1,#2}}
\newcommand{\h}[1]{\mathbf{h}_{#1}}
\newcommand{\y}[2]{\mathbf{y}_{#1,#2}}
\newcommand{\A}[2]{\mathbf{A}_{#1,#2}}

\section{Scaling System Designs}
\label{scaling system designs}
We focus on a straight and effective setting of test-time scaling, majority voting, \textit{i.e.}, Self-Consistency \citep{self-consistency}, which selects the most consistent answer among several samples.
Our goal is to study what prompting strategy performs best under the equivalent scaling overhead, particularly when largely increasing the scaling extent. 

\subsection{Models}
We conduct experiments on 4 open-sourced LLMs including Qwen2.5-7B-Instruct \citep{qwen2.5}, LLaMA-3-8B-Instruct \citep{llama3}, GLM-4-9B-Chat \citep{glm2024chatglm} and Phi-3.5-mini-Instruct, and 2 closed-sourced LLMs including Gemini-1.5-Flash \citep{gemini1.5} and GPT-4o-mini \citep{gpt-4o-mini}.

\subsection{Prompting Strategies}
We mainly focus on generalizable reasoning prompting strategies, excluding those individually designed for specific tasks or involving fine-tuning, training auxiliary models, or incorporating other models, tools, or human assistance. In this setting, the model's performance is only related to the input prompt, thus making it fairly compare the scaling performance of those prompting strategies. The prompting strategies we test are listed as follows.

\paragraph{Direct Prompting (DiP):} Directly input the question to the model, without any additional instruction or restrictions to the output.
   \vspace{-2pt}
\paragraph{Chain-of-Thought (CoT) \citep{cot, cot-0shot}:} Use the prompt ``Let's think step by step.'' to solve the problem step by step.
   \vspace{-2pt}
\paragraph{Least-to-Most (L2M) \citep{least-to-most}:} Break down the question into progressive sub-questions. Answer the sub-questions and get the final result according to them and their answers.
   \vspace{-2pt}
\paragraph{Tree-of-Thoughts (ToT) \citep{tot}:} Explore multiple reasoning paths to get several solutions, then analyze each solution and decide which one is the most promising.
   \vspace{-2pt}
\paragraph{Self-Refine (S-RF) \citep{self-refine}:} First, answer the question to get an initial answer. Next, evaluate the previous answer and get feedback. Finally, refine the previous answer according to feedback. This will last for several rounds.
   \vspace{-2pt}
\paragraph{Step-Back Prompting (SBP) \citep{SBP}:} First, extract the discipline concepts and principles involved in solving the problem. Then, solve the problem step by step by following the principles.
   \vspace{-2pt}
\paragraph{Analogous Prompting (AnP) \citep{AnP}:} Recall relevant problems as examples. Afterward, solve the analogous problems and proceed to solve the initial problem according to them.
   \vspace{-2pt}
\paragraph{Multi-Agent Debate (MAD) \citep{MAD}:} Set three model instances as different agents to debate for several rounds, and select the most consistent result among them.

\subsection{Benchmarks}
We evaluate across 6 reasoning benchmarks used in the original papers of the above prompting strategies, including GSM8K \citep{GSM8K}, GSM-Hard \citep{GSM-Hard}, MATH-500 \citep{math, lightmanlet}, MMLU-high-school-biology, chemistry and physics \citep{MMLU}.

\begin{figure*}
    \centering  
    \includegraphics[width=0.95\linewidth]{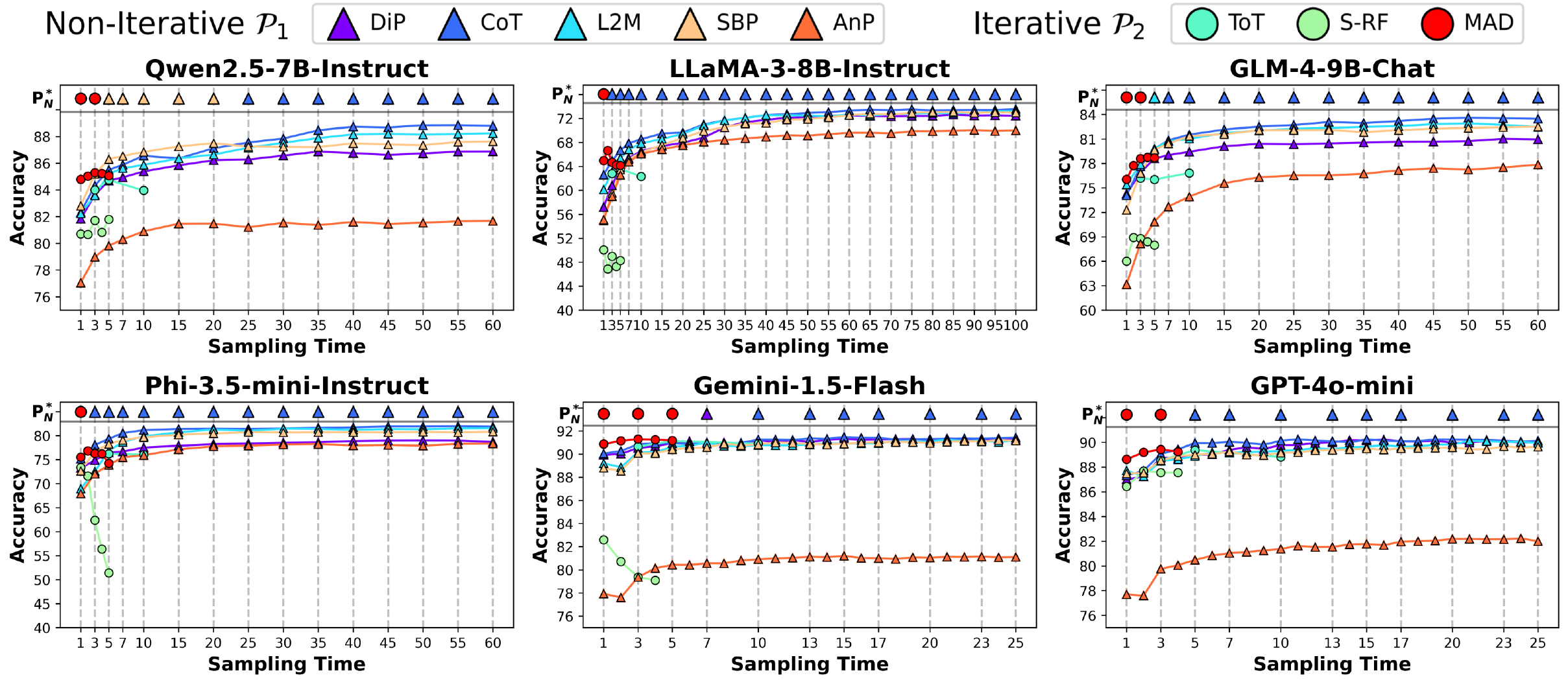}
    \vskip -0.04in
    \caption{Average performances of distinct prompting strategies and the best one $\promptbest_{\numbersample}$ across benchmarks on each LLM under constrained sampling time $\numbersample$. As increasing the sample time $\numbersample$, the accuracy of CoT grows rapidly and it dominates on all models when $\numbersample$ is large enough.}
    \vskip -0.1in
    \label{average_performance_N}
\end{figure*}

\begin{figure*}[!ht]
    \centering
    \includegraphics[width=0.95\linewidth]{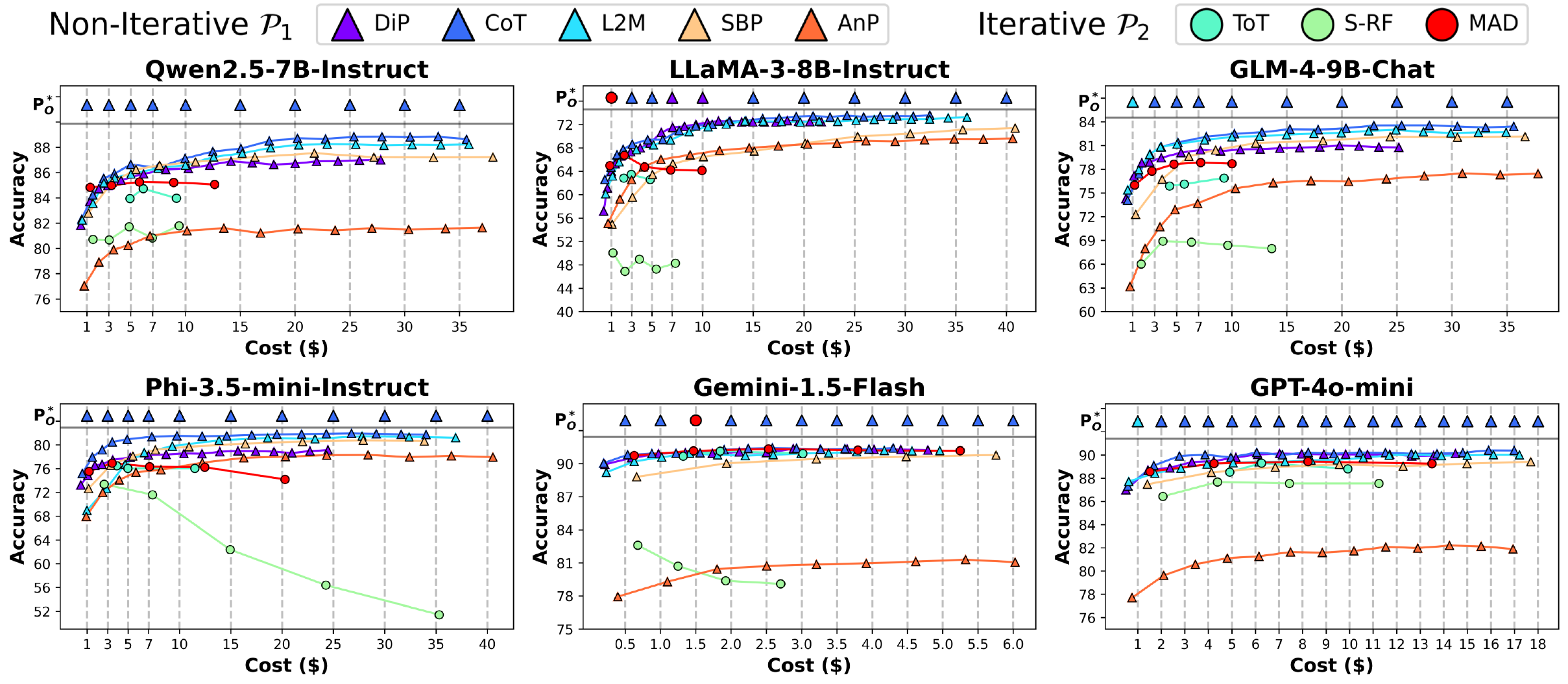}
    \vskip -0.04in
    \small
    \caption{Average performances of distinct prompting strategies and the best one $\promptbest_{\overhead}$ across benchmarks on each LLM under constrained cost $\overhead$. 
    Under the equal cost $\overhead$, CoT performs best most of the time. When $\overhead$ grows larger, CoT gradually becomes the best prompt strategy $\promptbest_\overhead$ on all models.}
    \vskip -0.15in
    \label{average_performance}
\end{figure*}

\subsection{Formal Expression}
We divided the prompting strategies into two groups: iterative methods (S-RF, MAD, and ToT) and the other non-iterative methods.
For S-RF and MAD, we run them $\numbersample$ rounds and get the final result in the $\numbersample$th round. For ToT, we set the model to explore and evaluate $N$ different reasoning paths to get the best one. For others, we parallel sample $\numbersample$ generations and get their most consistent answer with majority voting. For convenience, we refer to all of the above processes as sampling $\numbersample$ times.
Therefore, we can categorize those iterative strategies that require multiple rounds or reasoning paths as $\promptgroup{2}=\{\text{S-RF, MAD, ToT}\}$, and other non-iterative ones as $\promptgroup{1}=\{\text{DiP, CoT, L2M, SBP, AnP}\}$. 

Formally, assuming that we have $\numberprompt$ prompting strategies $\{\prompt{i}\,|\,i=1,2,...,\numberprompt\}$, when using the prompt strategy $\prompt{i}$ to answer a text question $x$ on a model $\M$, we can get the answered result of one sample with an answer extractor $\phi$, which extracts the answer in the output sentence using regular expressions. Then we can formalize the process of getting the final answer when sampling $\numbersample$ times as $\phi [\M(x\,|\,\prompt{i}); \numbersample] = $
\begin{equation}
    \left\{ \begin{matrix}
\operatorname{mode} \{ \phi [\M(x\,|\,\prompt{i})]\}_{1}^{N}, \prompt{i} \in \promptgroup{1}\\
~~~~~ \phi [\M(x\,|\,\prompt{i}; \numbersample)], ~~~~~~ \prompt{i} \in \promptgroup{2}\\
\end{matrix} \right \}
\end{equation}
\vskip -0.06in
With a fixed sampling time $\numbersample$, the best prompting strategy $\promptbest_{\numbersample}$ on the dataset $\dataset{}$ is
\vskip -0.25in
\begin{equation}
    \promptbest_{\numbersample} = \operatorname*{argmax}\limits_{\prompt{i}}\,  \mathbb{E}_{x \in \dataset{}}\, \mathds{1}\{\phi [\M \,(x\,|\,\prompt{i}); \numbersample] = y\},
\end{equation}
where $y$ is the ground truth answer for $x$. However, sampling with distinct $\prompt{i}$ may cause different computation overhead. It would be fairer to compare them with a fixed overhead $\overhead$. To calculate the overhead of using a model $\M$ to answer a question $x$ by sampling $\numbersample$ times with the prompting strategy $\prompt{i}$, we can consider it as a function of $x, \M, \prompt{i}, N$, noted as $\mathcal{C}(x\,|\,\M; \prompt{i};\numbersample)$. Under a fixed overhead $\overhead$, the best prompting strategy $\promptbest_{\overhead}$ on the dataset $\dataset{}$ is
\vspace{-3pt}
    \[
        \promptbest_{\overhead} \!=\! \operatorname*{argmax}\limits_{\prompt{i}} \operatorname*{max}\limits_{N} \mathbb{E}_{x \in \dataset{}} \mathds{1}\{\phi [\M \,(x\,|\,\prompt{i}); \numbersample] = y\},\]
\vspace{-16pt}
\begin{equation}
     s.t. ~\sum_{x \in \dataset{}}\mathcal{C}(x\,|\,\M; \prompt{i};\numbersample) \leq O.
     \vspace{-2pt}
\end{equation}
Given that completion tokens are more computationally expensive than prompt tokens, we define the overhead as the weighted sum of prompt tokens and completion tokens (Cost). For the models Gemini-1.5-Flash and GPT-4o-mini, we utilize their respective pricing metrics.\footnote{The price of Gemini-1.5-Flash: \$0.075/1M prompt tokens, \$0.3/1M completion tokens. The price of GPT-4o-mini: \$0.15/1M prompt tokens, \$0.6/1M completion tokens.} For other open-sourced models, we adopt the pricing of GPT-4o-mini as a proxy.

\section{CoT Dominates as Test-Time Scaling}
\label{cot dominates}
Under each sampling time 
$\numbersample$, we test five times to obtain the average performance of majority voting. 
We evaluate under two kinds of budget constraints: (1) a fixed sampling time budget $\numbersample$, and (2) a fixed inference cost budget $\overhead$.
Figure \ref{average_performance_N} and \ref{average_performance} summarize the average performances across benchmarks of different $\prompt{i}$ under constrained sampling time $\numbersample$ and cost $\overhead$ on each model, and display the best prompting strategy $\promptbest_{\numbersample}$ under different values of $\numbersample$ and $\promptbest_{\overhead}$ under different values of $\overhead$, respectively.\footnote{We don't test the performance with very large $\numbersample$ for $\prompt{i}\in \promptgroup{2}$, as this will lead to extremely long context, large cost and computation time, and marginally increased or even decreased performance, which is no better than Self-Consistency \citep{smitshould}. S-RF performs poorly even with multiple rounds. This is consistent with the results of \cite{huanglarge}, which points out the limitations of S-RF.}
We can see that when scaling test-time compute, CoT performs best among all prompting strategies under a constrained $\numbersample$ and $\overhead$ most of the time. Although some complicated prompting strategies perform best under lower $\numbersample$ and $\overhead$, CoT dominates without exception on all models when largely scaling. 
We theoretically and experimentally analyze this phenomenon, whose reasons come from two aspects. We explain these in detail in Section \ref{section analyze}.

What's more, we find that about 80\% of the results conform to this trend on each model and each benchmark. 
On certain datasets and LLMs, DiP also performs best as largely scaling. This is particularly evident on powerful models, such as Gemini-1.5-Flash and GPT-4o-mini.
More detailed results can be found in Appendix \ref{detailed results}.
These indicate that simple CoT is more efficient and has the potential to surpass other complicated prompting strategies under the same scaling setting. Current LLMs can achieve remarkable reasoning capabilities by only relying on simple prompting strategies. Complicated prompting with superior pass@1 accuracy may not always be better as test-time scaling.

\newcommand{\answer}[1]{{a}_{#1}}
\newcommand{\answersample}[1]{{a}^*_{#1}}
\newcommand{\Answer}{\mathcal{A}}
\newcommand{\prob}[1]{p_{#1}}
\newcommand{\x}{\mathbf{x}}
\newcommand{\occurrence}[1]{\mathbf{X}_{#1}}

\newtheorem{definition}{Definition}
\newtheorem{theorem}{Theorem}
\newtheorem{lemma}{Lemma}

\section{Why CoT Performs worse with Lower $\numbersample$ while better with Larger $\numbersample$?}
\label{section analyze}
Let us consider a specific input question $x$, note the answer space $\Answer=\{\answer{1}, \answer{2}, \dots , \answer{m}\}$ as the set of all probable values of $\phi [\M(x\,|\,\prompt{i})]$ for all $\prompt{i}$, \textit{i.e.}, $\phi [\M(x\,|\,\prompt{i})] \in \Answer \,\,  for \,\, \forall \,\, \prompt{i}$. We omit $\numbersample=1$ in $\phi [\M(x\,|\,\prompt{i})]$ for brevity. $\{\prob{i,1}, \prob{i,2}, \dots , \prob{i,m}\}$ denotes the corresponding probabilities, \textit{i.e.}, $\prob{i,j} = \text{Pr}\,(\phi [\M(x\,|\,\prompt{i})]=\answer{j})$. Note $\answersample{i}$ as the final result of $\prompt{i}$ by scaling sampling $\numbersample$ times, \textit{i.e.}, $\answersample{i} = \phi [\M(x\,|\,\prompt{i}); \numbersample]$.
Then the occurrence number $\occurrence{i} = (\x_{i,1},\dots,\x_{i,m})$ of each probable answer for $\prompt{i}$ follows a multinomial distribution, \textit{i.e.}, $\occurrence{i} \sim \mathit{Mult}(\numbersample, \prob{i,1}, \prob{i,2}, \dots , \prob{i,m})$. The process of getting the final result $\answersample{i}$ of $\prompt{i}$ by sampling $\numbersample$ times can be formalized as:

\begin{equation}
\begin{split}
    \mathcal{J}_i = \{j\,|\,\x_{i,j}=\max\{\occurrence{i}\}\} \\
    k \sim \text{Uniform}(\mathcal{J}_i), \quad \answersample{i} = \answer{k} 
\end{split}
\end{equation}

Next, we will introduce several lemmas and theorems to explain the two reasons why CoT sometimes performs worse with lower $\numbersample$ while better with larger $\numbersample$. In the following proof, we omit the input $x$, assume $\answer{1}$ is the correct answer, and note the probability of getting $\answer{1}$ when sampling $\numbersample$ times with $\prompt{i}$ as $\text{Pr}(\answer{1}|\prompt{i};\numbersample)$, which can be regarded as the expectation of the accuracy $\mathds{1}\{\phi [\M \,(x\,|\,\prompt{i}); \numbersample] = y\}$.  Details about the proof process can be found in Appendix \ref{proofs}.
\begin{definition}
    Note $\prob{max}\!=\!\max\{\prob{i,1}, ..., \prob{i,m}\}$, $\mathcal{S}\!=\!\{\answer{j}\,|\,\prob{i,j}=\prob{max}\}$, we can define the difficulty of the input question $x$ for $\prompt{i}$. If $\answer{1} \in \mathcal{S} ~and~|\mathcal{S}|=1$, we call $x$ an easy question for $\prompt{i}$. If $\answer{1} \in \mathcal{S} ~ and ~ |\mathcal{S}|>1$, we call $x$ a moderate question for $\prompt{i}$. If $\answer{1} \notin \mathcal{S}$, we call $x$ a hard question for $\prompt{i}$. 
\end{definition}
\begin{theorem}
\label{theorem4.2}
If $x$ is an easy question for $\prompt{i}$, $\text{Pr}(\answer{1}|\prompt{i};\numbersample)$ is non-decreasing \textit{w.r.t.} $\numbersample$, \!$\lim \limits_{\numbersample \to +\infty} \!\!\!\! \text{Pr}(\answer{1}|\prompt{i};\numbersample) = 1$.
\end{theorem}
\begin{theorem}
\label{theorem4.3}
    If $x$ is a moderate question for $\prompt{i}$, $\text{Pr}(\answer{1}|\prompt{i};\numbersample)$ is non-decreasing \textit{w.r.t.} $\numbersample$, $\lim \limits_{\numbersample \to +\infty} \text{Pr}(\answer{1}|\prompt{i};\numbersample) = 1/|\mathcal{S}|$.
\end{theorem}
\begin{theorem}
\label{theorem4.4}
        If $x$ is a hard question for $\prompt{i}$, $\text{Pr}(\answer{1}|\prompt{i};\numbersample)$ exhibits a general declining trend \textit{w.r.t.} $\numbersample$, \!$\lim \limits_{\numbersample \to +\infty}\!\!\!\! \text{Pr}(\answer{1}|\prompt{i};\numbersample) = 0$.
\end{theorem}
\begin{lemma}
\label{lemma4.5}
    Consider a specific condition with answer space $|\Answer|=3$. For $\numbersample=3$, $\text{Pr}(\answer{1}|\prompt{i};\numbersample)=3\prob{i,1}^2-2\prob{i,1}^3+2\prob{i,1}\prob{i,2}\prob{i,3}$. For $\numbersample=5$, $\text{Pr}(\answer{1}|\prompt{i};\numbersample)=6\prob{i,1}^5-15\prob{i,1}^4+10\prob{i,1}^3+15\prob{i,1}^2\prob{i,2}\prob{i,3}(\prob{i,2}+\prob{i,3})$.
\end{lemma}
\begin{theorem}
\label{theorem4.6}
    For two prompting strategies $\prompt{i}$ and $\prompt{i'}$, note $\prob{i,q} = \max\{\prob{i,2},\dots,\prob{i,m}\}$, $\prob{i',q'} = \max\{\prob{i',2},\dots,\prob{i',m}\}$, if $\prob{i,1}-\prob{i,q}<\prob{i',1}-\prob{i',q'}$ and $\prob{i,1}+\prob{i,q}-\prob{i,1}^2-\prob{i,q}^2>\prob{i',1}+\prob{i',q'}-\prob{i',1}^2-\prob{i',q'}^2$, there exits a sufficiently large $\numbersample_0$ such that for $\numbersample>\numbersample_0$, $\text{Pr}(\answer{1}|\prompt{i};\numbersample)<\text{Pr}(\answer{1}|\prompt{i'};\numbersample)$.
\end{theorem}

\begin{table*}[ht]
    \centering
    \caption{\textbf{Difficulty proportion of questions and extreme peformance (denote by ``Acc'') for each $\prompt{i}$ and LLM across benchmarks.} CoT has more easy questions and fewer hard questions, and can reach the best extreme performance on all LLMs.}
    \setlength{\extrarowheight}{0.9pt}
    \setlength{\tabcolsep}{4.5pt}
    \small
    \begin{tabular}{l|cccc|cccc|cccc}
    \toprule
     $\prompt{i}$ &\textbf{Easy} & \textbf{Moderate} & \textbf{Hard} & \textbf{Acc} &\textbf{Easy} & \textbf{Moderate} & \textbf{Hard} & \textbf{Acc} &\textbf{Easy} & \textbf{Moderate} & \textbf{Hard} & \textbf{Acc}  \\ \midrule
     \rowcolor{blue!10}
        &\multicolumn{4}{|c}{\textbf{Qwen2.5-7B-Instruct}} &\multicolumn{4}{|c}{\textbf{LLaMA-3-8B-Instruct}} &\multicolumn{4}{|c}{\textbf{GLM-4-9B-Chat}}  \\ 
        DiP & 86.3\% & 0.3\% & 13.4\% & 86.4 & 69.7\% & 1.0\% & 29.3\% & 70.2 & 79.8\% & 0.6\% & 19.6\% & 80.1  \\ 
        CoT & \textbf{88.1\%} & 0.2\% & \textbf{11.6\%} & \textbf{88.2} & \textbf{70.9\%} & 0.9\% & 28.2\% & \textbf{71.3} & \textbf{82.8\%} & 0.8\% & \textbf{16.5\%} & \textbf{83.1}  \\ 
        L2M & 87.4\% & 0.3\% & 12.3\% & 87.6 & 70.3\% & 1.6\% & \textbf{28.1\%} & 71.0 & 81.9\% & 0.4\% & 17.7\% & 82.1  \\ 
        SBP & 87.1\% & 0.1\% & 12.8\% & 87.2 & 67.3\% & 1.3\% & 31.3\% & 68.0 & 81.4\% & 0.9\% & 17.6\% & 81.9  \\ 
        AnP & 81.1\% & 0.5\% & 18.4\% & 81.3 & 67.5\% & 1.4\% & 31.1\% & 68.2 & 76.4\% & 1.2\% & 22.4\% & 77.0  \\ \midrule
        \rowcolor{blue!10}
         &\multicolumn{4}{|c}{\textbf{Phi-3.5-mini-Instruct}}  &\multicolumn{4}{|c}{\textbf{Gemini-1.5-Flash}}  &\multicolumn{4}{|c}{\textbf{GPT-4o-mini}} \\ 
        DiP & 78.4\% & 0.6\% & 21.1\% & 78.6 & 91.0\% & 0.0\% & 9.0\% & 91.0 & 89.7\% & 0.4\% & \textbf{9.9\%} & 89.9 \\ 
        CoT & \textbf{81.2\%} & 0.4\% & \textbf{18.4\%} & \textbf{81.4} & \textbf{91.2\%} & 0.2\% & \textbf{8.6\%} & \textbf{91.3} & \textbf{89.8\%} & 0.3\% & \textbf{9.9\%} & \textbf{90.0}  \\ 
        L2M & 80.2\% & 0.6\% & 19.2\% & 80.5 & 90.9\% & 0.2\% & 89.8\% & 90.9 & \textbf{89.8\%} & 0.3\% & 10.0\% & 89.9 \\
        SBP & 79.0\% & 0.6\% & 20.4\% & 79.3 & 90.6\% & 0.4\% & 9.0\% & 90.8 & 81.4\% & 0.2\% & 10.4\% & 89.5 \\
        AnP & 77.0\% & 1.2\% & 21.8\% & 77.6 & 80.5\% & 0.6\% & 18.8\% & 90.9 & 81.4\% & 1.1\% & 17.5\% & 81.9 \\ 
        \bottomrule
    \end{tabular}
    \vskip -0.1 in
    \label{table difficulty}
\end{table*}

\begin{table}[ht]
    \centering
    \caption{\textbf{Quantity of questions described in Section \ref{cot less affected}.} The value $v_{ii'}$ in the $i$th row and $i'$th column represents the quantity of data that satisfies Theorem \ref{theorem4.6}. Results prove that CoT has greater potential to significantly increase performance as scaling.}
    \vskip 0in
    \small 
    \setlength{\tabcolsep}{4pt}
    \setlength{\extrarowheight}{0.3pt}
    \begin{tabular}{c|ccccc|>{\columncolor{gray!10}}c}
    \toprule
    \rowcolor{blue!10}
    \multicolumn{7}{c}{\textbf{Qwen2.5-7B-Instruct}} \\ \midrule
        \diagbox{$\prompt{i}$ }{$\prompt{i'}$ }  &DiP$\downarrow$ & CoT$\downarrow$ & L2M$\downarrow$ & SBP$\downarrow$ & AnP$\downarrow$ & Sum $\downarrow$\\ \midrule
        DiP $\uparrow$ & - & 447 & 414 & 457 & 393 & 1711 \\ 
        CoT $\uparrow$ & 423 & - & 374 & 416 & 361 & \textbf{1574} \\ 
        L2M $\uparrow$ & 505 & 510 & - & 494 & 403 & 1912 \\ 
        SBP $\uparrow$ & 599 & 601 & 564 & - & 429 & 2193 \\ 
        AnP $\uparrow$ & 800 & 817 & 799 & 776 & - & 3192 \\ \midrule
        \rowcolor{gray!10} Sum $\uparrow$ & 2327 & \textbf{2375} & 2151 & 2143 & 1586 &-\\ \bottomrule
        \toprule

        \rowcolor{blue!10}
        \multicolumn{7}{c}{\textbf{LLaMA-3-8B-Instruct}} \\ \midrule
        \diagbox{$\prompt{i}$ }{$\prompt{i'}$ }  &DiP$\downarrow$ & CoT$\downarrow$ & L2M$\downarrow$ & SBP$\downarrow$ & AnP$\downarrow$ & Sum$\downarrow$ \\ \midrule
        DiP $\uparrow$ & - & 816 & 794 & 459 & 513 & 2582 \\ 
        CoT $\uparrow$ & 620 & - & 646 & 382 & 408 & \textbf{2056} \\ 
        L2M $\uparrow$ & 639 & 677 & - & 432 & 393 & 2141 \\ 
        SBP $\uparrow$ & 1316 & 1433 & 1398 & - & 923 & 5070 \\ 
        AnP $\uparrow$ & 1243 & 1381 & 1380 & 871 & - & 4875 \\ \midrule
       \rowcolor{gray!10} Sum $\uparrow$ & 3818 & \textbf{4307} & 4218 & 2144 & 2237 & -\\ 
    \bottomrule
    \end{tabular}
    \vskip -0.05in
    \label{table4}
\end{table}

\begin{figure}[!t]
    \centering
    \includegraphics[width=0.98\linewidth]{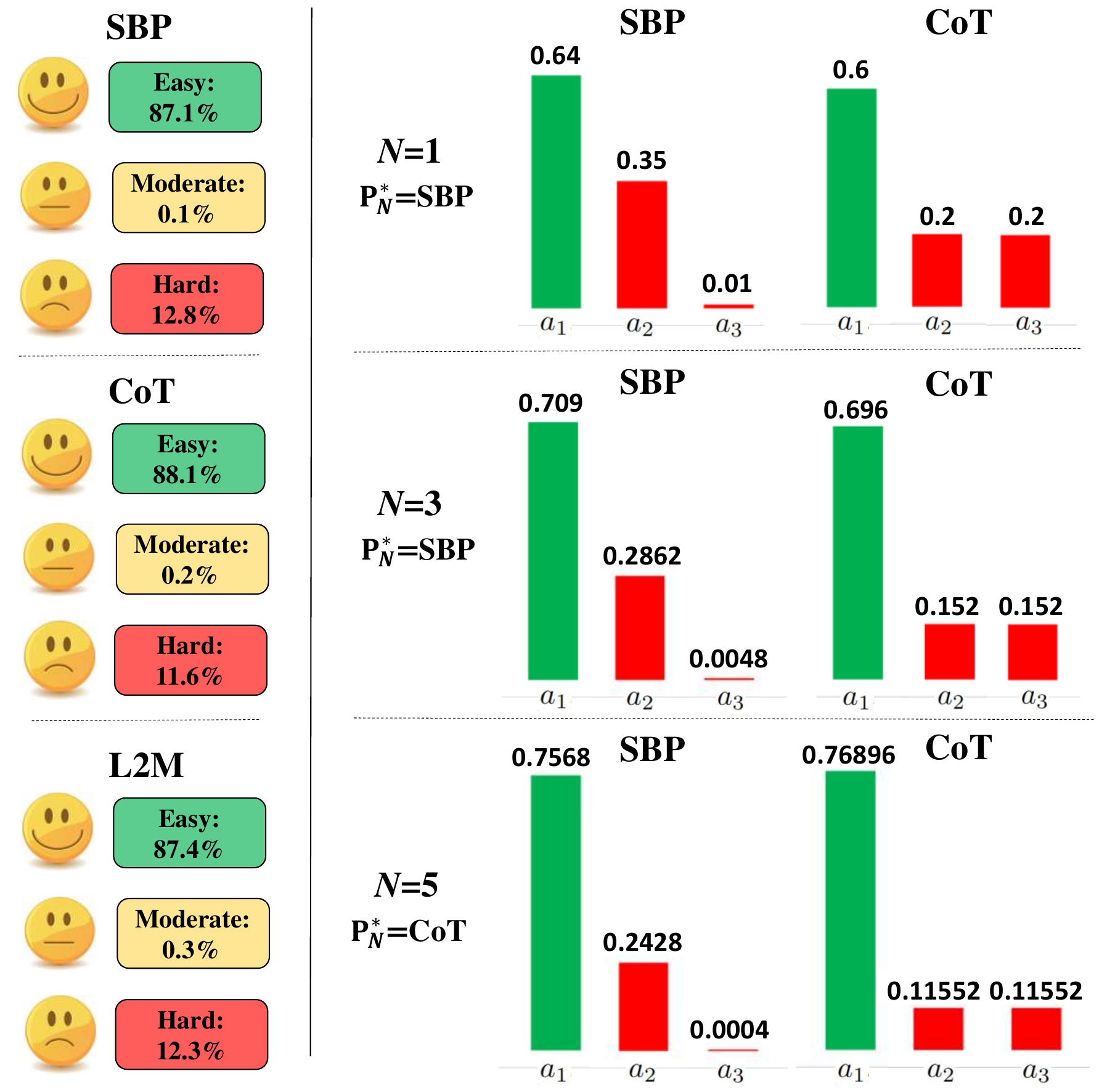}
    \vskip  -0.02in
    \caption{\textbf{Illustration of the two reasons why CoT sometimes performs worse with
lower $\numbersample$ while better with larger $\numbersample$}. \textit{Left:} CoT has more easy questions and fewer hard questions. For example, the probability distribution of L2M is $\{0.4,0.5,0.1,0.0,0.0\}$ (hard question), and $\{0.3,0.2,0.2,0.2,0.1\}$ (easy question) for CoT. Although L2M has higher pass@1 accuracy, its accuracy reduces until 0\% as scaling while CoT increases until 100\%. \textit{Right:} CoT is less likely to be affected by wrong answers due to their relatively uniform distribution. The probability of obtaining the right answer $\answer{1}$ grows more rapidly as increasing $\numbersample$.}
    \label{reason}
    \vskip -0.18in
\end{figure}

\subsection{CoT Has More Easy Questions and Fewer Hard Questions}
We identify two primary reasons why CoT sometimes performs worse with lower sample sizes ($N$) but achieves better performance among these prompting approaches with larger $N$.
The first reason relates to the distribution of question difficulty for CoT. CoT has more easy questions and fewer hard questions. When sampling with lower $\numbersample$, $\prompt{i}$ still has a small probability of obtaining the right answer for hard questions, while the probability diminishes to zero as increasing $\numbersample$.
This is the opposite of easy questions. Figure \ref{Easy vs Hard} shows an example of the accuracy changes on easy/hard questions. The prompting strategy with fewer hard questions and more easy questions will improve performance more rapidly when scaling.
According to Theorem \ref{theorem4.2} to \ref{theorem4.4}, we can calculate the extreme performance of $\prompt{i}$ according to the difficulty proportion of questions, \textit{i.e.}, $\sum \limits_{x \in \dataset{}} \lim \limits_{\numbersample \to +\infty} \!\!\!\! \text{Pr}(\answer{1}|\prompt{i};\numbersample)$.
Table \ref{table difficulty} summarizes the difficulty proportion of the questions and extreme performance for each $\prompt{i}$ on each model. It can be observed that CoT has more easy questions and fewer hard questions, and can reach the best extreme performance on all models, thus making CoT gradually dominate as increasing $N$ even if it has a lower pass@1 accuracy.

\subsection{CoT is Less Likely to be Affected by Wrong Answers}
\label{cot less affected}
The second reason for this phenomenon is that CoT is less likely to be affected by wrong answers. $\text{Pr}(\answer{1}|\prompt{i};\numbersample)$ is not a function of only the probability $\prob{i,1}$ of the right answer $\answer{1}$, but also related to the probability distribution of other wrong answers.
According to Theorem \ref{theorem4.6}, even if $\prob{i,1}>\prob{i',1}$, \textit{i.e.}, $\text{Pr}(\answer{1}|\prompt{i};\numbersample=1)>\text{Pr}(\answer{1}|\prompt{i'};\numbersample=1)$, there still may exist an $\numbersample_0$ that $\text{Pr}(\answer{1}|\prompt{i};\numbersample=\numbersample_0)>\text{Pr}(\answer{1}|\prompt{i'};\numbersample=\numbersample_0)$. 
Considering a question $x$ in GSM8K as an example and $\answer{1}$ is the correct answer, the result probability distribution of $\prompt{i}=\text{SBP}$ is $\{0.64, 0.35, 0.01\}$, and $\{0.6, 0.2, 0.2\}$ for $\prompt{i'}=\text{CoT}$, which satisfies the condition in Theorem \ref{theorem4.6}. According to Lemma \ref{lemma4.5}, $\text{Pr}(\answer{1}|\prompt{i};\numbersample=1)=0.640>\text{Pr}(\answer{1}|\prompt{i'};\numbersample=1)=0.600$, $\text{Pr}(\answer{1}|\prompt{i};\numbersample=3)=0.709>\text{Pr}(\answer{1}|\prompt{i'};\numbersample=3)=0.696$, while $\text{Pr}(\answer{1}|\prompt{i};\numbersample=5)=0.757<\text{Pr}(\answer{1}|\prompt{i'};\numbersample=5)=0.769$. This means, although complicated prompting strategies may have higher pass@1 accuracy, they are easier to be affected by wrong answers. In contrast, simple CoT has a relatively flat distribution on wrong answers, thus making it focus more on the correct answer, which makes it more rapidly improve performance in easy questions and more slowly reduce accuracy in hard questions as increasing $\numbersample$, as shown in Figure \ref{reason}.
We record the quantity of such questions for each two prompting strategies and display the results of Qwen2.5-7B-Instruct and LLaMA-3-8B-Instruct in Table \ref{table4}. If CoT is $\prompt{i'}$, there are the most data satisfying Theorem \ref{theorem4.6}. If CoT is $\prompt{i}$, there are the least such questions. These demonstrate that CoT has greater potential to significantly increase scaling performance compared with other strategies.

To quantify the uniformity of the erroneous answer distributions, we measure the KL divergence between them and the uniform distribution, where lower values indicate closer alignment with greater uniformity. Table \ref{kl_results} shows the average KL divergence on each $\prompt{i}$ and LLM across all benchmarks. We can see that CoT has the most uniform distribution in most cases. While L2M achieves the lowest KL divergence on GLM-4-9B-Chat, CoT exhibits a marginal difference of merely 0.0003, indicating near-identical uniformity.

\begin{table}[!t]
\centering
\caption{Average KL divergence between the erroneous answer distribution and uniform distribution of each $\prompt{i}$ across all tested 6 benchmarks on each LLM.}
\label{kl_results_transposed}
\small
\setlength{\tabcolsep}{2pt}
\setlength{\extrarowheight}{0.5pt}
\resizebox{0.48\textwidth}{!}{ 
\begin{tabular}{l|c|c|c|c|c|c}
\toprule
\rowcolor{blue!10}
 & \textbf{Qwen } & \textbf{LLaMA } & \textbf{GLM } & \textbf{Phi } & \textbf{Gemini } & \textbf{GPT } \\ 
\midrule
DiP  & 0.0774       & 0.0830       & 0.0655       & 0.0770       & 0.0679       & 0.0649 \\
CoT  & \textbf{0.0668} & \textbf{0.0829} & 0.0575       & \textbf{0.0678}       & \textbf{0.0674} & \textbf{0.0624} \\
L2M  & 0.0708       & 0.0857       & \textbf{0.0572} & 0.0724       & 0.0692       & 0.0647 \\
SBP  & 0.0794       & 0.0952       & 0.0603       & 0.0821       & 0.0757       & 0.0647 \\
AnP  & 0.1357       & 0.1154 & 0.1137       & 0.1097       & 0.1990       & 0.1593 \\
\bottomrule
\end{tabular}
}
\label{kl_results}
\vskip -0.12in
\end{table}

As for why CoT has a more uniform probability distribution, we speculate this stems from its simpler approach that avoids imposing specific constraints or guidance on reasoning patterns, compared with other specifically designed prompting approaches. By minimizing explicit guidance, it preserves the model's natural exploration of the solution space. While DiP is also simple, it fails to adequately activate the model's step-by-step reasoning capabilities and free exploration potential. In contrast, more complex prompting strategies introduce explicit guidance mechanisms that constrain reasoning patterns and narrow search directions. This results in probability mass concentrating around specific potential correct solutions at the expense of broader exploratory behavior, ultimately leading to less uniform answer distributions.

\begin{figure*}[ht]
    \vspace{-1mm}
    \centering
    \includegraphics[width=0.98\linewidth]{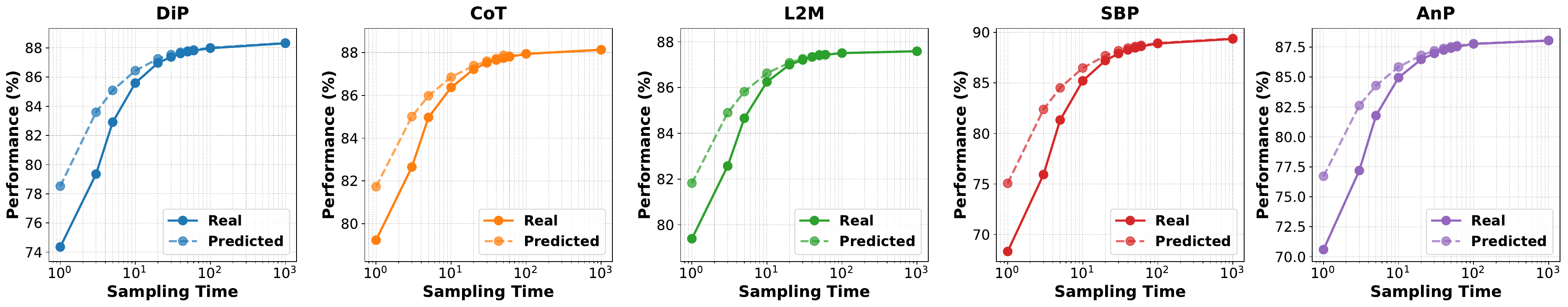}
    \vskip -0.02in
    \caption{Real and predicted performance using our method of different $\prompt{i}$ under various sampling time constraints. Our method can accurately estimate the scaling performance of arbitrary $\prompt{i}$, especially with large $\numbersample$.}
\label{predicted performance}
    \vspace{-2.5mm}
\end{figure*}

\begin{table*}[ht]
    \centering
    \caption{The true best prompting strategy $\promptbest_{\numbersample}$ and the predicted $\promptbest_{\numbersample}$ using our method under various sampling time constraints. Our method can correctly predict the best prompting strategy under any constraints evaluated.}
    \small
    \begin{tabular}{l|cccccccccccc}
    \toprule
    \multirow{2}{*}{\textbf{{$\promptbest_{\numbersample}$}}} & \multicolumn{11}{c}{\textbf{Sampling Time $\numbersample$}} \\ 
    \cmidrule{2-12}
          &\textbf{1} & \textbf{3} & \textbf{5} & \textbf{10} & \textbf{20} & \textbf{30} & \textbf{40} & \textbf{50} & \textbf{60} & \textbf{100} & \textbf{1000}  \\ \midrule
       Oracle & L2M & CoT & CoT & CoT & SBP & SBP & SBP & SBP & SBP & SBP & SBP \\ 
       Predicted \textbf{(Ours)} & L2M & CoT & CoT & CoT & SBP & SBP & SBP & SBP & SBP & SBP & SBP \\
       \hline
       Correctness & \textcolor{darkgreen}{\checkmark} & \textcolor{darkgreen}{\checkmark} & \textcolor{darkgreen}{\checkmark} & \textcolor{darkgreen}{\checkmark} & \textcolor{darkgreen}{\checkmark} & \textcolor{darkgreen}{\checkmark} & \textcolor{darkgreen}{\checkmark} & \textcolor{darkgreen}{\checkmark} & \textcolor{darkgreen}{\checkmark} & \textcolor{darkgreen}{\checkmark} & \textcolor{darkgreen}{\checkmark} \\
        \bottomrule
    \end{tabular}
    \label{predict prompt}

\end{table*}

\section{Predicting Scaling Performance and $\promptbest_{\numbersample}$}
\label{predict scaling performance section}

In practice, evaluating the test-time scaling performance requires significantly intensive resource consumption, especially with very large sampling time $\numbersample$. For pretraining, it is feasible to predict the train-time scaling performance based on the scaling law \citep{scaling_law} through a series of low-cost experiments, while maintaining the model architecture largely unchanged and minimizing the risks associated with large-scale training. 
Similarly, we can also use the sample results of $\prompt{i}$ with fewer $\numbersample$ to approximately get the distribution $\{\prob{i,1}, \prob{i,2}, \dots , \prob{i,m}\}$ to predict the test-time scaling performance with larger $\numbersample$. Directly, one can utilize the multinomial distribution probability calculation formula (Equation \ref{equation_mult} and \ref{Pr equation5}) to calculate $\text{Pr}(\answer{1}|\prompt{i};\numbersample)$ with enumeration or leverage numerical simulation to estimate. However, their computational complexities are both $O(\numbersample)$, and the former needs to traverse all situations and is difficult to operate. Therefore, we propose a method with the computational complexity $O(1)$ to quickly predict the scaling performance of majority voting for arbitrary $\prompt{i}$, which can serve as the test-time scaling law for majority voting. It can also select the best prompting strategy $\promptbest_{\numbersample}$ according to the predicted performances of each $\prompt{i}$.

Here we omit the prompting index $i$ and input question $x$, and assume $\answer{1}$ is the correct answer in the following. 
According to Khinchin's Law of Large Numbers and Lindeberg-Levy Central Limit Theorem, when $N$ is large enough, each occurrence number $\x_j$ can be approximated by a normal distribution. Specifically, for $\x_1$, we have
\vskip -0.06in
\begin{equation}
    \x_1 \sim \mathcal{N}(\numbersample p_1, \numbersample p_1(1-p_1)),
\end{equation}
\vskip -0.04in
\noindent \textit{i.e.}, a normal distribution with mean $\numbersample p_1$ and variance $\numbersample p_1(1-p_1)$. Considering the maximum value among all other $\x_j$ ($j \neq 1$), denoted as $M = \max(\x_2, ..., \x_m)$, when $N$ is large enough, the distribution of $M$ can be approximated by
\vskip -0.2in
\begin{equation}
    M \sim \mathcal{N}(Np_{max}, Np_{max}(1-p_{max})),
\end{equation}
\vskip -0.07in
\noindent where $p_{max}$ is the second highest probability excluding $p_1$. 
We now need to calculate $P(\x_1 > M)$, which can be approximated by comparing two normal distributions. Let $Z = \x_1 - M$, then the distribution of $Z$ is
\vskip -0.25in
\begin{equation}
   \!\!\!\! Z \!\sim\! \mathcal{N}(N(p_1 - p_{max}), N(p_1(1-p_1) + p_{max}(1-p_{max})).
\end{equation}
\vskip -0.02in
\noindent Therefore,
\vskip -0.14in
\begin{equation}
\text{Pr}(\answer{}^*=\answer{1} ) \approx \text{Pr}(Z > 0),
\end{equation}
\vskip -0.05in

\noindent where $\answer{}^*$ is the final sample result. Using properties of the standard normal distribution, we can write
\begin{equation}
\begin{split}
\text{Pr}(Z > 0) &= \text{Pr}\left(\frac{Z - E[Z]}{\sqrt{\text{Var}[Z]}} > \frac{-E[Z]}{\sqrt{\text{Var}[Z]}}\right) \\
&= 1 - \Phi\left(\frac{-E[Z]}{\sqrt{\text{Var}[Z]}}\right),\\
\end{split} 
\end{equation}
\vskip -0.16in
\[
    \!\!\!\!\!\!\!\!\!\!\!\!\!\!\!\!\!\!\!\!\mathbb{E}[Z] = N(p_1 - p_{max}),~~~~~~~~~~~~~~~~~~~\]
\[
    \mathbb{V}[Z] = N(p_1(1-p_1) + p_{max}(1-p_{max})),
\]
\vskip -0.06in


\noindent where $\Phi$ is the standard normal cumulative distribution function. Thus, we can quickly predict the scaling performance and select the best prompting strategy $\promptbest_\numbersample$ with given $\numbersample$ by
\vskip -0.22in
\begin{equation}
\label{predict equation}
   \text{Pr}(\answer{1}|\prompt{i};\numbersample ) \approx 1 - \Phi\left(\frac{-(p_1 - p_{max})}{\sqrt{\frac{p_1(1-p_1) + p_{max}(1-p_{max})}{N}}}\right)\!\!, 
\end{equation}
\begin{equation}
    \text{Accuracy}(\prompt{i},\numbersample) = \mathbb{E}_{x\in \dataset{}} \,\text{Pr}(\answer{1}|\prompt{i},\numbersample),
\end{equation}
\begin{equation}
    \promptbest_\numbersample = \operatorname*{argmax}\limits_{\prompt{i}}\, \text{Accuracy}(\prompt{i}, \numbersample).
\end{equation}


\noindent\textbf{Experiment.~} We verify our method on LLaMA-3-8B-Instruct on GSM8K, only using 40 samples to estimate $\prob{i,j}$. Results are shown in Figure \ref{predicted performance}. We can see that our method can accurately estimate the scaling performance, and the error decreases until 0\% as scaling sampling time. When $\numbersample \geq 10$, the error is already less than 1\%. This makes sense as our method is based on the assumption that $\numbersample$ is large enough. Although the prediction accuracy is not very high when $\numbersample$ is small, the difference in predicted performances between distinct $\prompt{i}$ is similar to that in true performances, so our method can correctly select the best prompting strategy $\promptbest_\numbersample$ with arbitrary $\numbersample$, as shown in Table \ref{predict prompt}.

\section{Improving Scaling Peformance}
\label{improving scaling performance}
According to the analysis in Section \ref{section analyze}, we can further improve the scaling performance in two ways. Extensive experiments confirm their effectiveness, leading to significant improvements. We will further explore them in the future. All following results are conducted on Qwen-2.5-7B-Instruct on GSM8K. Please refer to Appendix \ref{more discussions} for more results on other LLMs and benchmarks.

\subsection{Adaptively Scaling Based on the Difficulty}
\label{adaptive}
\begin{figure*}[!h]
    \centering
    \vskip -0.02in
    \includegraphics[width=0.98\linewidth]{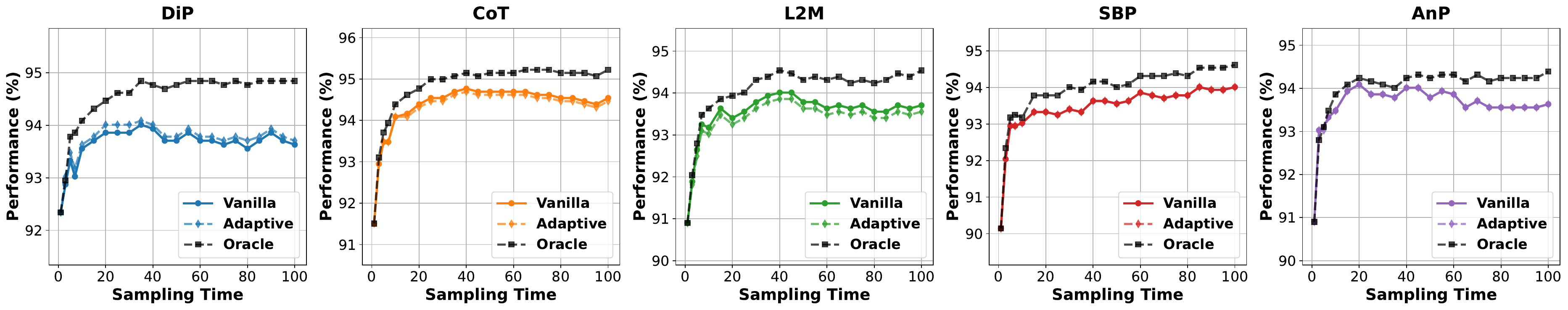}
    \caption{Results of adaptively scaling for each $\prompt{i}\in \promptgroup{1}$ based on oracle and predicted question difficulty.}
    \label{qwen adapative}
    \vskip -0.05in
\end{figure*}

According to Theorem \ref{theorem4.2} to \ref{theorem4.6}, it will lead to decreased performance when scaling sampling time on hard questions. Performances only continuously improve on easy questions. 
\textbf{Therefore, when facing a hard question, we can force LLMs to only answer it once without scaling more. If the question is a moderate or easy question, LLMs scale sampling time as usual.}
We evaluate the performance both when forcing the LLM to determine the question difficulty itself (noted as ``Adaptive'') and providing the difficulty oracle to the LLM as an upper bound reference (noted as ``Oracle''), as shown in Figure \ref{qwen adapative}. 
``Adaptive'' performance is almost equal to the usual scaling performance (noted as ``Vanilla''), which is because the LLM is more inclined to believe a question is easy, especially on more complicated $\prompt{i}$ such as SBP and AnP. Nevertheless, all $\prompt{i}$ can significantly improve their scaling performances with question difficulty oracles, proving the potential of this method. 

\subsection{Dynamically Choosing the Optimal $\prompt{i}$}
\label{section dynamic}
\begin{figure}[!t]
    \centering
    \includegraphics[width=0.99\linewidth]{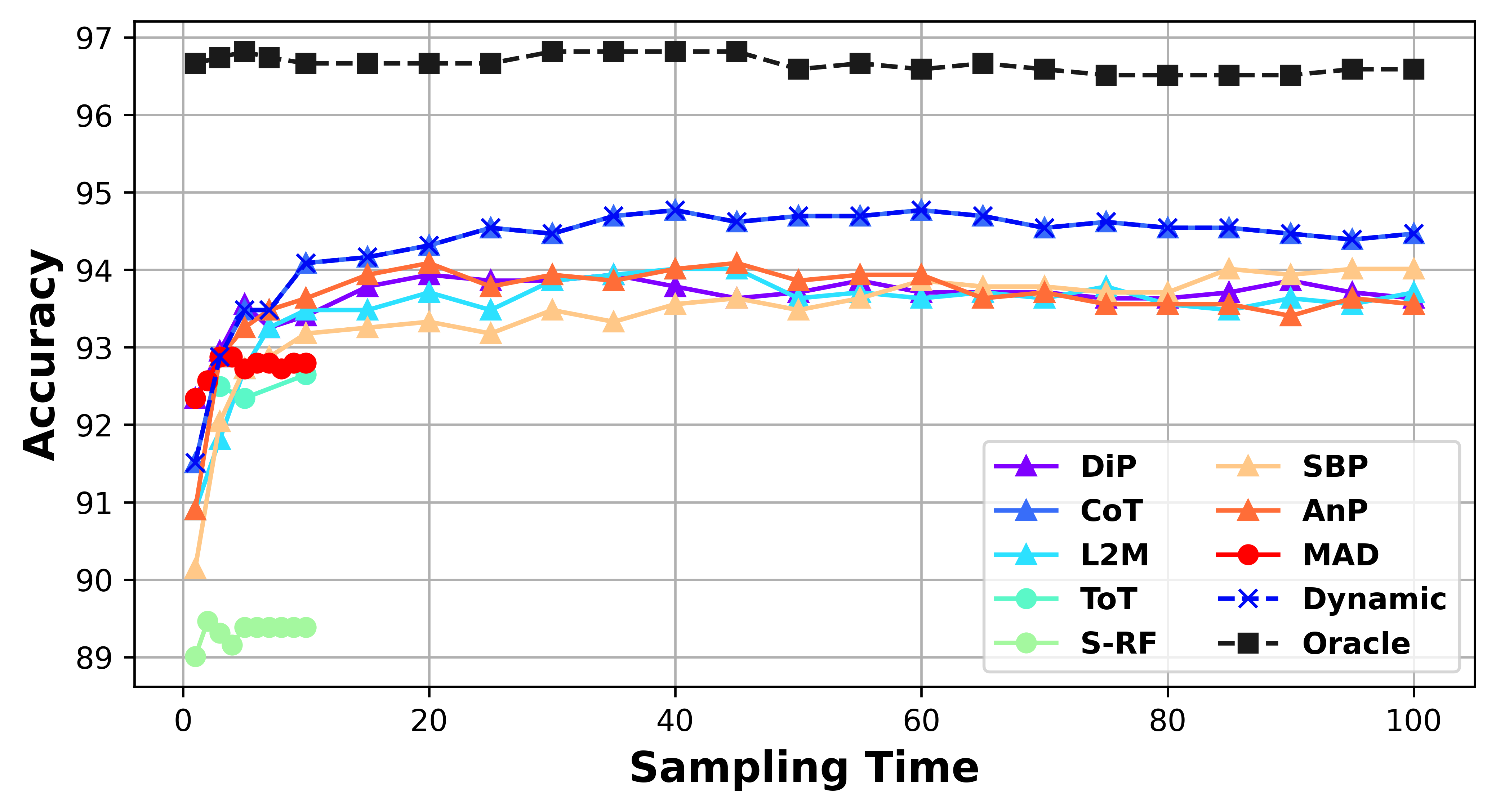}
     \vskip -0.05in
    \caption{Results of dynamically choosing the optimal $\prompt{i}$.$\!\!\!\!\!\!$}
    \label{dynamicw}
    \vskip -0.1in
\end{figure}

For a question $x$, it may be a hard question for a prompting strategy $\prompt{i}$ with higher accuracy, while an easy question for another strategy $\prompt{i'}$ with lower accuracy. 
\textbf{So if we can choose the optimal prompting strategy for each question, it will largely improve the performance.}
We test the scaling performance both when forcing the LLM to choose the most suitable $\prompt{i}$ (noted as ``Dynamic'') and providing the oracles as an upper bound (noted as ``Oracle''), \textit{i.e.}, telling the LLM which $\prompt{i}$ maximizes $\text{Pr}(\answer{1}|\prompt{i};\numbersample)$, as shown in Figure \ref{dynamicw}. ``Dynamic'' performance is almost equal to CoT. This is because Qwen believes CoT is the best $\prompt{i}$ among 8 prompting strategies in 99.7\% of the questions. However, it can achieve significant improvement with oracles. This means that selecting the best $\prompt{i}$ for each question is more effective than majority voting, as ``Oracle'' performance with only $\numbersample=1$ is much higher than $\forall \,\prompt{i}$ with even $\numbersample \rightarrow +\infty$. However, ``Oracle'' performance does not increase with scaling. This is because there are questions that are hard for all $\prompt{i}$. Even if we select the best $\prompt{i}$ on a question, its accuracy still reduces with scaling. So if we can combine the two methods in Section \ref{adaptive} and \ref{section dynamic}, it would lead to much more improvement.

\subsection{Combining Adaptively Scaling and Dynamically Choosing the Optimal $\prompt{i}$}
\begin{figure}[!t]
    \centering
    \includegraphics[width=0.99\linewidth]{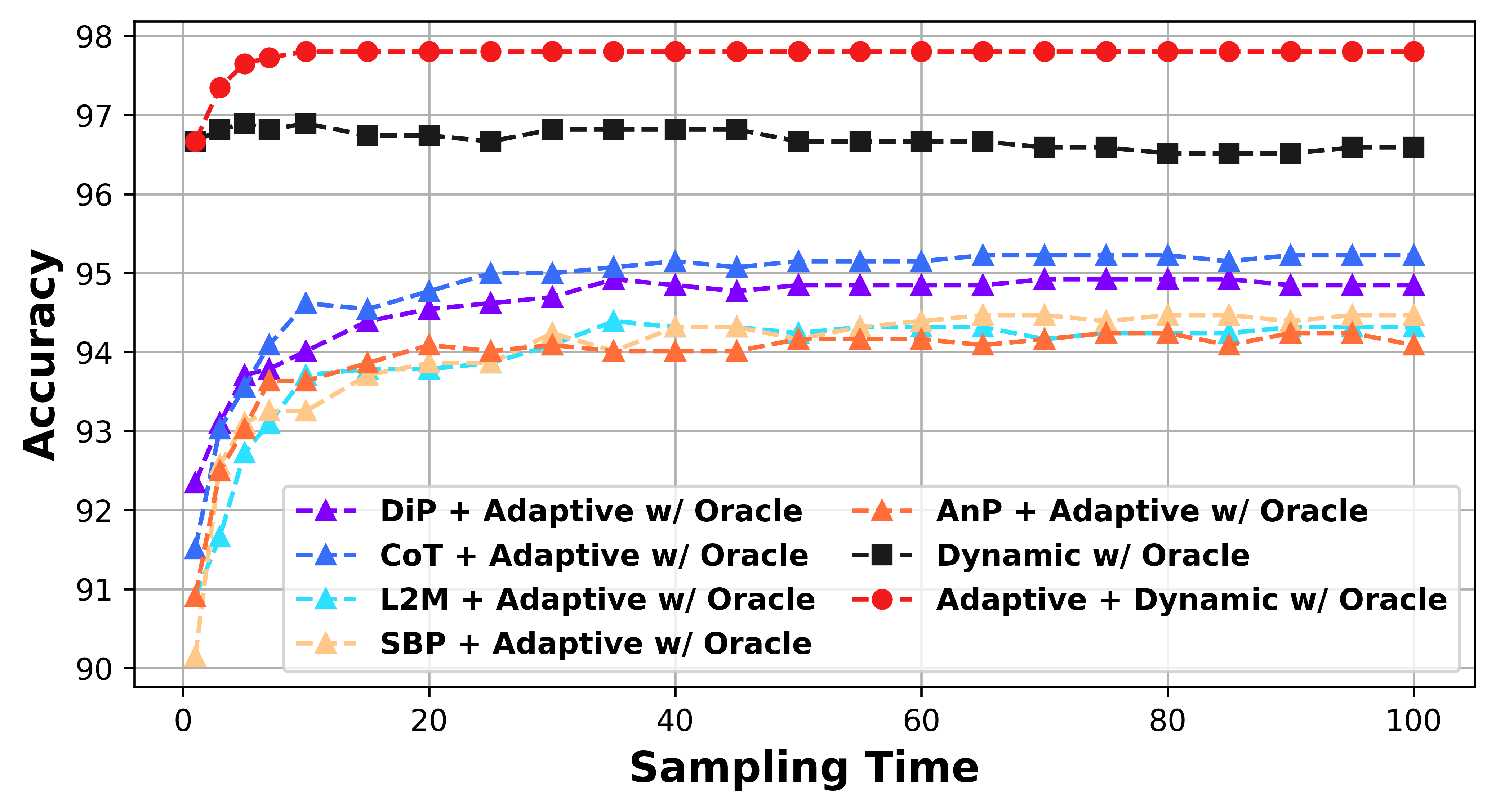}
    \vskip -0.05in
    \caption{Results of combining adaptively scaling and dynamically choosing the optimal $\prompt{i}$ with oracles.}
    \label{combine figure}
    \vskip -0.15in
\end{figure}

Figure \ref{combine figure} reports the peformance upper bounds of each $\prompt{i} \in \promptgroup{1}$ + ``Adaptive'', ``Dynamic'', and combining ``Adaptive'' and ``Dynamic''. Experiment results demonstrate the powerful potential of the combined method. We will explore more feasible methods to reach this upper bound in future work.

\section{Related Work}
\paragraph{Reasoning Prompting Strategies.}
CoT series carefully design exemplars or 0-shot prompts to unleash the potential of step-by-step solving \citep{cot, cot-0shot, zhangautomatic,fu2022complexity}. \cite{least-to-most,dua2022successive,khotdecomposed} break down the question into smaller, more manageable subproblems. \cite{self-refine, kim2023language} force LLMs to self-evaluate and correct. \cite{MAD, liang2023encouraging, chanchateval, DMAD, huang2025towards} utilize multi-agent debate to collaborate reasoning. \cite{AnP, yuthought} guide LLMs to draw experience from analogous problems. \cite{SBP,gao2024efficient} promote LLMs on abstract reasoning.

\paragraph{Scaling Test-Time Compute.} 
Self-Consistency is a simple but effective scaling method \citep{self-consistency}. \cite{li2023making, hosseini2024v} train a verifier to evaluate samples and select the best solution.  Some use iterative refinement \citep{self-refine} or multiple rounds of debate\citep{MAD}. Others leverage the theory of tree search \citep{tot, ding-etal-2024-everything, zhang2024accessing} and graph search \citep{besta2024graph, jin-etal-2024-graph} to expand and aggregate reasoning paths \citep{besta2024topologies}. 

Several studies have shown that scaling test-time compute optimally can be more effective than scaling model parameters \citep{snell2024scaling, o1}. \cite{snell2024scaling} investigates the most effective test-time scaling approach with the basic fixed standard prompting given auxiliary resources, \textit{e.g.}, available datasets for training, specifically trained verifiers, and fine-tuned models. They analyze two kinds of test-time scaling approaches: 1) searching against verifier reward models, and 2) sequential revisions with specifically trained models. 
In contrast, our aim is, completely relying on the LLM itself, which prompting strategy is the most effective with the basic majority voting scaling. We provide new perspectives and theories to 1) understand, 2) predict, and 3) improve the scaling performance of different prompting strategies with the most basic test-time scaling setting.


\section{Conclusion}
We comprehensively study the behavior of various prompting strategies when scaling majority voting. Our experiments on 6 LLMs $\times$ 8 prompting strategies $\times$ 6 benchmarks consistently show that CoT has the potential to perform best as scaling. Theoretical analysis reveals that it benefits from fewer hard questions, more easy questions, and less susceptibility to incorrect answers, enabling more rapid performance gains. 
Additionally, our proposed method for predicting scaling performance offers a practical tool to select the optimal prompting strategy under given sampling time budgets. What's more, we introduce two effective methods to further improve scaling performance.

We also extend experiments on two more challenging reasoning benchmarks, GPQA \citep{GPQA} and AIME \citep{MAA_AIME2024}, further verifying the generality of our findings and methods. Our combined method can achieve a significant boost, elevating accuracy from just over 30\% (majority@100) to 75.7\%. Please refer to Appendix \ref{appendix extended experiments} for more details.

\section*{Limitations}
\label{limitations}
In this paper, we mainly focus on majority voting, which is a simple but effective scaling approach. However, we don't test on other more complex scaling approaches such as Monte Carlo Tree Search. Our finding that CoT dominates as scaling most of the time does not always hold for every LLM on every dataset, \textit{e.g.}, Table \ref{predict prompt}. Nevertheless, 80\% of the results conform to this rule.
In fact, it depends on the composition of the dataset. If we specifically collect hard questions for $\prompt{i}$ as a dataset, it will lead to a continuous decline performance of $\prompt{i}$.
Our experiments and analysis indicate that, even though some $\prompt{i}$ may perform poorly with lower sampling time, they hold the potential to exhibit superior performance than other prompting strategies as test-time scaling.
We propose two superior methods according to rigorous theories, which can significantly improve scaling performance on each model and each benchmark we test, and we are confident in the universality of our methods. However, our experiment results indicate that LLMs alone cannot readily achieve the intended effects, pushing us to explore more practicable and effective methods in our future work.

\section*{Ethical Considerations}
\label{Ethical Considerations}
There are many potential societal consequences of our work, none which we feel must be specifically highlighted here. The sole potential risk we acknowledge is that scaling compute may result in substantial electricity consumption and carbon dioxide emissions.

\section*{Acknowledgment}
We thank Jie Cao and Huaibo Huang for their insightful discussions. The work is supported by National Natural Science Foundation of China (Grant No. 62425606, 32341009, U21B2045) and the Strategic Priority Research Program of Chinese Academy of Sciences (Grant No. XDA0480302).

\bibliography{custom}

\appendix

\clearpage
\section*{Appendix}
In Appendix, we present broader related work, proofs for our theorems, detailed results on each benchmark, more discussions on improving the scaling performance, results of extended experiments, implementation details and prompts. The content structure is outlined as follows:

\begin{itemize}
    \item Appendix~\ref{appendix broader related work} - Broader Related Work
    \item Appendix \ref{proofs} - Proofs
    \item Appendix \ref{detailed results} - Detailed Results
    \item Appendix \ref{more discussions} - More Discussions on Improving the Scaling Performance 
    \begin{itemize}
        \item Appendix \ref{appendix adaptively} - Adaptively Scaling Based on the Difficulty
        \item Appendix \ref{appendix dynamical} - Dynamically Choosing the Optimal $\prompt{i}$
        \item Appendix \ref{appendix combine} - Combining Adaptively Scaling and Dynamically Choosing the Optimal $\prompt{i}$
    \end{itemize}
    \item Appendix \ref{appendix extended experiments} - Extended Experiments
    \item Appendix \ref{implementation details and prompts} - Implementation Details and Prompts
\end{itemize}

\section{Broader Related Work}
\label{appendix broader related work}
\paragraph{Efficient Reasoning.} \cite{ASC, ESC, chen2024more} improve the reasoning efficiency with majority voting by adjusting the sampling time. \cite{ damani2024learning, zhang2024scaling} learn to dynamically allocate resources under limited sampling time budgets. \cite{chen2023frugalgpt, yuelarge, vsakota2024fly} leverage multiple models with different prices to reduce cost while maintaining performance. \cite{yang2025towards, chen2024not, yu2025think} reduce the length of the thinking process to alleviate the overthinking issue, to achieve efficient reasoning.

\paragraph{Role and Mechanism of CoT and Test-Time Scaling.} \cite{jin-etal-2024-impact} studies the impact of reasoning step length of CoT. \cite{wang2023towards} studies what makes CoT
prompting effective, indicating that being relevant to the query and correctly ordering the reasoning steps are more important. \cite{feng2024towards,cuitheoretical} analyze the mechanism of CoT from a theoretical perspective. \cite{sprague2024cot} points out that CoT helps mainly on math and symbolic reasoning by sorting and analyzing a large number of experimental results. \cite{chenunlocking} proposes a framework to quantify the reasoning boundary of CoT. \cite{yang2024context} provides an in-context learning analysis of CoT. \cite{chen2024more} investigates and analyzes the performance changes with more LLM calls. \cite{chen2024simple} proves that the failure probability of test-time scaling decays to zero exponentially or by a power law.

\section{Proofs}
\label{proofs}

\textbf{Theorem} \ref{theorem4.2}. If $x$ is an easy question for $\prompt{i}$, $\text{Pr}(\answer{1}|\prompt{i};\numbersample)$ is non-decreasing \textit{w.r.t.} $\numbersample$, \!$\lim \limits_{\numbersample \to +\infty} \!\!\!\! \text{Pr}(\answer{1}|\prompt{i};\numbersample) = 1$. 

\textbf{Theorem} \ref{theorem4.3}. If $x$ is a moderate question for $\prompt{i}$, $\text{Pr}(\answer{1}|\prompt{i};\numbersample)$ is non-decreasing \textit{w.r.t.} $\numbersample$, $\lim \limits_{\numbersample \to +\infty} \text{Pr}(\answer{1}|\prompt{i};\numbersample) = 1/|\mathcal{S}|$.

\textbf{Theorem} \ref{theorem4.4}.  If $x$ is a hard question for $\prompt{i}$, $\text{Pr}(\answer{1}|\prompt{i};\numbersample)$ exhibits a general declining trend \textit{w.r.t.} $\numbersample$, \!$\lim \limits_{\numbersample \to +\infty}\!\!\!\! \text{Pr}(\answer{1}|\prompt{i};\numbersample) = 0$.

\textit{Proof.} The occurrence number $\occurrence{i} = (\x_{i,1},\dots,\x_{i,m})$ of each probable answer for $\prompt{i}$ follows a multinomial distribution, \textit{i.e.}, $\occurrence{i} \sim \mathit{Mult}(\numbersample, \prob{i,1}, \prob{i,2}, \dots , \prob{i,m})$. When sampling $\numbersample$ times, the specific probability of a certain occurrence number can be calculated with Equation \ref{equation_mult}. For brevity, we omit the input $x$, sampling time $\numbersample$, and the prompting index $i$ in $\x_{i,j}$ and $\prob{i,j}$ in the following equations.
\begin{equation} 
\begin{matrix}
    \!\!\!\text{Pr}\,(\x_{1}=k_{1}, \x_{2}=k_{2},\dots, \x_{m}=k_{m}) \\\\
    = \underbrace{\underbrace{\dfrac{\numbersample!}{k_1!k_2! \cdots k_m!}}_{coefficient}\, \underbrace{\prob{1}^{k_1}\prob{2}^{k_2} \cdots \prob{m}^{k_m}}_{probability\,\,term}}_{a\,\,term\,\,in\,\, \text{Pr}(\answer{1}|\prompt{i};\numbersample)}  \\\\
    s.t. \quad \sum_{j=1}^{m} k_{j}=\numbersample, ~ \sum_{j=1}^{m}\prob{j}=1 
\end{matrix}
\label{equation_mult}
\end{equation}

Assuming the correct answer is $\answer{1}$, $M=\max(k_2,\dots,k_m)$, the probability of obtaining the right answer by sampling $\numbersample$ times with $\prompt{i}$ is
\renewcommand{\arraystretch}{1.5}
\begin{equation}
\begin{matrix}
    \!\!\!\!\!\!\!\!\text{Pr}\,(\answer{1}|\prompt{i})=\text{Pr}\,(\x_{1}>M)+ 
    \sum_{|\mathcal{J}|=1}^{m-1}~~~~~~~~~~~~\vspace{10pt}\\
    \dfrac{\text{Pr}\,(\x_{1}=\x_{j} >\x_{q} ~ {for ~ all} ~ j \in \mathcal{J}, q \notin \{1\} \cup \mathcal{J})}{|\mathcal{J}|+1}, \\
\end{matrix}
\label{Pr equation5}
\end{equation}
where $\mathcal{J}$ is the set of all indexes $j$ ($j \neq 1$) of $\x_{j}$ that satisfies $\x_{j}=\x_{1}$.
\begin{equation}
\begin{matrix}
      \text{Pr}_1 = \text{Pr}\,(\x_{1}>M)= ~~~~~~~~~~~~~~~~~~~~~~~~~~~~~~~~~~~~~~~~~~~~~~~~~~~~~~~\\
     \!\!\!\!\!\!\!\!\sum \limits_{\sum \limits_{j=1}^m k_j=N}\dfrac{\numbersample!}{k_1! \cdots k_m!} \prob{1}^{k_1}\prod_{j=2}^{m}\prob{j}\mathds{1}(k_j < k_1), 
\end{matrix}
\end{equation}
\begin{equation}
\begin{matrix}
\text{Pr}_2\!=\!\text{Pr}(\x_{1}\!=\!\x_{j}\!
    >\!\x_{q} \, {for \, all} \, j \in \mathcal{J}, q \notin \{1\}\! \cup\! \mathcal{J})) ~~~~~~~~~~~~~~~~~~\\  
    \!\!\!\!\!\!\!\!\!\!\!\!\!\!\!=\!\!\!\!\! \sum \limits_{\sum \limits_{j=1}^m k_j=N}\dfrac{\numbersample!}{{k_1!}^{|\mathcal{J}|} \!\!\!\!\!\! \prod \limits_{q \notin \{1\} \cup {J}} \!\!\! \!\!\! k_k!}\prob{1}\prod \limits_{j \in \mathcal{J}}\prob{j} \!\!\!\!\! \prod \limits_{q \notin \{1\} \cup \mathcal{J}}\!\!\!\!\! \prob{k}\mathds{1}(k_k < k_1),~~~~~~~~~~
\end{matrix}
\end{equation}
\renewcommand{\arraystretch}{1}
Pr$_{1}$ represents the probability that $\x_{1}$ is the only maximum number in $\occurrence{i}$. Pr$_{2}$ denotes the probability that there exists more than one maximum number and correctly obtains $\answer{1}$ by randomly choosing from them.

 Here we present a generalized representation. As shown in Equation \ref{Pr equation5}, $\text{Pr}\,(\answer{1}|\prompt{i})$ only includes the cases where $\occurrence{i,1}$ is the maximum value (maybe not the only one). Therefore, for a certain occurrence number $\occurrence{i} = (\x_{1},\dots,\x_{m})$ of each probable answer $\{\answer{1},\dots,\answer{m}\}$, we can reorder $\answer{2},\dots,\answer{m}$ to obtain $\x_{1}=\x_{2}=\dots=\x_{l}>\x_{l+1},\,\x_{l+2},\,\dots,\x_{m}$, where $1\leq l\leq m$. When $l=1$, $\x_{1}$ is the only maximum value. So each term in $\text{Pr}(\answer{1}|\prompt{i};\numbersample)$ can be written as 
\begin{equation}
\frac{1}{l} \cdot \frac{\left(lk+\sum_{j=l+1}^{m}{k_j}\right)!}{(k!)^l \cdot \prod_{j=l+1}^{m}(k_j!)}p_1^kp_2^k\cdots p_l^kp_{l+1}^{k_{l+1}}p_{l+2}^{k_{l+2}}\cdots p_m^{k_m},
\label{term written}
\end{equation}
where $k_1=k_2=\dots =k_l=k>k_j$, $j=l+1,...,m$ and $lk+\sum_{j=l+1}^{m}{k_j}=N$. 

Now we prove Theorem \ref{theorem4.2} and \ref{theorem4.3}. We aim to prove that given the set of answers $\{\answer{1}, \answer{2}, \dots, \answer{m} \}$ with associated probabilities $\{\prob{1}, \prob{2}, \dots , \prob{m}\}$ from $\prompt{i}$, we have $\text{Pr}(\answer{1}|\prompt{i};\numbersample+1) \geq \text{Pr}(\answer{1}|\prompt{i};\numbersample)$ for any $N \in \mathbb{N}^+$. Due to $\sum_{j=1}^{m}{p_j}=1$, the given proposition can be restated as
\begin{equation}
\text{Pr}(\answer{1}|\prompt{i};\numbersample+1) - (\sum_{j=1}^{m}{p_j})\cdot\text{Pr}(\answer{1}|\prompt{i};\numbersample) \geq 0.
\end{equation}

We consider the probability term $p_1^kp_2^k\cdots p_l^kp_{l+1}^{k_{l+1}}p_{l+2}^{k_{l+2}}\cdots p_m^{k_m}$ in each term in $\text{Pr}(\answer{1}|\prompt{i};\numbersample)$, \textit{i.e}, Equation \ref{term written}. When it times $\sum_{j=1}^{m}{p_j}$, there will be three cases.
\vspace{5pt}

Case 1: When it times $\prob{1}$, $\x_1$ is the only maximum value. The probability term becomes

\vspace{5pt}
$p_1^{k+1}p_2^k\cdots p_l^kp_{l+1}^{k_{l+1}}p_{l+2}^{k_{l+2}}\cdots p_m^{k_m}$,
\vspace{5pt}

Case 2: When it times $\prob{s}$, where $s\in\{2,...,l\}$, $\x_s$ become the only maximum value. In this situation, its final result would be an incorrect answer. If $l=1$, case 2 will not exist. The probability term becomes  

\vspace{5pt}
$p_1^kp_2^k\cdots p_{s-1}^kp_s^{k+1}p_{s+1}^k \cdots p_l^kp_{l+1}^{k_{l+1}}p_{l+2}^{k_{l+2}}\cdots p_m^{k_m}$
\vspace{5pt}

Case 3: When it times $\prob{t}$, where $t \in \{l+1,...,m\}$, the value of $\x_t$ changes from $k_t$ to $k_{t}+1$. If $k_t = k-1$, $\x_t$ becomes a new maximum value. If $l=m$, case 3 will not exist. The probability term becomes

\vspace{5pt}
$\!\!\!\!\!\!p_1^kp_2^k\cdots p_l^kp_{l+1}^{k_{l+1}}p_{l+2}^{k_{l+2}}\cdots p_{t-1}^{k_{t-1}}p_t^{k_t+1}p_{t+1}^{k_{t+1}} \cdots p_m^{k_m}$. 
\vspace{5pt}

It can be seen that case 1 and case 3 are also present in $\text{Pr}(\answer{1}|\prompt{i};\numbersample+1)$, whereas case 2 does not.
We begin by considering case 1 and case 2. The terms in $\text{Pr}(\answer{1}|\prompt{i};\numbersample+1)$ corresponding to case 1 $p_1^{k+1}p_2^k\cdots p_l^kp_{l+1}^{k_{l+1}}p_{l+2}^{k_{l+2}}\cdots p_m^{k_m}$ are shown in Equation \ref{equation19}, and no term in $\text{Pr}(\answer{1}|\prompt{i};\numbersample+1)$ involves case 2. The terms in $(\sum_{j=1}^{m}{p_j})\cdot\text{Pr}(\answer{1}|\prompt{i};\numbersample)$ involving case 1 are
shown in Equation \ref{equation20}.
The terms in $(\sum_{j=1}^{m}{p_j})\cdot\text{Pr}(\answer{1}|\prompt{i};\numbersample)$ involving case 2 $p_1^kp_2^k\cdots p_{s-1}^kp_s^{k+1}p_{s+1}^k \cdots p_l^kp_{l+1}^{k_{l+1}}p_{l+2}^{k_{l+2}}\cdots p_m^{k_m}$ are shown in Equation \ref{equation21}. Based on the fact of Equation \ref{equation22}, we can establish the inequality Equation \ref{equation23}, \textit{i.e.}, the terms corresponding to case 1 and case 2 in $\text{Pr}(\answer{1}|\prompt{i};\numbersample+1)$ are greater than or equal to those in $(\sum_{j=1}^{m}{p_j})\cdot\text{Pr}(\answer{1}|\prompt{i};\numbersample)$.
\begin{figure*}[!h]
    \begin{equation}
    \frac{\left(lk+1+\sum_{j=l+1}^{m}k_j\right)!}{(k+1)\cdot(k!)^l\cdot\prod_{j=l+1}^{m}{(k_j!)}}p_1^{k+1}p_2^k\cdots p_l^kp_{l+1}^{k_{l+1}}p_{l+2}^{k_{l+2}}\cdots p_m^{k_m}
    \label{equation19}
\end{equation}
\end{figure*}
\begin{figure*}[!h]
    \centering
    \begin{equation}
    \begin{split}
    &p_1 \cdot \frac{1}{l} \cdot \frac{\left(lk+\sum_{j=l+1}^{m}{k_j}\right)!}{(k!)^l \cdot \prod_{j=l+1}^{m}(k_j!)} p_1^kp_2^k\cdots p_l^kp_{l+1}^{k_{l+1}}p_{l+2}^{k_{l+2}}\cdots p_m^{k_m} \\
    +& \sum_{s=2}^l{p_s\cdot\frac{\left(lk+\sum_{j=l+1}^{m}{k_j}\right)!}{\frac{k+1}{k}\cdot(k!)^l \cdot \prod_{j=l+1}^{m}(k_j!)}}p_1^{k+1}p_2^k\cdots p_{s-1}^kp_s^{k-1}p_{s+1}^k \cdots p_l^kp_{l+1}^{k_{l+1}}p_{l+2}^{k_{l+2}}\cdots p_m^{k_m} \\
    +&\sum_{t=l+1}^m{p_t\cdot\frac{\left(lk+\sum_{j=l+1}^{m}{k_j}\right)!}{\frac{k+1}{k_t}\cdot(k!)^l\cdot\prod_{j=l+1}^m(k_j!)}p_1^{k+1}p_2^k\cdots p_l^kp_{l+1}^{k_{l+1}}p_{l+2}^{k_{l+2}}\cdots p_{t-1}^{k_{t-1}}p_t^{k_t-1}p_{t+1}^{k_{t+1}} \cdots p_m^{k_m}}
    \end{split}
    \label{equation20}
    \end{equation}
\end{figure*}

\begin{figure*}[!h]
    \centering
    \begin{align}
    \sum_{s=2}^l{p_s\cdot\frac{1}{l} \cdot \frac{\left(lk+\sum_{j=l+1}^{m}{k_j}\right)!}{(k!)^l \cdot \prod_{j=l+1}^{m}(k_j!)} p_1^kp_2^k\cdots p_{s-1}^kp_s^kp_{s+1}^k\cdots p_l^kp_{l+1}^{k_{l+1}}p_{l+2}^{k_{l+2}}\cdots p_m^{k_m}}
    \label{equation21}
    \end{align}
\end{figure*}

\begin{figure*}[!h]
    \begin{equation}
\begin{split}
    \frac{\left(lk+1+\sum_{j=l+1}^{m}k_j\right)!}{(k+1)\cdot(k!)^l\cdot\prod_{j=l+1}^{m}{(k_j!)}}&= \frac{\left[(k+1)+(l-1)k+\sum_{t=l+1}^mk_t](lk+\sum_{j=l+1}^{m}k_j\right)!}{(k+1)\cdot(k!)^l\cdot\prod_{j=l+1}^{m}{(k_j!)}}\\&= \frac{1}{l} \cdot \frac{\left(lk+\sum_{j=l+1}^{m}{k_j}\right)!}{(k!)^l \cdot \prod_{j=l+1}^{m}(k_j!)} + (l-1)\cdot\frac{\left(lk+\sum_{j=l+1}^{m}{k_j}\right)!}{\frac{k+1}{k}\cdot(k!)^l \cdot \prod_{j=l+1}^{m}(k_j!)}\\
    &+ \sum_{t=l+1}^m{\frac{\left(lk+\sum_{j=l+1}^{m}{k_j}\right)!}{\frac{k+1}{k_t}\cdot(k!)^l\cdot\prod_{j=l+1}^m(k_j!)}} + (l-1)\cdot\frac{1}{l} \cdot \frac{\left(lk+\sum_{j=l+1}^{m}{k_j}\right)!}{(k!)^l \cdot \prod_{j=l+1}^{m}(k_j!)}
\end{split}
\label{equation22}
\end{equation}
\end{figure*}

\begin{figure*}[!h]
   \begin{equation}
\vspace{10pt}
\begin{split}
    &\frac{\left(lk+1+\sum_{j=l+1}^{m}k_j\right)!}{(k+1)\cdot(k!)^l\cdot\prod_{j=l+1}^{m}{(k_j!)}}p_1^{k+1}p_2^k\cdots p_l^kp_{l+1}^{k_{l+1}}p_{l+2}^{k_{l+2}}\cdots p_m^{k_m} \\ 
    - &p_1 \cdot \frac{1}{l} \cdot \frac{\left(lk+\sum_{j=l+1}^{m}{k_j}\right)!}{(k!)^l \cdot \prod_{j=l+1}^{m}(k_j!)} p_1^kp_2^k\cdots p_l^kp_{l+1}^{k_{l+1}}p_{l+2}^{k_{l+2}}\cdots p_m^{k_m} \\
    -&\sum_{s=2}^l{p_s\cdot\frac{\left(lk+\sum_{j=l+1}^{m}{k_j}\right)!}{\frac{k+1}{k}\cdot(k!)^l \cdot \prod_{j=l+1}^{m}(k_j!)}}p_1^{k+1}p_2^k\cdots p_{s-1}^kp_s^{k-1}p_{s+1}^k \cdots p_l^kp_{l+1}^{k_{l+1}}p_{l+2}^{k_{l+2}}\cdots p_m^{k_m} \\
    -&\sum_{t=l+1}^m{p_t\cdot\frac{\left(lk+\sum_{j=l+1}^{m}{k_j}\right)!}{\frac{k+1}{k_t}\cdot(k!)^l\cdot\prod_{j=l+1}^m(k_j!)}p_1^{k+1}p_2^k\cdots p_l^kp_{l+1}^{k_{l+1}}p_{l+2}^{k_{l+2}}\cdots p_{t-1}^{k_{t-1}}p_t^{k_t-1}p_{t+1}^{k_{t+1}} \cdots p_m^{k_m}}  \\
    -&\sum_{s=2}^l{p_s\cdot\frac{1}{l} \cdot \frac{\left(lk+\sum_{j=l+1}^{m}{k_j}\right)!}{(k!)^l \cdot \prod_{j=l+1}^{m}(k_j!)} p_1^kp_2^k\cdots p_{s-1}^kp_s^kp_{s+1}^k\cdots p_l^kp_{l+1}^{k_{l+1}}p_{l+2}^{k_{l+2}}\cdots p_m^{k_m}} \\
    =&\sum_{s=2}^l\left[\frac{1}{l} \cdot \frac{\left(lk+\sum_{j=l+1}^{m}{k_j}\right)!}{(k!)^l \cdot \prod_{j=l+1}^{m}(k_j!)} p_1^kp_2^k\cdots p_{s-1}^kp_s^kp_{s+1}^k\cdots p_l^kp_{l+1}^{k_{l+1}}p_{l+2}^{k_{l+2}}\cdots p_m^{k_m}\right]\cdot{(p_1-p_s)} \geq 0
\end{split}
\label{equation23}
\end{equation} 
\vspace{10pt}
\end{figure*}
Now we consider case 3 $p_1^kp_2^k\cdots p_l^kp_{l+1}^{k_{l+1}}p_{l+2}^{k_{l+2}}\cdots p_{t-1}^{k_{t-1}}p_t^{k_t+1}p_{t+1}^{k_{t+1}} \cdots p_m^{k_m}$, which can be analyzed by splitting it into two distinct scenarios: $k_t+1<k$ and $k_t+1=k$. 

For scenario $k_t+1<k$, the terms in $\text{Pr}(\answer{1}|\prompt{i};\numbersample+1)$ corresponding to case 3 are shown in Equation \ref{equation24}.
\begin{figure*}[!h]
    \begin{equation}
    \frac{1}{l}\cdot\frac{\left(lk+1+\sum_{j=l+1}^{m}k_j\right)!}{(k_t+1)\cdot(k!)^l\cdot\prod_{j=l+1}^{m}{(k_j!)}}p_1^kp_2^k\cdots p_l^kp_{l+1}^{k_{l+1}}p_{l+2}^{k_{l+2}}\cdots p_{t-1}^{k_{t-1}}p_t^{k_t+1}p_{t+1}^{k_{t+1}} \cdots p_m^{k_m}
    \label{equation24}
\end{equation}
\end{figure*}
The terms in $(\sum_{j=1}^{m}{p_j})\cdot\text{Pr}(\answer{1}|\prompt{i};\numbersample)$ corresponding to case 3 are shown in Equation \ref{equation25}.
\begin{figure*}[!h]
   \begin{equation}
\begin{split}
    & \sum_{s=2}^l{p_s\cdot\frac{1}{l-1}\cdot\frac{\left(lk+\sum_{j=l+1}^{m}{k_j}\right)!}{\frac{k_t+1}{k}\cdot(k!)^l \cdot \prod_{j=l+1}^{m}(k_j!)}}p_1^kp_2^k\cdots p_{s-1}^kp_s^{k-1}p_{s+1}^k \cdots p_l^kp_{l+1}^{k_{l+1}}p_{l+2}^{k_{l+2}}\cdots p_{t-1}^{k_{t-1}}p_t^{k_t+1}p_{t+1}^{k_{t+1}} \cdots p_m^{k_m} \\
    +&\sum_{r=l+1,r\neq t}^m{p_r\cdot\frac{1}{l}\cdot\frac{\left(lk+\sum_{j=l+1}^{m}{k_j}\right)!}{\frac{k_t+1}{k_r}\cdot(k!)^l\cdot\prod_{j=l+1}^m(k_j!)}p_1^kp_2^k\cdots p_l^kp_{l+1}^{k_{l+1}}p_{l+2}^{k_{l+2}}\cdots p_{r-1}^{k_{r-1}}p_r^{k_r-1}p_{r+1}^{k_{r+1}} \cdots p_{t-1}^{k_{t-1}}p_t^{k_t+1}p_{t+1}^{k_{t+1}} \cdots p_m^{k_m}} \\
    +&p_t \cdot \frac{1}{l} \cdot \frac{\left(lk+\sum_{j=l+1}^{m}{k_j}\right)!}{(k!)^l \cdot \prod_{j=l+1}^{m}(k_j!)} p_1^kp_2^k\cdots p_l^kp_{l+1}^{k_{l+1}}p_{l+2}^{k_{l+2}}\cdots p_m^{k_m}
\end{split}
\label{equation25}
\end{equation} 
\end{figure*}
Evidently, we can obtain Equation \ref{equation26} similar to Equation \ref{equation22},
\begin{figure*}[!h]
    \begin{equation}
\begin{split}
    \frac{1}{l}\cdot\frac{\left(lk+1+\sum_{j=l+1}^{m}k_j\right)!}{(k_t+1)\cdot(k!)^l\cdot\prod_{j=l+1}^{m}{(k_j!)}} =& \sum_{s=2}^l{\frac{1}{l-1}\cdot\frac{\left(lk+\sum_{j=l+1}^{m}{k_j}\right)!}{\frac{k_t+1}{k}\cdot(k!)^l \cdot \prod_{j=l+1}^{m}(k_j!)}} \\
    +& \sum_{r=l+1,r\neq t}^m{p_r\cdot\frac{1}{l}\cdot\frac{\left(lk+\sum_{j=l+1}^{m}{k_j}\right)!}{\frac{k_t+1}{k_r}\cdot(k!)^l\cdot\prod_{j=l+1}^m(k_j!)}}\\
    +& \frac{1}{l} \cdot \frac{\left(lk+\sum_{j=l+1}^{m}{k_j}\right)!}{(k!)^l \cdot \prod_{j=l+1}^{m}(k_j!)}
\end{split}
\label{equation26}
\end{equation}
\end{figure*}
and then we can get Equation \ref{equation27}, which proves the terms corresponding to the scenario $k_t+1<k$ in $\text{Pr}(\answer{1}|\prompt{i};\numbersample+1)$ are equal to those in $(\sum_{j=1}^{m}{p_j})\cdot\text{Pr}(\answer{1}|\prompt{i};\numbersample)$.
\begin{figure*}[!h]
   \begin{equation}
\begin{split}
    &\frac{1}{l}\cdot\frac{\left(lk+1+\sum_{j=l+1}^{m}k_j\right)!}{(k_t+1)\cdot(k!)^l\cdot\prod_{j=l+1}^{m}{(k_j!)}}p_1^kp_2^k\cdots p_l^kp_{l+1}^{k_{l+1}}p_{l+2}^{k_{l+2}}\cdots p_{t-1}^{k_{t-1}}p_t^{k_t+1}p_{t+1}^{k_{t+1}} \cdots p_m^{k_m} \\
    -&\sum_{s=2}^l{p_s\cdot\frac{1}{l-1}\cdot\frac{\left(lk+\sum_{j=l+1}^{m}{k_j}\right)!}{\frac{k_t+1}{k}\cdot(k!)^l \cdot \prod_{j=l+1}^{m}(k_j!)}}p_1^kp_2^k\cdots p_{s-1}^kp_s^{k-1}p_{s+1}^k \cdots p_l^kp_{l+1}^{k_{l+1}}p_{l+2}^{k_{l+2}}\cdots p_{t-1}^{k_{t-1}}p_t^{k_t+1}p_{t+1}^{k_{t+1}} \cdots p_m^{k_m} \\
    -&\sum_{r=l+1,r\neq t}^m\!\!\!\!\!\!{p_r\cdot\frac{1}{l}\cdot\frac{\left(lk+\sum_{j=l+1}^{m}{k_j}\right)!}{\frac{k_t+1}{k_r}\cdot(k!)^l\cdot\prod_{j=l+1}^m(k_j!)}p_1^kp_2^k\cdots p_l^kp_{l+1}^{k_{l+1}}p_{l+2}^{k_{l+2}}\cdots p_{r-1}^{k_{r-1}}p_r^{k_r-1}p_{r+1}^{k_{r+1}} \cdots p_{t-1}^{k_{t-1}}p_t^{k_t+1}p_{t+1}^{k_{t+1}} \cdots p_m^{k_m}} \\
    -&p_t \cdot \frac{1}{l} \cdot \frac{\left(lk+\sum_{j=l+1}^{m}{k_j}\right)!}{(k!)^l \cdot \prod_{j=l+1}^{m}(k_j!)} p_1^kp_2^k\cdots p_l^kp_{l+1}^{k_{l+1}}p_{l+2}^{k_{l+2}}\cdots p_m^{k_m}
    = 0
\end{split}
\label{equation27}
\end{equation} 
\end{figure*}

For scenario $k_t+1=k$, in a similar manner, according to the Equation \ref{equation28},
\begin{figure*}[!h]
   \begin{equation}
\begin{split}
    \frac{1}{l+1}\cdot\frac{\left(lk+1+\sum_{j=l+1}^{m}k_j\right)!}{(k_t+1)\cdot(k!)^l\cdot\prod_{j=l+1}^{m}{(k_j!)}} =& \sum_{s=2}^l{\frac{1}{l}\cdot\frac{\left(lk+\sum_{j=l+1}^{m}{k_j}\right)!}{(k!)^l \cdot \prod_{j=l+1}^{m}(k_j!)}} \\
    +& \sum_{r=l+1,r\neq t}^m{p_r\cdot\frac{1}{l+1}\cdot\frac{\left(lk+\sum_{j=l+1}^{m}{k_j}\right)!}{\frac{k_t+1}{k_r}\cdot(k!)^l\cdot\prod_{j=l+1}^m(k_j!)}}\\
    +& \frac{1}{l} \cdot \frac{\left(lk+\sum_{j=l+1}^{m}{k_j}\right)!}{(k!)^l \cdot \prod_{j=l+1}^{m}(k_j!)}
\end{split}
\label{equation28}
\end{equation} 
\end{figure*}
we obtain the same result as the above scenario, \textit{i.e.}, the terms corresponding to the scenario $k_t+1=k$ in $\text{Pr}(\answer{1}|\prompt{i};\numbersample+1)$ are equal to those in $(\sum_{j=1}^{m}{p_j})\cdot\text{Pr}(\answer{1}|\prompt{i};\numbersample)$. 

Thus far, let us revisit the proof steps. In the first step, we expand the expression $(\sum_{j=1}^{m}{p_j})\cdot\text{Pr}(\answer{1}|\prompt{i};\numbersample)$ and divide it into three cases, where case 1 and case 3 are present in $\text{Pr}(\answer{1}|\prompt{i};\numbersample+1)$ whereas case 2 does not. It has been proven that the coefficients of the terms in $\text{Pr}(\answer{1}|\prompt{i};\numbersample+1)$, where case 3 appears as the probability term part, are identical to those in $(\sum_{j=1}^{m}{p_j})\cdot\text{Pr}(\answer{1}|\prompt{i};\numbersample)$. Consequently, these terms cancel out in the expression $\text{Pr}(\answer{1}|\prompt{i};\numbersample+1) - (\sum_{j=1}^{m}{p_j})\cdot\text{Pr}(\answer{1}|\prompt{i};\numbersample)$. However, case 2 is not present in $\text{Pr}(\answer{1}|\prompt{i};\numbersample+1)$, which implies that the terms in $(\sum_{j=1}^{m}{p_j})\cdot\text{Pr}(\answer{1}|\prompt{i};\numbersample)$, where case 2 appears as the probability term part, cannot be combined with any terms in $\text{Pr}(\answer{1}|\prompt{i};\numbersample+1)$ by extracting the exponent and performing subtraction on the coefficients like the terms containing case 3. It is fortunate that the terms in $\text{Pr}(\answer{1}|\prompt{i};\numbersample+1)$, where case 1 appears as the probability term part, cancel out with the corresponding terms in $(\sum_{j=1}^{m}{p_j})\cdot\text{Pr}(\answer{1}|\prompt{i};\numbersample)$ which share the same probability terms and the remaining terms have coefficients identical to those of the terms in $(\sum_{j=1}^{m}{p_j})\cdot\text{Pr}(\answer{1}|\prompt{i};\numbersample)$, where case 2 appears as the probability terms. Therefore, these terms can be combined by factoring out the shared coefficients and partial probability terms. The remaining part after factoring out the common factor is $\sum_{s=2}^l(p_1 - p_s)$. It is undeniable that $p_1 \geq p_s$ when $x$ is an easy question or moderate question for $\prompt{i}$, therefore $\text{Pr}(\answer{1}|\prompt{i};\numbersample+1) - (\sum_{j=1}^{m}{p_j})\cdot\text{Pr}(\answer{1}|\prompt{i};\numbersample) \geq 0$. Namely, $\text{Pr}(\answer{1}|\prompt{i};\numbersample)$ is strictly non-decreasing \textit{w.r.t.} $\numbersample$ if $x$ is an easy or moderate question.

For sufficiently large $\numbersample$, the strong law of large numbers implies that $\text{Pr}(\lim \limits_{\numbersample \rightarrow +\infty}{\x_{j}}/{\numbersample}=\prob{j})=1$. When $x$ is an easy question, $\prob{1}>\prob{j}, \x_1/\numbersample > \x_j/\numbersample$ for $j=2,3,...,m$. As $\numbersample$ is sufficiently large, it is sure that $\x_1$ is the only maximum value, making the final result must be the correct answer $\answer{1}$. Therefore, if $x$ is an easy question, $\numbersample$, \!$\lim \limits_{\numbersample \to +\infty} \!\!\!\! \text{Pr}(\answer{1}|\prompt{i};\numbersample) = 1$. If $x$ is a moderate question, there are $|\mathcal{S}|$ equivalent answers in the probability sense, whose probabilities are all the maximum value. Therefore, $\lim \limits_{\numbersample \to +\infty} \!\!\!\! \text{Pr}(\answer{1}|\prompt{i};\numbersample) = 1/|\mathcal{S}|$. Similarly, if $x$ is a hard question, the maximum probability is not $\prob{1}$, the final result must be a wrong answer, so $\lim \limits_{\numbersample \to +\infty} \!\!\!\! \text{Pr}(\answer{1}|\prompt{i};\numbersample) = 0$. Theorem \ref{theorem4.2} to \ref{theorem4.4} is proved.

\textbf{Lemma} \ref{lemma4.5}. Consider a specific condition with answer space $|\Answer|=3$. For $\numbersample=3$, $\text{Pr}(\answer{1}|\prompt{i};\numbersample)=3\prob{i,1}^2-2\prob{i,1}^3+2\prob{i,1}\prob{i,2}\prob{i,3}$. For $\numbersample=5$, $\text{Pr}(\answer{1}|\prompt{i};\numbersample)=6\prob{i,1}^5-15\prob{i,1}^4+10\prob{i,1}^3+15\prob{i,1}^2\prob{i,2}\prob{i,3}(\prob{i,2}+\prob{i,3})$.

\textit{Proof.} For $\numbersample=3$, we can calculate $\text{Pr}(\answer{1}|\prompt{i};\numbersample)$ with Equation \ref{Pr equation5} as follows:
\begin{equation}
\begin{split}
\text{Pr}(\answer{1}|\prompt{i};\numbersample=3) =&  \binom{3}{3}\prob{i,1}^3 + \binom{3}{2}\prob{i,1}^2(1-\prob{i,1}) \\
&+ A(3,1)\,\prob{i,1}\prob{i,2}\prob{i,3}\\
=\,&3\prob{i,1}^2-2\prob{i,1}^3+2\prob{i,1}\prob{i,2}\prob{i,3}\,,
\end{split}
\end{equation}
where $A(n,k)$ is the permutation number formula $A(n,k)=\frac{n!}{(n-k)!}$.
For $\numbersample=5$, we can get
\begin{equation}
\begin{split}
&\text{Pr}(\answer{1}|\prompt{i};\numbersample=5)=\binom{5}{5}\prob{i,1}^5+\binom{5}{4}\prob{i,1}^4(1-\prob{i,1})+\\
&\binom{5}{3}\prob{i,1}^3(1-\prob{i,1})^2 +\binom{5}{2}\prob{i,1}^2\binom{3}{2}(\prob{i,2}^2\prob{i,3}+\prob{i,3}^2\prob{i,2})\\
&=6\prob{i,1}^5-15\prob{i,1}^4+10\prob{i,1}^3+15\prob{i,1}^2\prob{i,2}\prob{i,3}(\prob{i,2}+\prob{i,3}).
\end{split}
\end{equation}
Lemma \ref{lemma4.5} is proved.

\textbf{Theorem} \ref{theorem4.6}. For two prompting strategies $\prompt{i}$ and $\prompt{i'}$, note $\prob{i,q} = \max\{\prob{i,2},\dots,\prob{i,m}\}$, $\prob{i',q'} = \max\{\prob{i',2},\dots,\prob{i',m}\}$, if $\prob{i,1}-\prob{i,q}<\prob{i',1}-\prob{i',q'}$ and $\prob{i,1}+\prob{i,q}-\prob{i,1}^2-\prob{i,q}^2>\prob{i',1}+\prob{i',q'}-\prob{i',1}^2-\prob{i',q'}^2$, there exits a sufficiently large $\numbersample_0$ such that for $\numbersample>\numbersample_0$, $\text{Pr}(\answer{1}|\prompt{i};\numbersample)<\text{Pr}(\answer{1}|\prompt{i'};\numbersample)$.

\textit{Proof.} According to Khinchin's Law of Large Numbers and Lindeberg-Levy Central Limit Theorem, when $N$ is sufficiently large, each $X_i$ can be approximated by a normal distribution. Specifically, for each $\x_{i,j}$, we have $\x_{i,j} \sim \mathcal{N}(\numbersample \prob{i,j}, \numbersample \prob{i,j}(1-\prob{i,j}))$. Note $M_i = \max(\x_{i,2}, ..., \x_{i,m})$ and 
$\prob{i,q}=\max \{\prob{i,2},\dots,\prob{i,m}\}$, 
the distribution of $M_i$ can be approximated by $M_i \sim \mathcal{N}
(\numbersample \prob{i,n}, \numbersample \prob{i,1}(1-\prob{i,1}))$. 
So $\x_{i,j}-M_i$ obey the normal distribution $\mathcal{\numbersample}
(\numbersample(\prob{i,1} - \prob{i,q}), \numbersample(\prob{i,1}(1-\prob{i,1}) + \prob{i,q}(1-\prob{i,q})))$. Thus, we can get Equation \ref{equation31}:

\vspace{-6pt}
\begin{equation}
\begin{split}
&\text{Pr}(\x_{i,1}>M_i)\\
         &= \text{Pr}(\x_{i,1}-M_i>0) \\
         &= 1- \Phi(f(\prob{i,1},\prob{i,q},\numbersample)), \\
         &\Phi(f(\prob{i,1},\prob{i,q},\numbersample))=\\
         & \Phi(\sqrt{\frac{\numbersample}{\prob{i,1}(1-\prob{i,1}) + \prob{i,q}(1-\prob{i,q})}} (\prob{i,q}-\prob{i,1})),
\end{split}
\label{equation31}
\end{equation}
where $\Phi$ is the standard normal cumulative distribution function. This also holds for any other $\text{Pr}(\x_{i',1}>M_i')$. If $\prob{i,1}-\prob{i,q}<\prob{i',1}-\prob{i',q'}$ and $\prob{i,1}+\prob{i,q}-\prob{i,1}^2-\prob{i,q}^2>\prob{i',1}+\prob{i',q'}-\prob{i',1}^2-\prob{i',q'}^2$, we can get $\prob{i,q}-\prob{i,1}>\prob{i',q}-\prob{i',1}$ and $\prob{i,1}(1-\prob{i,1}) + \prob{i,q}(1-\prob{i,q})<\prob{i',1}(1-\prob{i',1}) + \prob{i',q'}(1-\prob{i',q'})$, and $\Phi(f(\prob{i,1},\prob{i,q},\numbersample))>\Phi(f(\prob{i',1},\prob{i',q'},\numbersample))$. So there exists a large $\numbersample_0$ such that for  $\numbersample>\numbersample_0$, $\text{Pr}(\x_{i',1}>M_i')>\text{Pr}(\x_{i,1}>M_i)$, \textit{i.e.}, $\text{Pr}(\answer{1}|\prompt{i};\numbersample)>\text{Pr}(\answer{1}|\prompt{i'};\numbersample)$. Theorem \ref{theorem4.6} is proved.

\section{Detailed Results}
\label{detailed results}

\begin{figure*}[!t]
    \centering
    \includegraphics[width=1\linewidth]{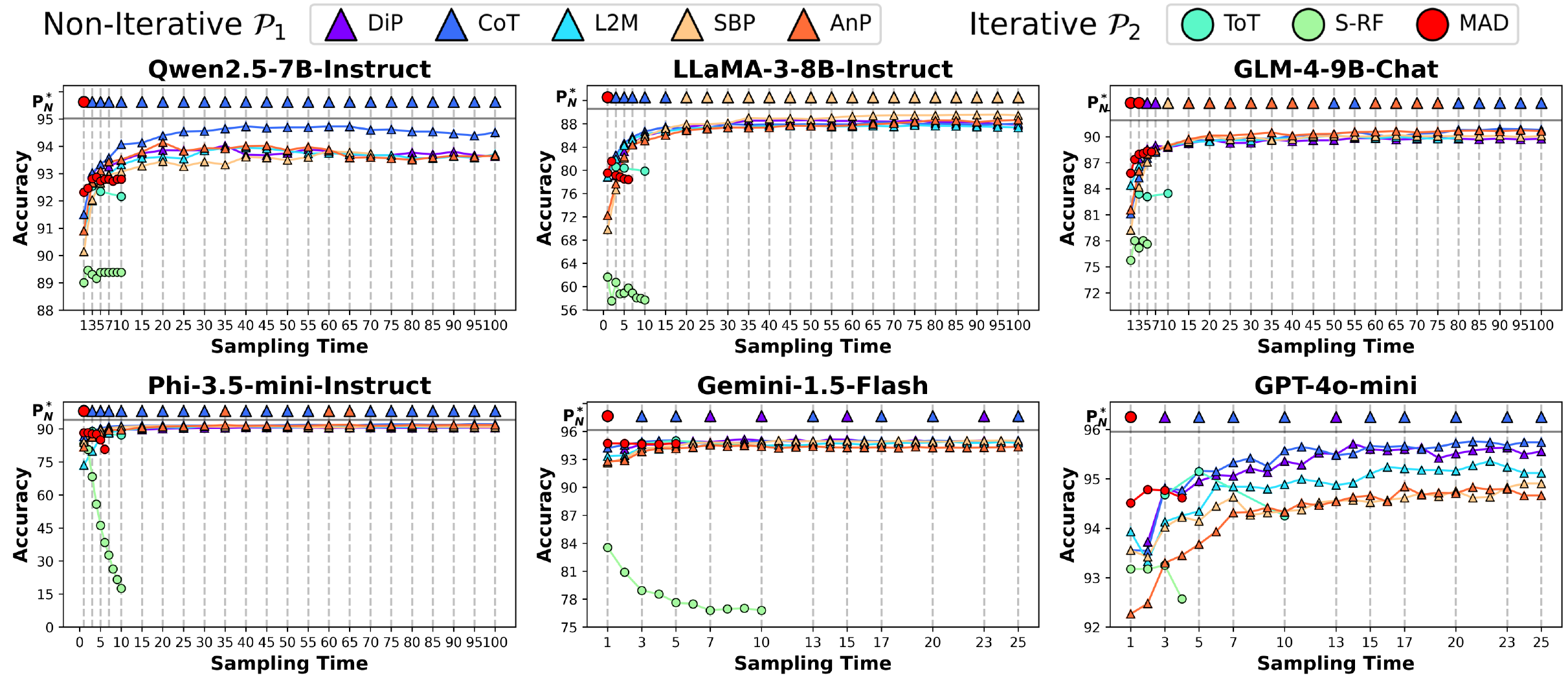}
    \caption{Performance of each prompting strategy under given sampling time $\numbersample$ on GSM8K.}
    \label{GSM8K-N}
\end{figure*}

\begin{figure*}[!h]
    \centering
    \includegraphics[width=1\linewidth]{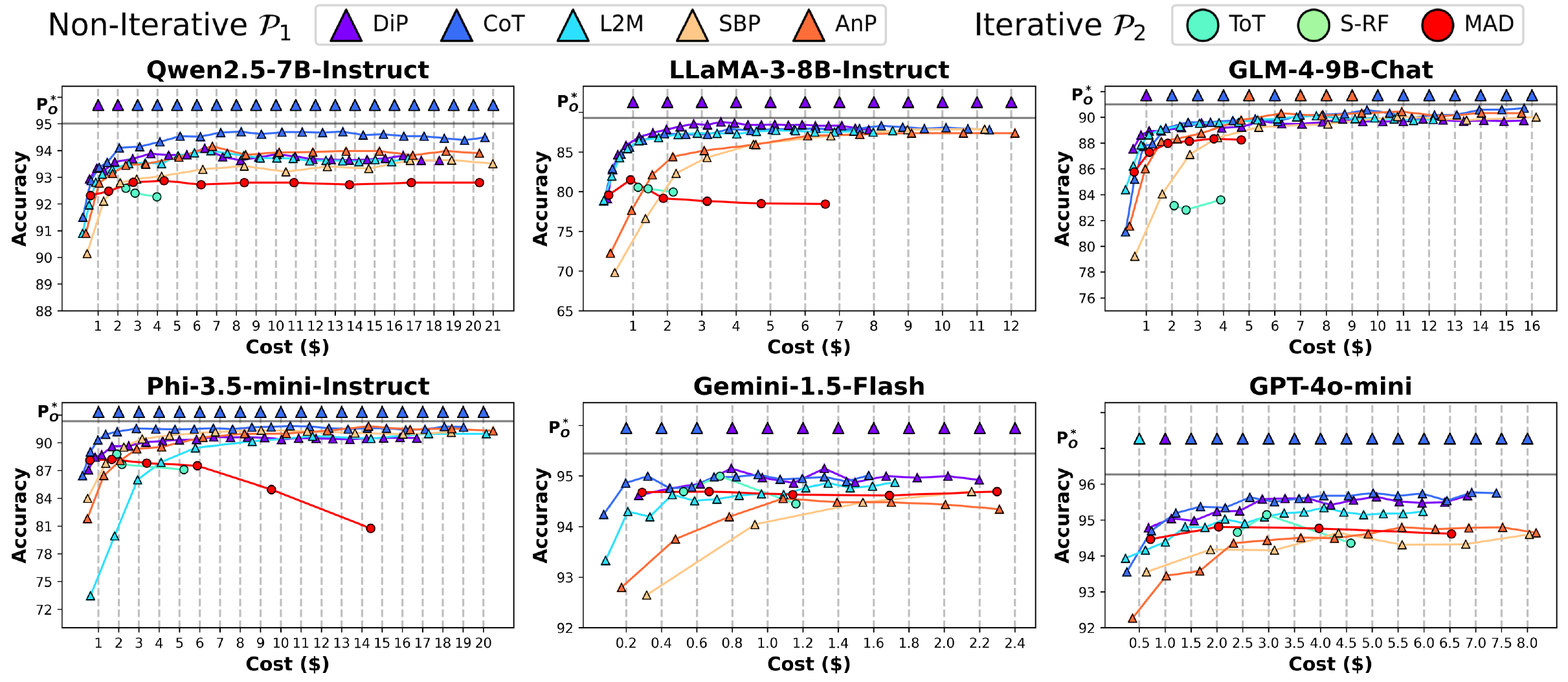}
    \caption{Performance of each prompting strategy under given cost $\overhead$ on GSM8K.}
    \label{GSM8K-cost}
\end{figure*}

\begin{figure*}
    \centering
    \includegraphics[width=1\linewidth]{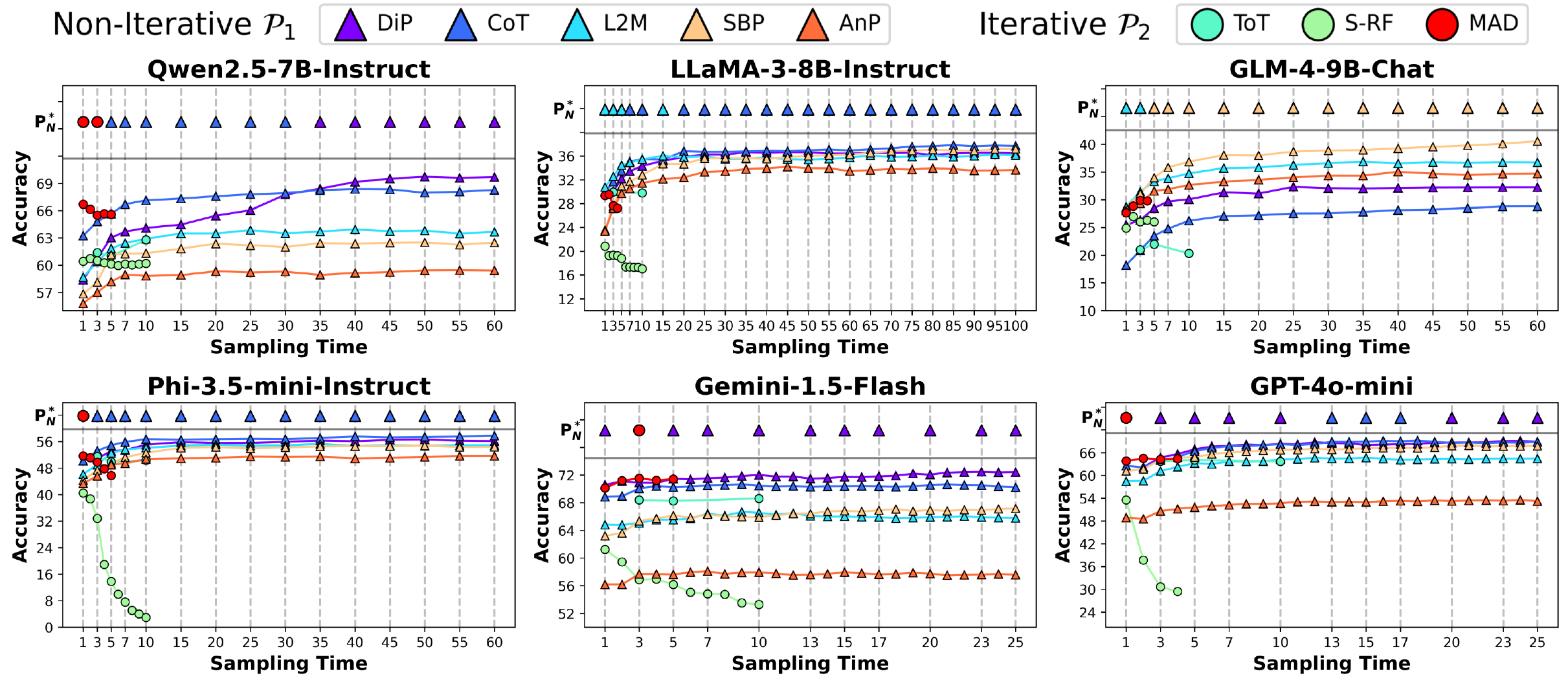}
    \caption{Performance of each prompting strategy under given sampling time $\numbersample$ on GSM-Hard.}
    \label{GSM-hard-N}
\end{figure*}

\begin{figure*}[!h]
    \centering
    \includegraphics[width=1\linewidth]{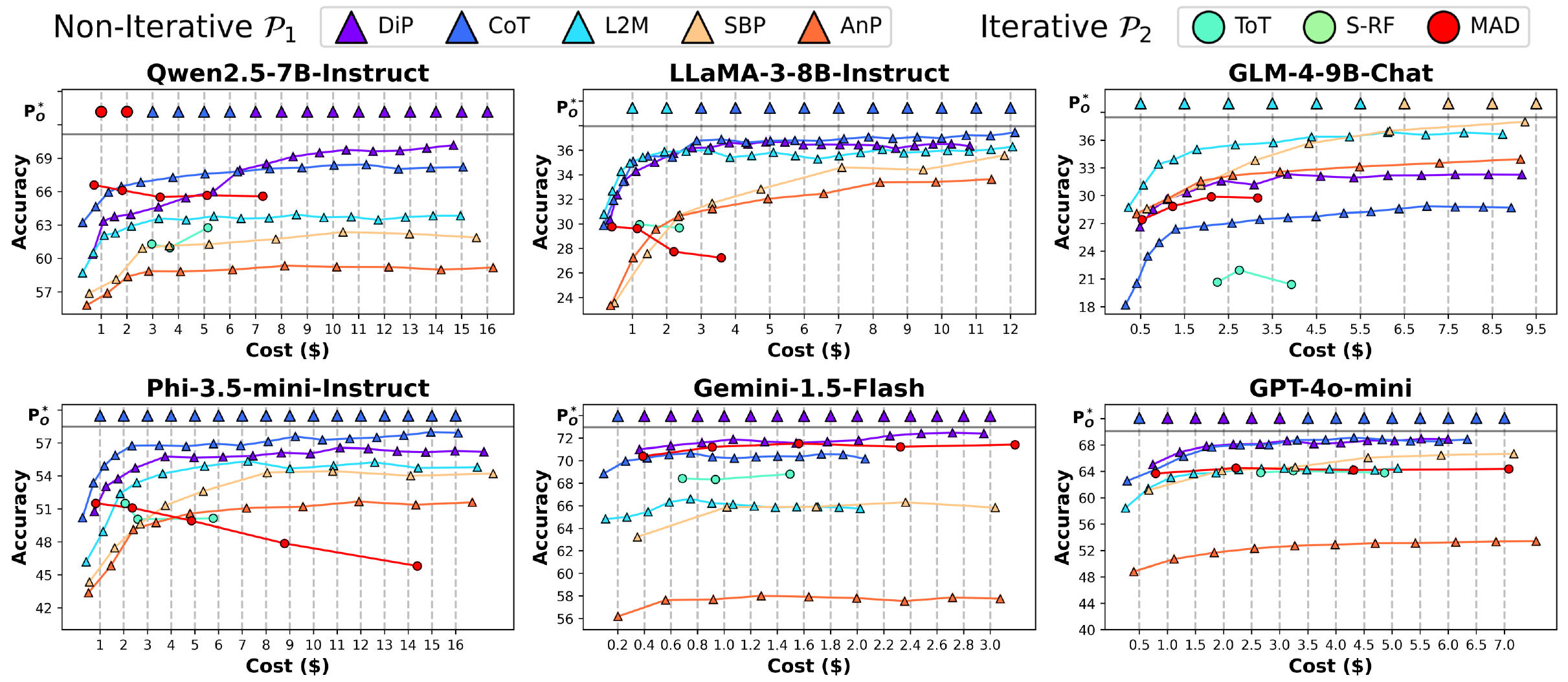}
    \caption{Performance of each prompting strategy under given cost $\overhead$ on GSM-Hard.}
    \label{GSM-hard-cost}
\end{figure*}

\begin{figure*}[!h]
    \centering
    \includegraphics[width=1\linewidth]{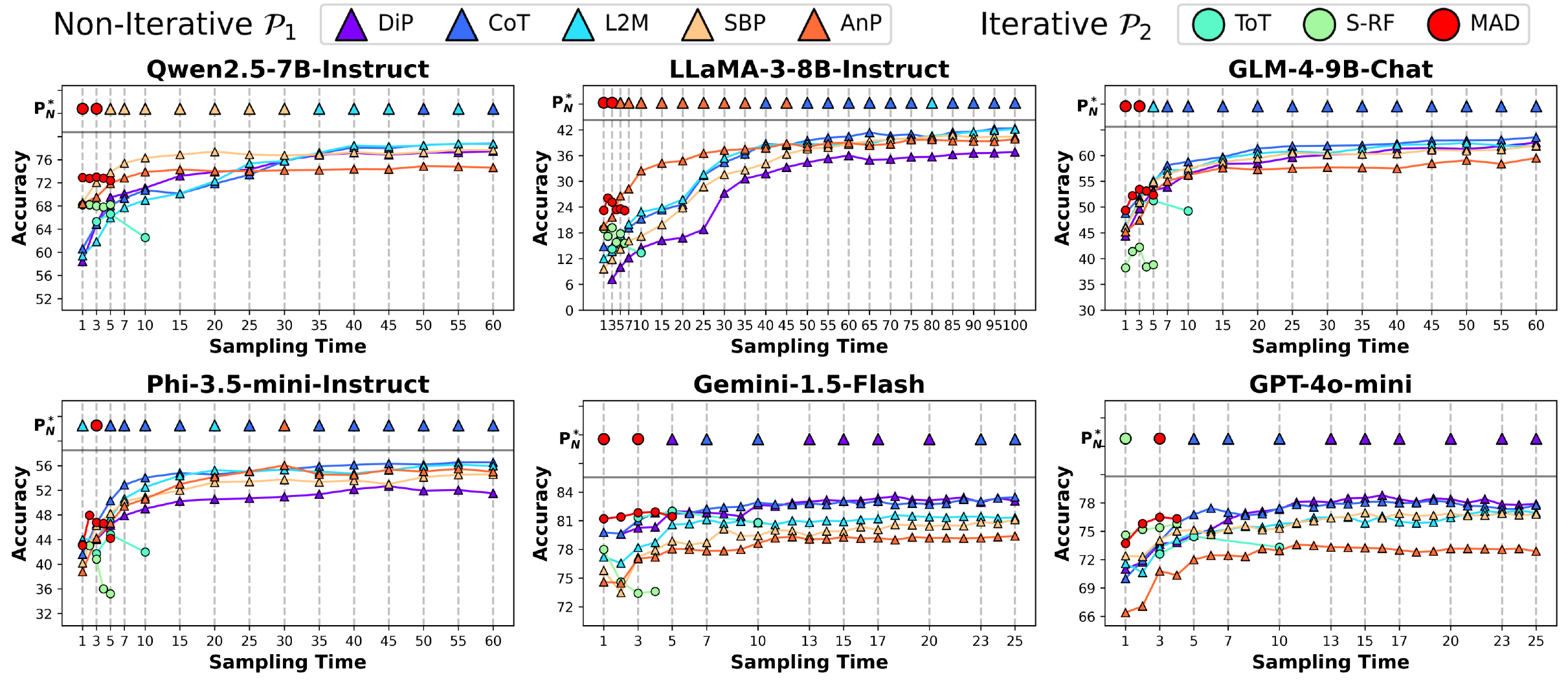}
    \caption{Performance of each prompting strategy under given sampling time $\numbersample$ on MATH.}
    \label{MATH-N}
\end{figure*}

\begin{figure*}[!h]
    \centering
    \includegraphics[width=1\linewidth]{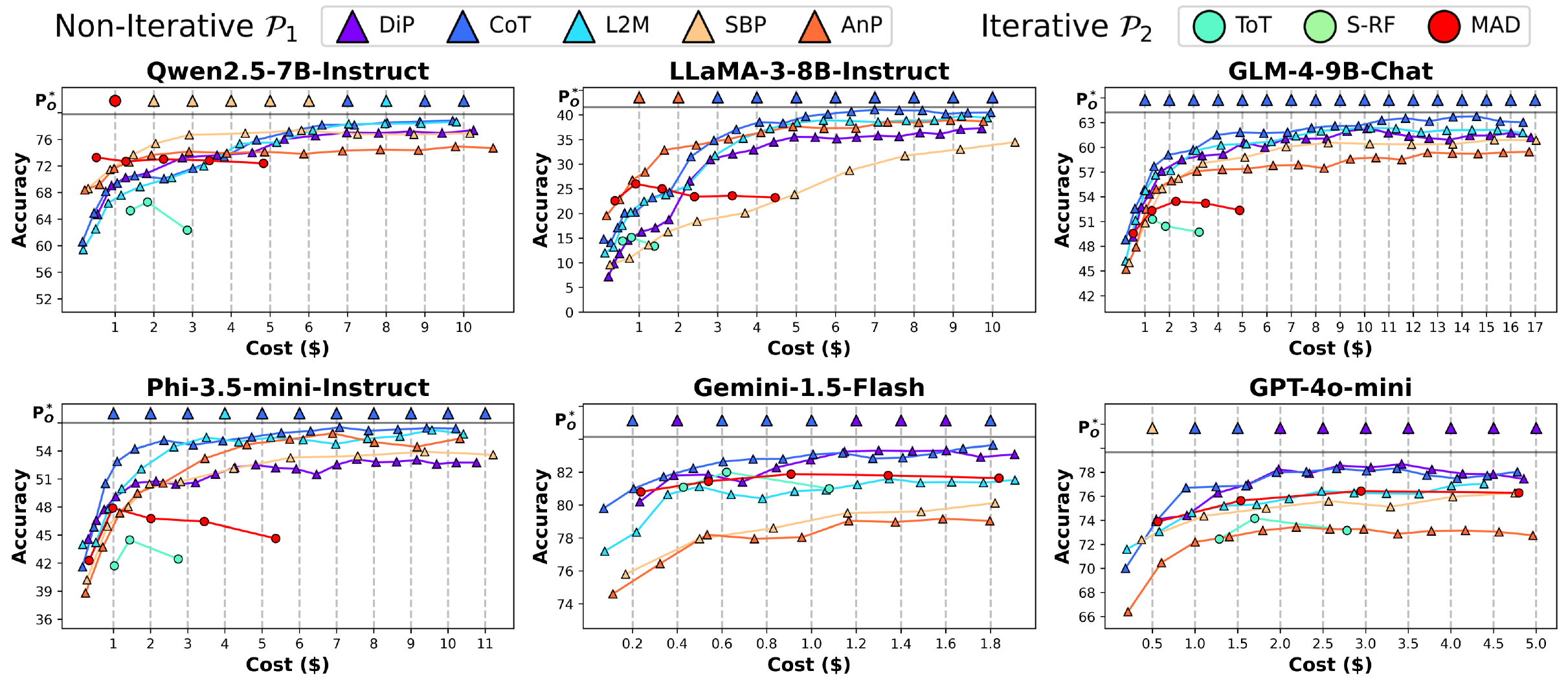}
    \caption{Performance of each prompting strategy under given cost $\overhead$ on MATH.}
    \label{MATH-cost}
\end{figure*}

\begin{figure*}[!h]
    \centering
    \includegraphics[width=1\linewidth]{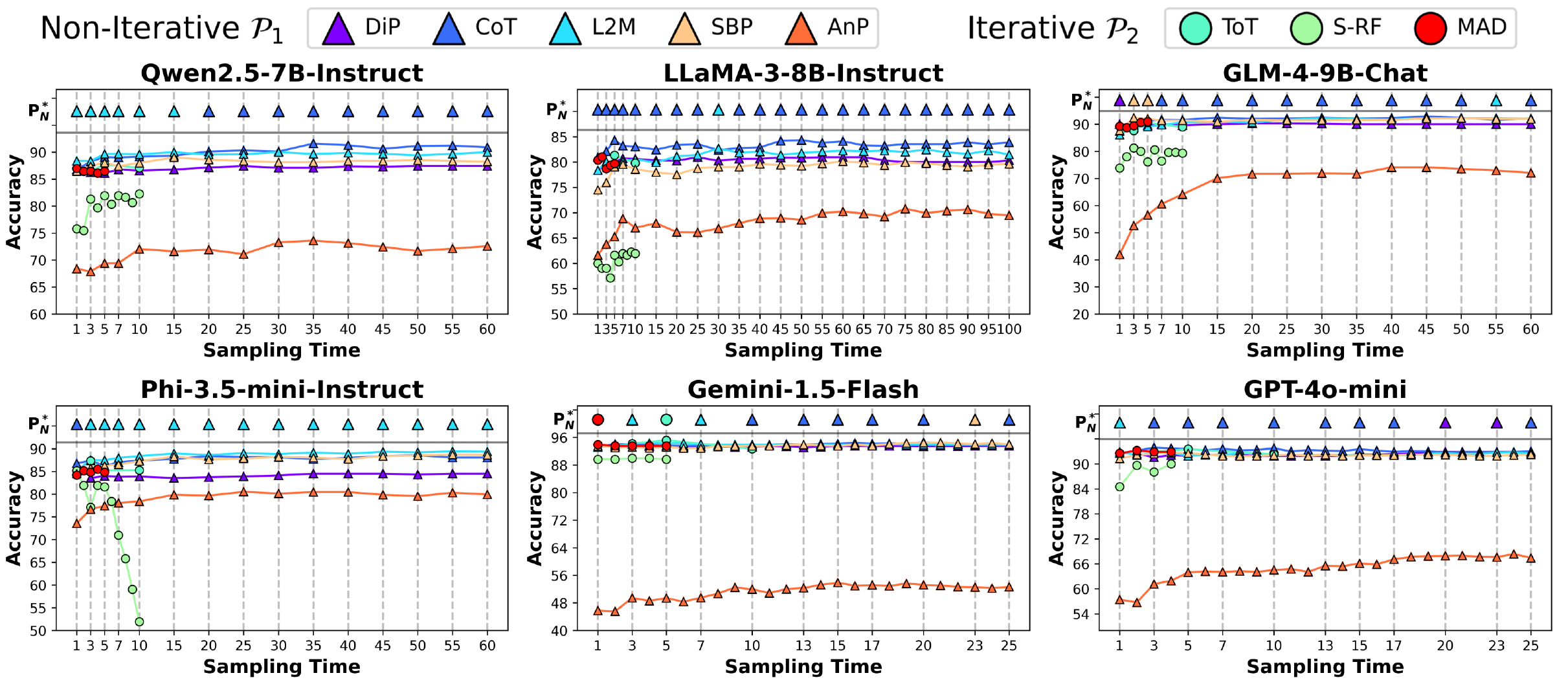}
    \caption{Performance of each prompting strategy under given sampling time $\numbersample$ on MMLU.}
    \label{MMLU-N}
\end{figure*}

\begin{figure*}[!h]
    \centering
    \includegraphics[width=1\linewidth]{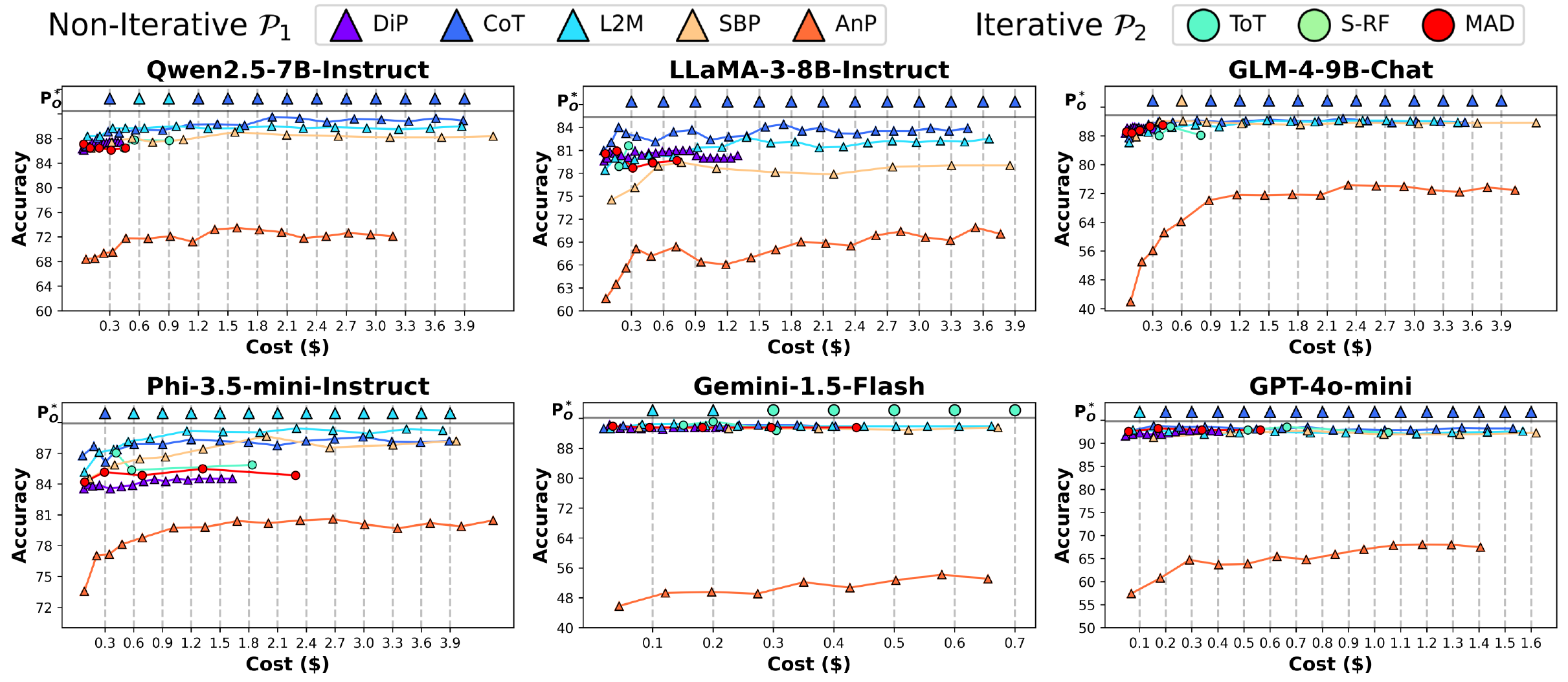}
    \caption{Performance of each prompting strategy under given cost $\overhead$ on MMLU.}
    \label{MMLU-cost}
\end{figure*}

Here we display the scaling performance of different prompting strategies on each LLM and benchmark under given sampling time $\numbersample$ and cost $\overhead$, as shown in Figures \ref{GSM8K-N} to \ref{MMLU-cost}.
We find that, aside from CoT, DiP also exhibits superior performance compared to other complex prompting strategies on certain models and datasets, \textit{e.g.}, GPT-4o-mini on MATH. This also comes from the two reasons, \textit{i.e.}, DiP has more hard questions and easy questions, and a flat probability distribution of wrong answers on the specific dataset. This phenomenon is particularly prominent on powerful LLMs such as Gemini-1.5-Flash on GSM8K and GSM-Hard, where DiP and CoT exhibit comparable performance. Almost 83\% of results satisfy that CoT or DiP performs best as significantly scaling. Besides, this trend is also observed on other prompting strategies on few datasets and models. This encourages us to fully unleash the potential of simple prompting strategies, and indicates that the scaling performance does not only depend on the prompting strategies' pass@1 accuracy.

\section{More Discussions on Improving the Scaling Performance}
\label{more discussions}
In this section, we will discuss more about our further exploration of the two ways to improve the scaling performance. We display more results of (1) adaptively scaling based on the question difficulty, (2) dynamically choosing the optimal $\prompt{i}$ and (3) combining adaptively scaling and dynamically choosing the optimal $\prompt{i}$ in Section \ref{appendix adaptively}, \ref{appendix dynamical} and \ref{appendix combine}, respectively. 

\subsection{Adaptively Scaling Based on the Difficulty}
\label{appendix adaptively}
We use the following prompt to force the LLM to determine if the question is hard for given $\prompt{i}$.

\vspace{10pt}
\itshape 
Question:

\{question\}

\vspace{1\baselineskip}
Using the method \#\{method\}\# to solve the question:

\{description\}

\vspace{1\baselineskip}

If the method is more likely to get the right answer, the question is easy. Otherwise, if the method is more likely to get the wrong answer, the question is hard. Please determine the difficulty of the question for the used method, and answer in the following JSON format.

\{"Difficulty": "Easy or Hard", "Reason": ""\}
\upshape
\vspace{10pt}

\begin{figure*}[htbp]
    \centering
    \includegraphics[width=0.981\linewidth]{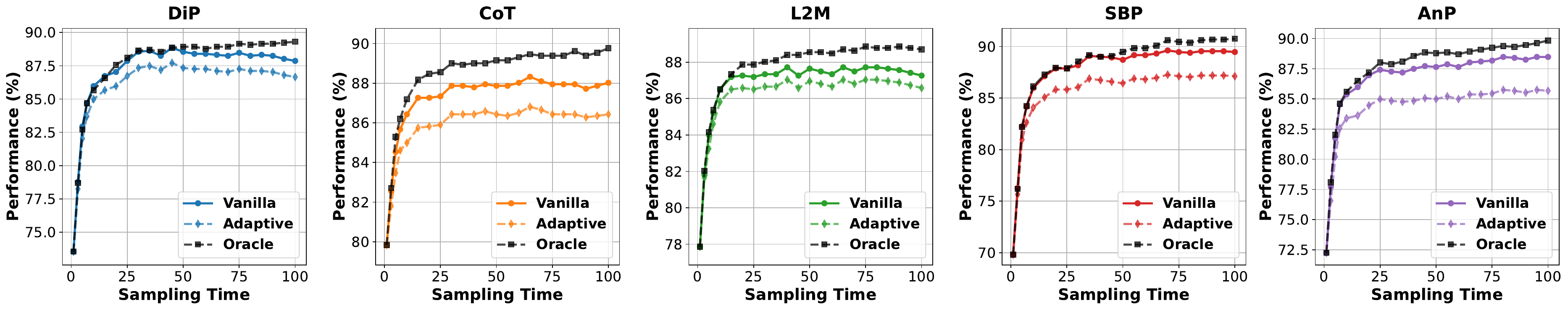}
    \caption{Results of adaptively scaling based on the question difficulty on Llama-3-8B-Instruct on GSM8K.}
    \label{adaptive llama}
\end{figure*}

\begin{figure*}[htbp]
    \centering
    \includegraphics[width=0.981\linewidth]{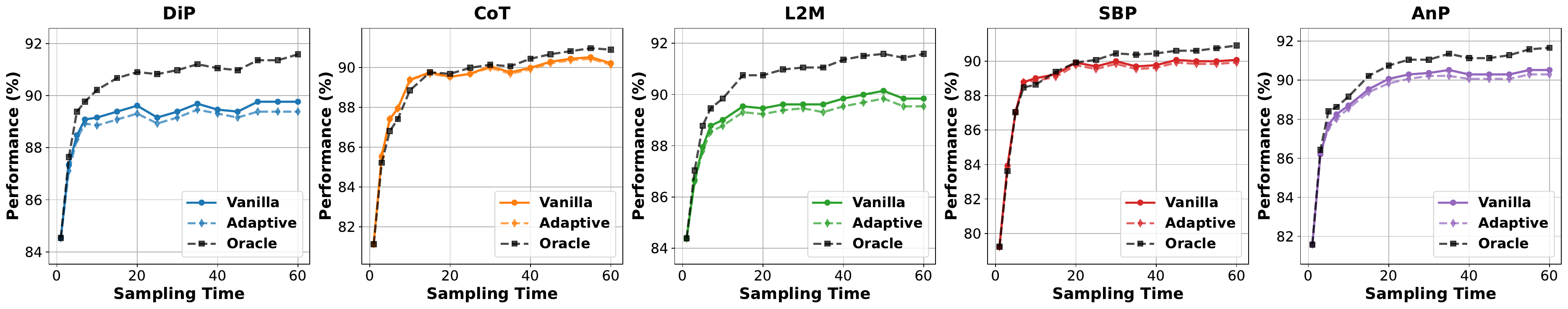}
    \caption{Results of adaptively scaling based on the question difficulty on GLM-4-9B-Chat on GSM8K.}
    \label{adaptive glm}
\end{figure*}

\begin{figure*}[htbp]
    \centering
    \includegraphics[width=0.981\linewidth]{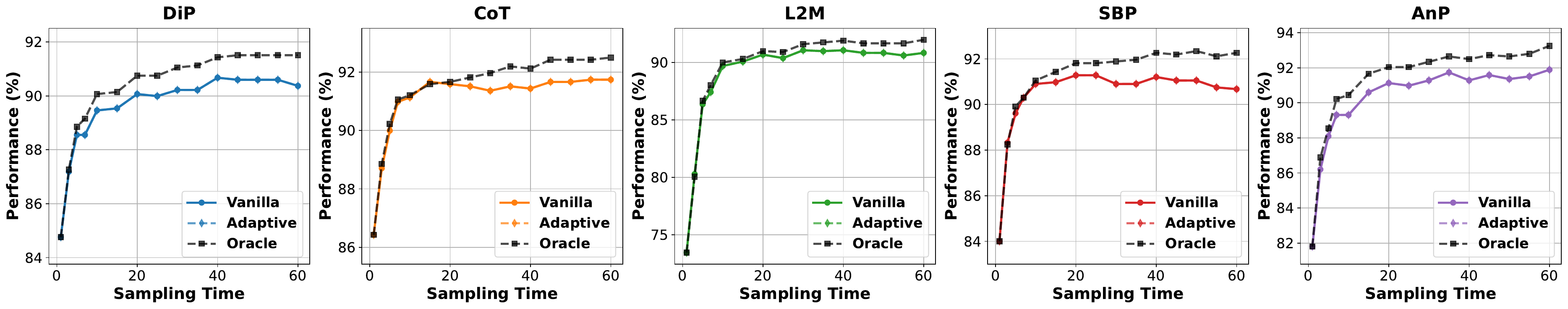}
    \caption{Results of adaptively scaling based on the question difficulty on Phi-3.5-mini-Instruct on GSM8K.}
    \label{adaptive phi}
\end{figure*}

\begin{figure*}[htbp]
    \centering
    \includegraphics[width=0.981\linewidth]{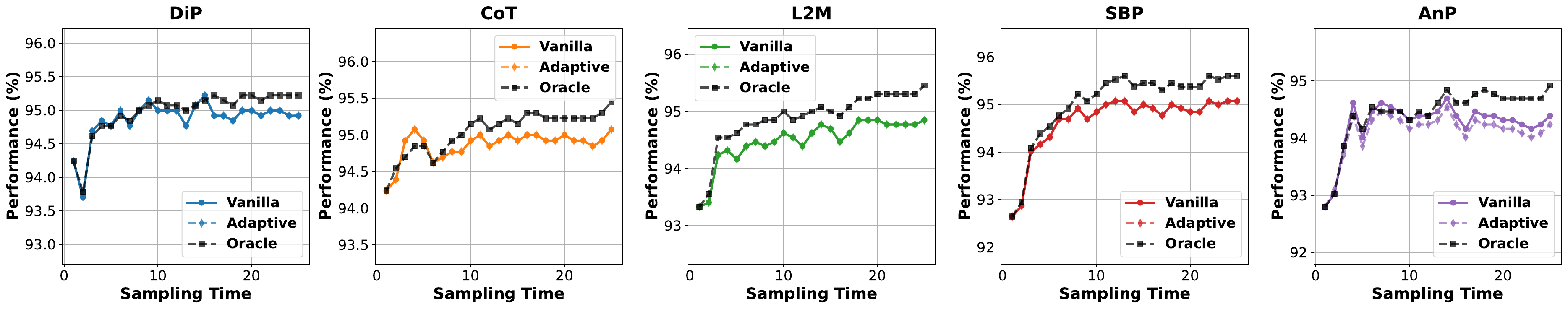}
    \caption{Results of adaptively scaling based on the question difficulty on Gemini-1.5-Flash on GSM8K.}
    \label{adaptive gemini}
\end{figure*}

\begin{figure*}[htbp]
    \centering
    \includegraphics[width=0.981\linewidth]{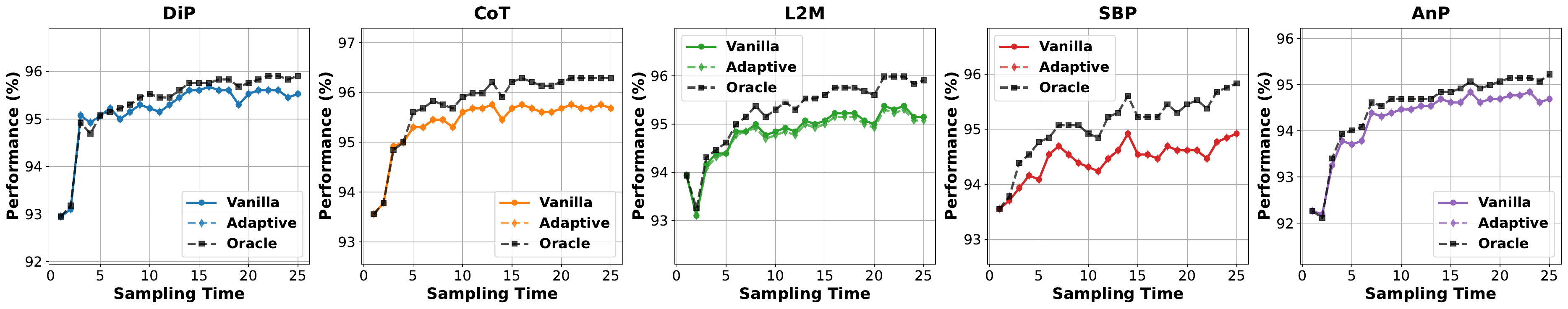}
    \caption{Results of adaptively scaling based on the question difficulty on GPT-4o-mini on GSM8K.}
    \label{adaptive gpt}
\end{figure*}

Figures \ref{adaptive llama} to \ref{adaptive gpt} report the results of each prompting strategy when adaptively scaling based on the question difficulty.
Our experiment results show that LLMs cannot accurately judge the difficulty of the input question most of the time, thus even leading to reduced performance. Nevertheless, this method is theoretically capable of enhancing the scaling performance, thereby motivating us to explore other approaches to accurately assess the question difficulty.

\subsection{Dynamically Choosing the Optimal $\prompt{i}$}
\label{appendix dynamical}
\begin{figure}[htbp]
    \centering
    \includegraphics[width=1\linewidth]{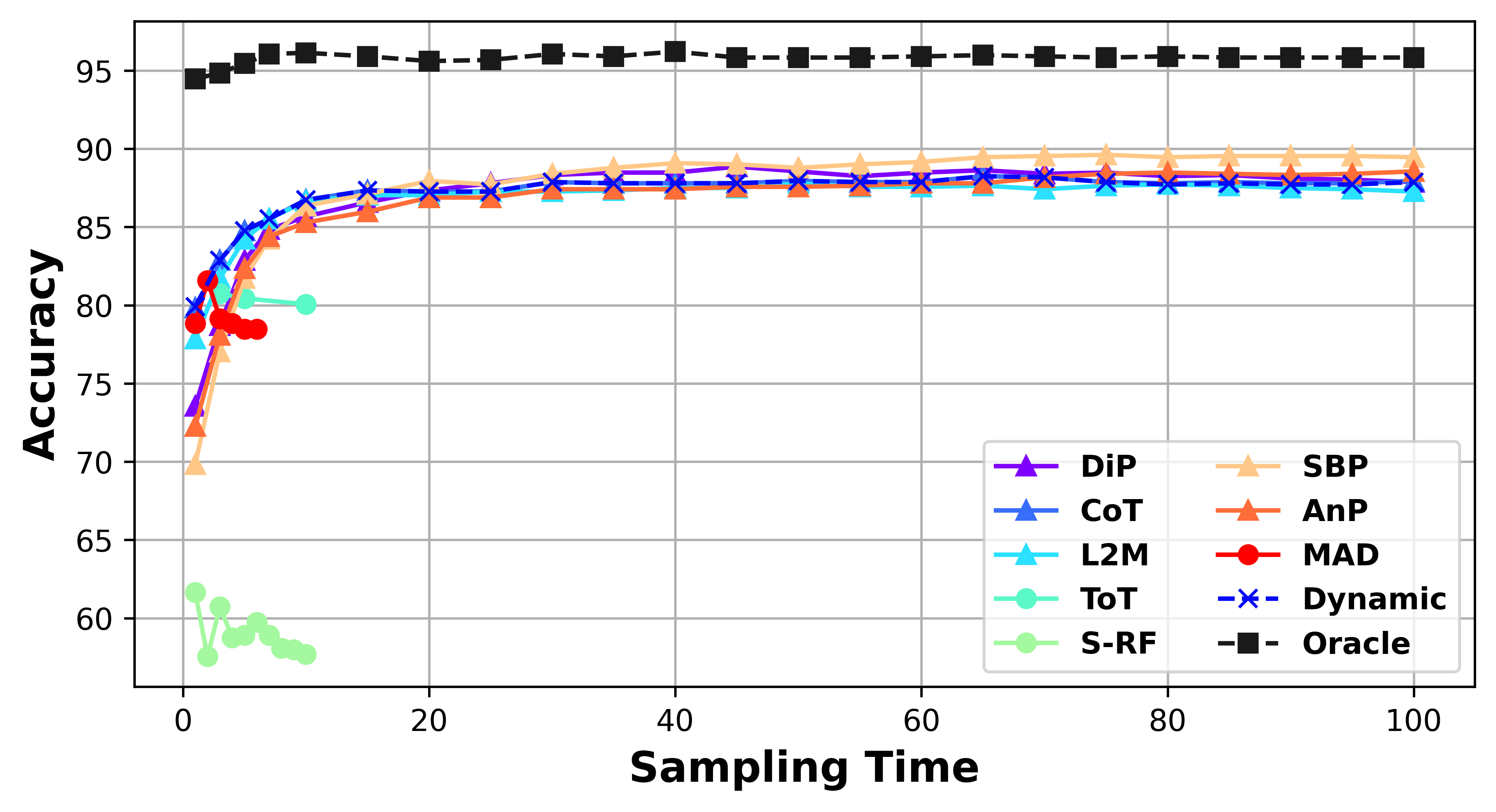}
    \caption{Results of dynamically choosing the optimal $\prompt{i}$ on LLaMA-3-8B-Instruct on GSM8K.}
    \label{dynamic gsm8k llama}
\end{figure}

\begin{figure}[htbp]
    \centering
    \includegraphics[width=1\linewidth]{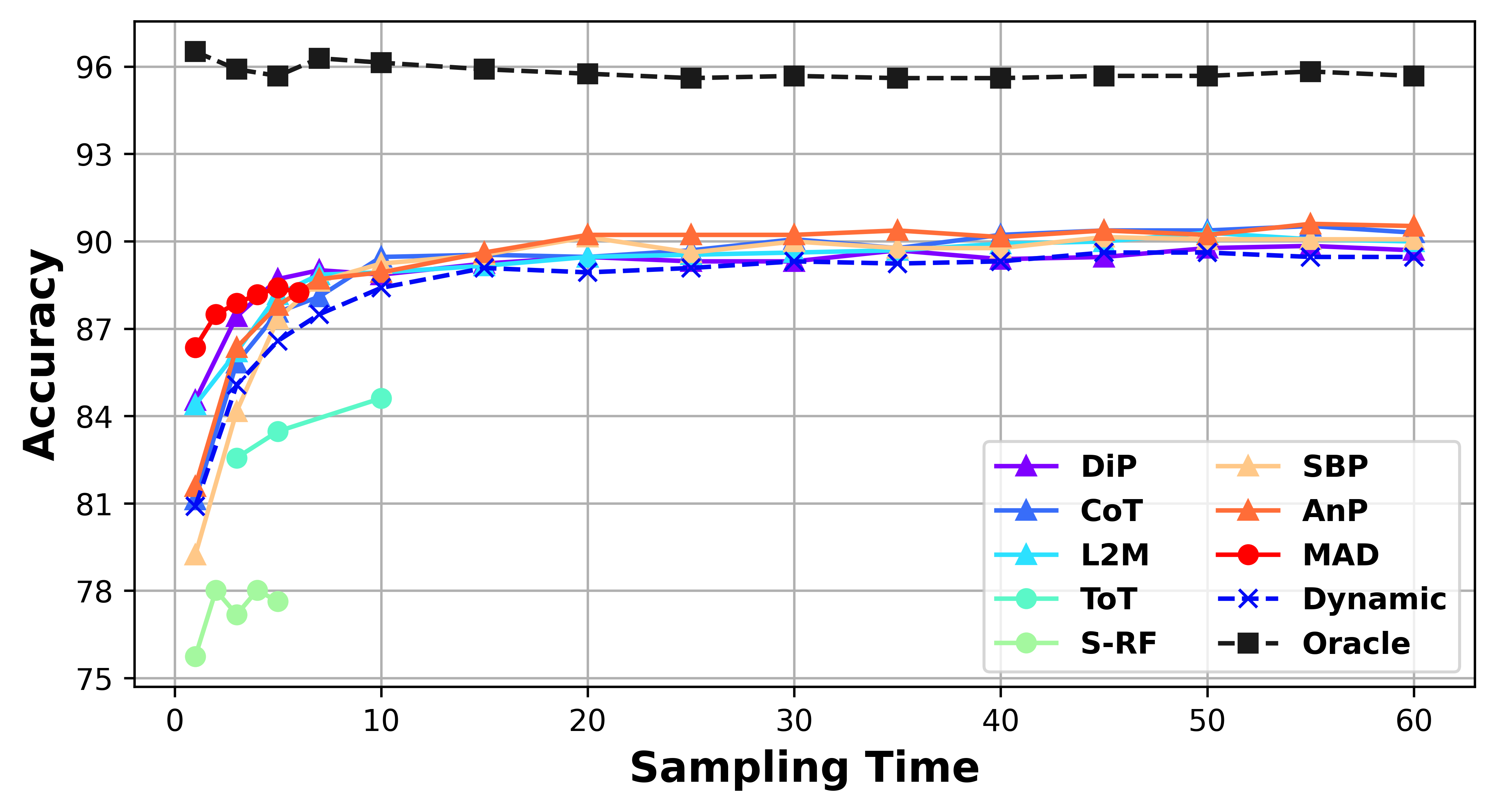}
    \caption{Results of dynamically choosing the optimal $\prompt{i}$ on GLM-9B-Chat on GSM8K.}
    \label{dynamic gsm8k glm}
\end{figure}

\begin{figure}[htbp]
    \centering
    \includegraphics[width=1\linewidth]{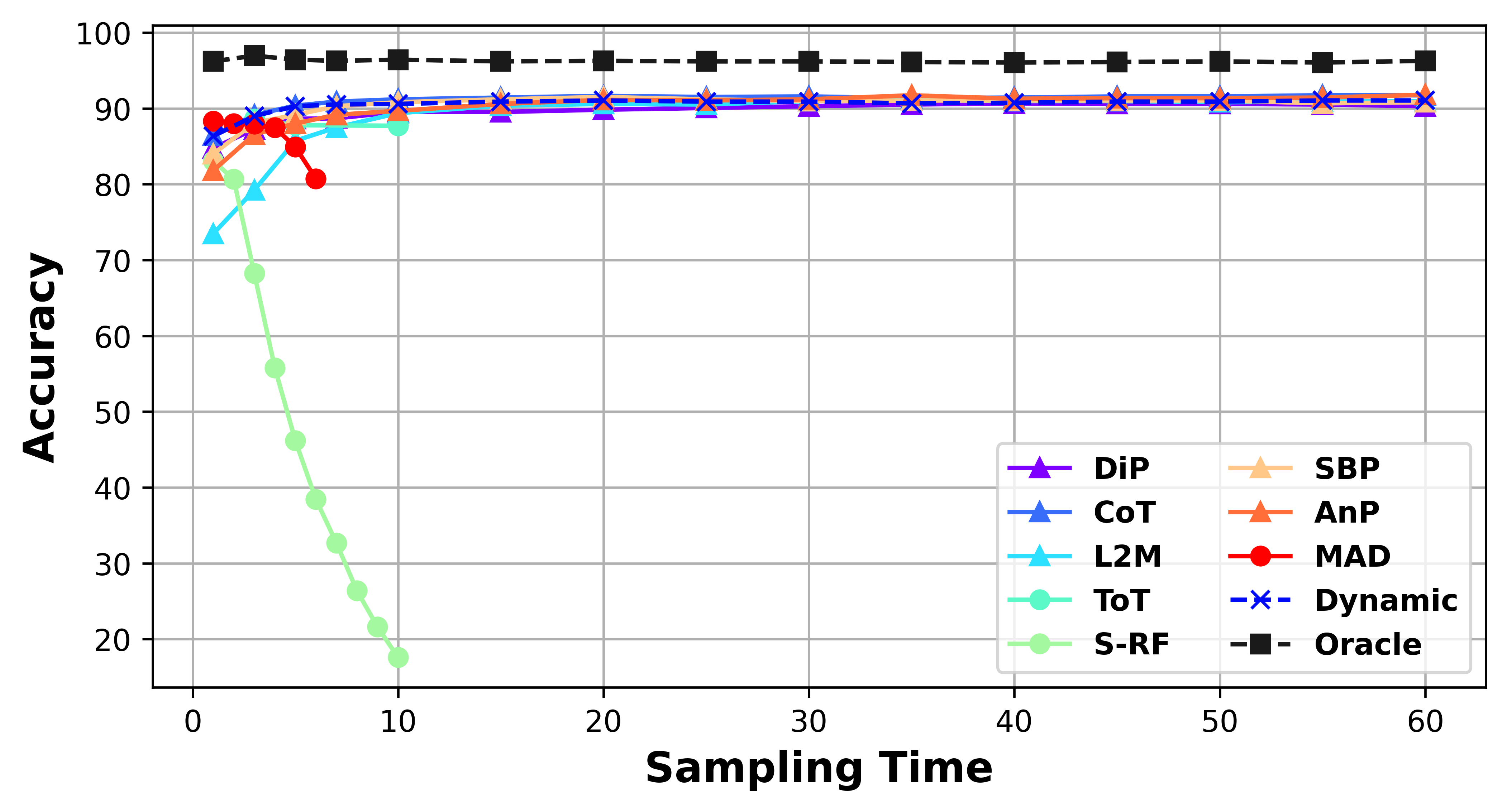}
    \caption{Results of dynamically choosing the optimal $\prompt{i}$ on Phi-3.5-mini-Instruct on GSM8K.}
    \label{dynamic gsm8k phia}
\end{figure}

\begin{figure}[htbp]
    \centering
    \includegraphics[width=1\linewidth]{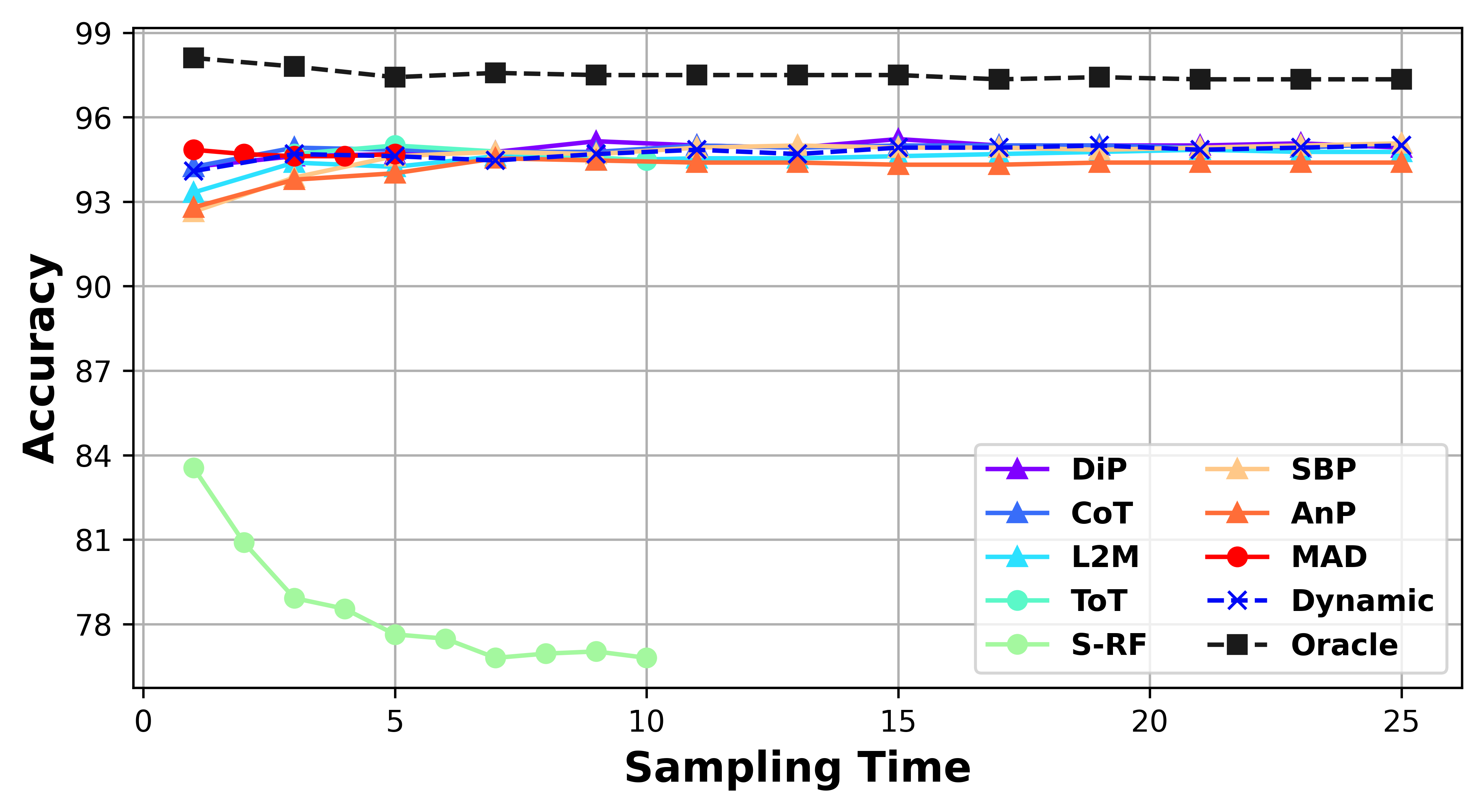}
    \caption{Results of dynamically choosing the optimal $\prompt{i}$ on Gemini-1.5-Flash on GSM8K.}
    \label{dynamic gsm8k gemini}
\end{figure}

\begin{figure}[htbp]
    \centering
    \includegraphics[width=1\linewidth]{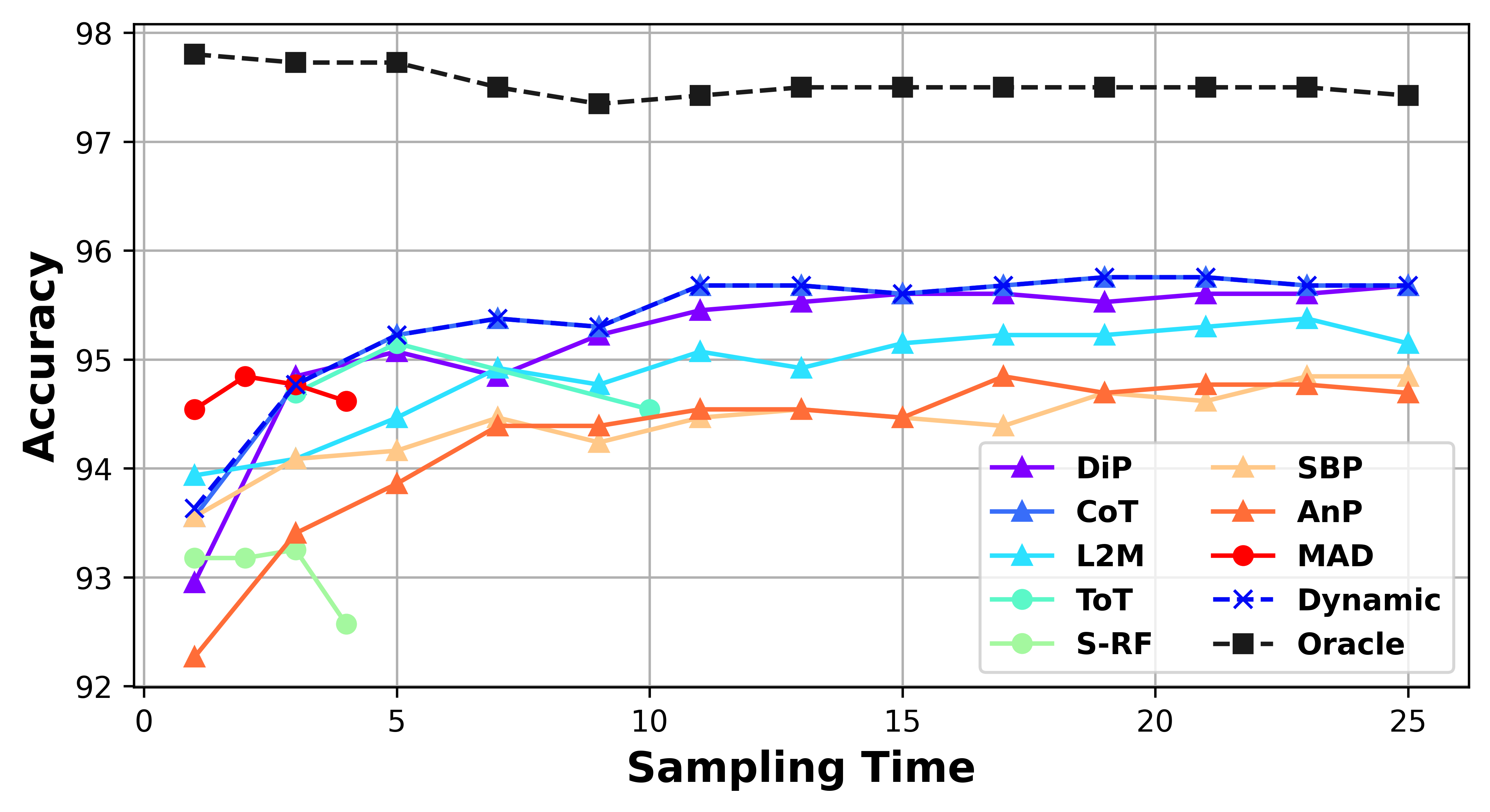}
    \caption{Results of dynamically choosing the optimal $\prompt{i}$ on GPT-4o-mini on GSM8K.}
    \label{dynamic gsm8k gpt}
\end{figure}

Figures \ref{dynamic gsm8k llama} to \ref{dynamic gsm8k gpt} display the results on GSM8K on LLaMA-3-8B-Instruct, GLM-4-9B-Chat, Phi-3.5-mini-Instruct, Gemini-1.5-Flash and GPT-4o-mini, respectively. It can be observed that all LLMs tend to believe that CoT is the best prompting strategy, while CoT does not excel at every question. With oracles to provide the optimal $\prompt{i}$ labels, all LLMs demonstrate significant performance improvements, even with only one sampling time, proving the enormous potential of this method. we will explore how to approach this upper bound in the future.

\subsection{Combining Adaptively Scaling and Dynamically Choosing the Optimal $\prompt{i}$}
\label{appendix combine}
\begin{figure}[!h]
    \centering
    \includegraphics[width=1\linewidth]{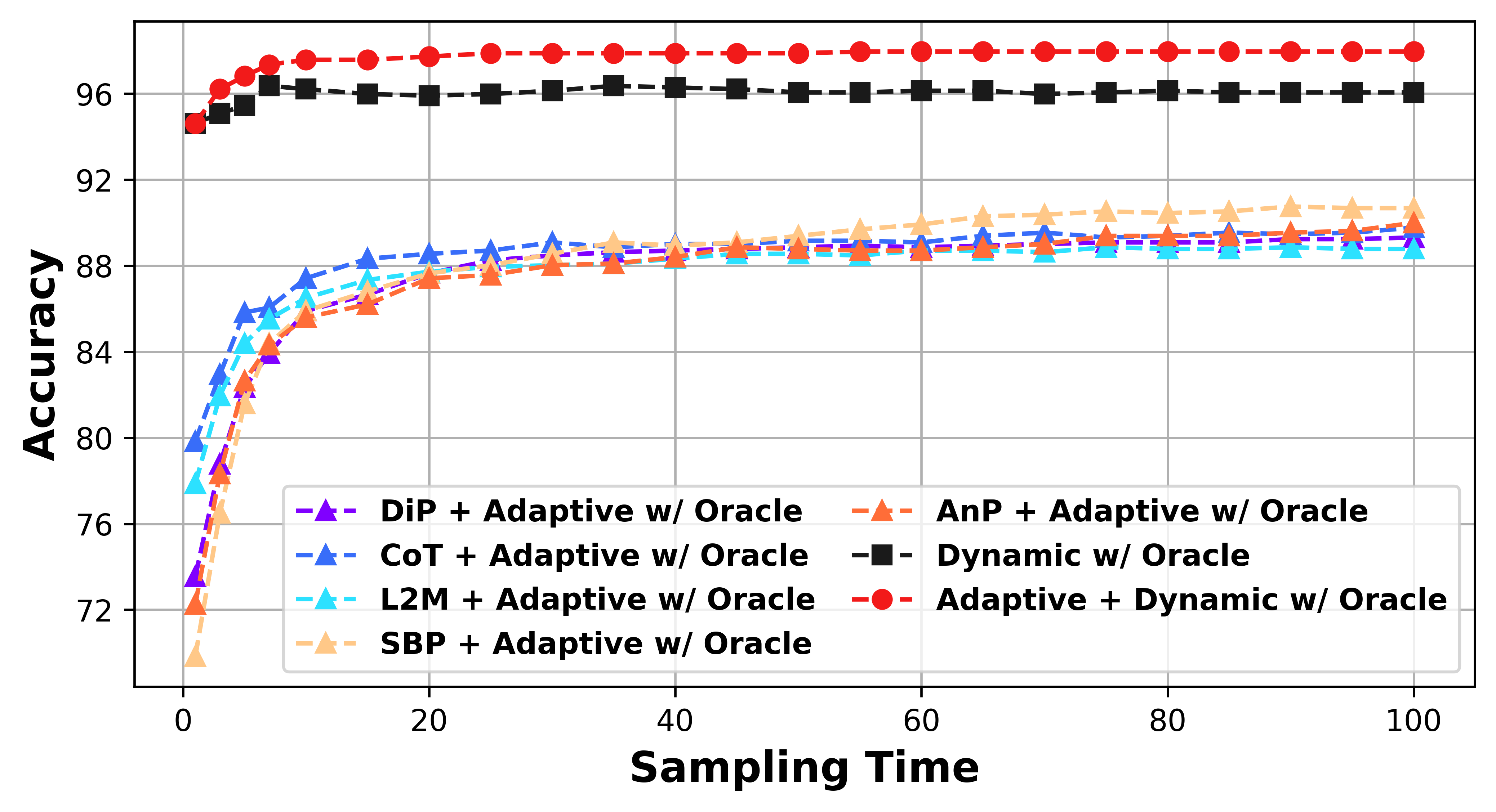}
    \caption{Results of combining adaptively scaling and dynamically choosing the optimal $\prompt{i}$ on LLaMA-3-8B-Instruct on GSM8K.}
    \label{combine gsm8k llama}
\end{figure}

\begin{figure}[!h]
    \centering
    \includegraphics[width=1\linewidth]{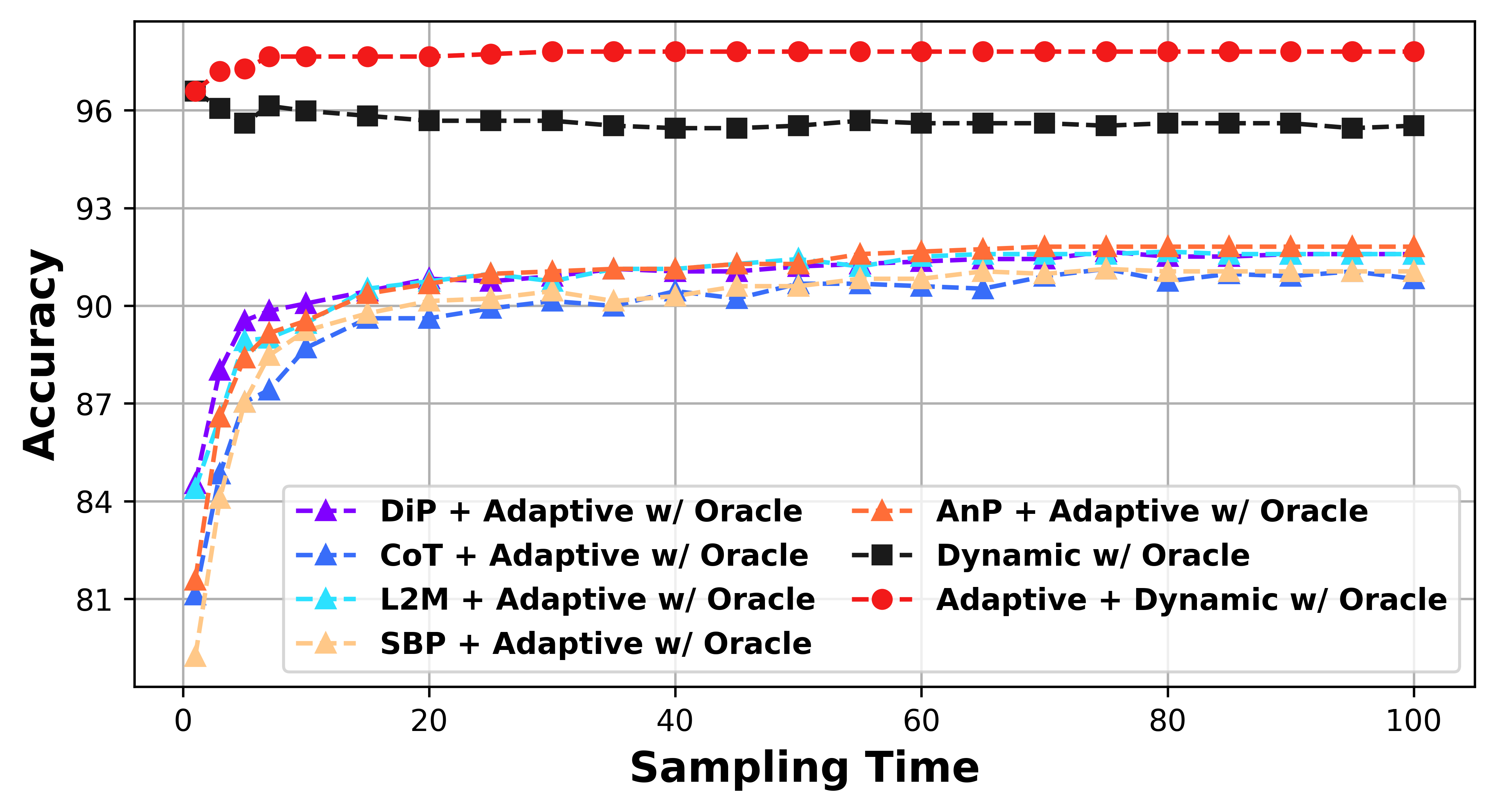}
    \caption{Results of combining adaptively scaling and dynamically choosing the optimal $\prompt{i}$ on GLM4-9B-Chat on GSM8K.}
    \label{combine gsm8k glm}
\end{figure}

\begin{figure}[!h]
    \centering
    \includegraphics[width=1\linewidth]{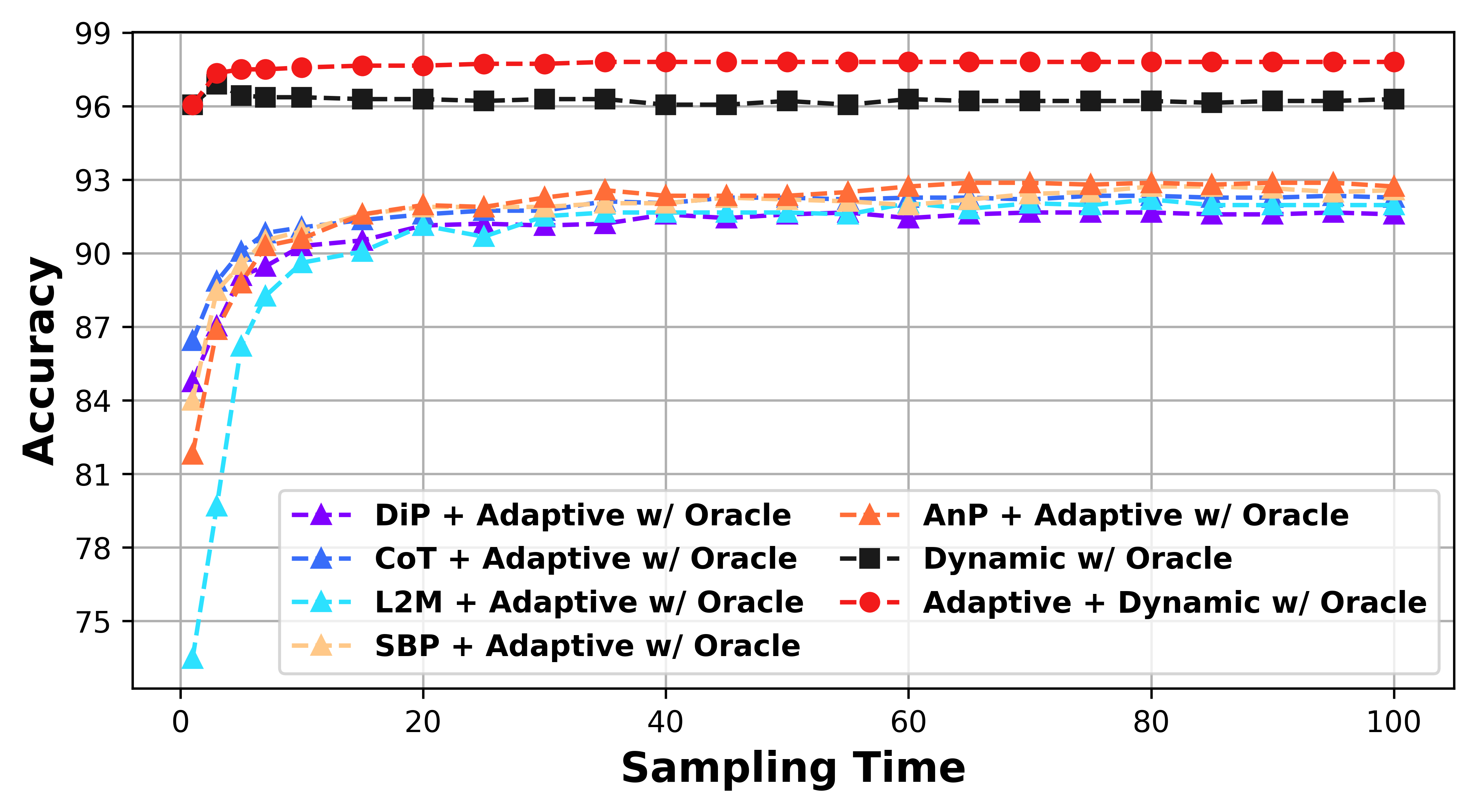}
    \caption{Results of combining adaptively scaling and dynamically choosing the optimal $\prompt{i}$ on Phi-3.5-mini-Instruct on GSM8K.}
    \label{combine gsm8k phi}
\end{figure}

\begin{figure}[!h]
    \centering
    \includegraphics[width=1\linewidth]{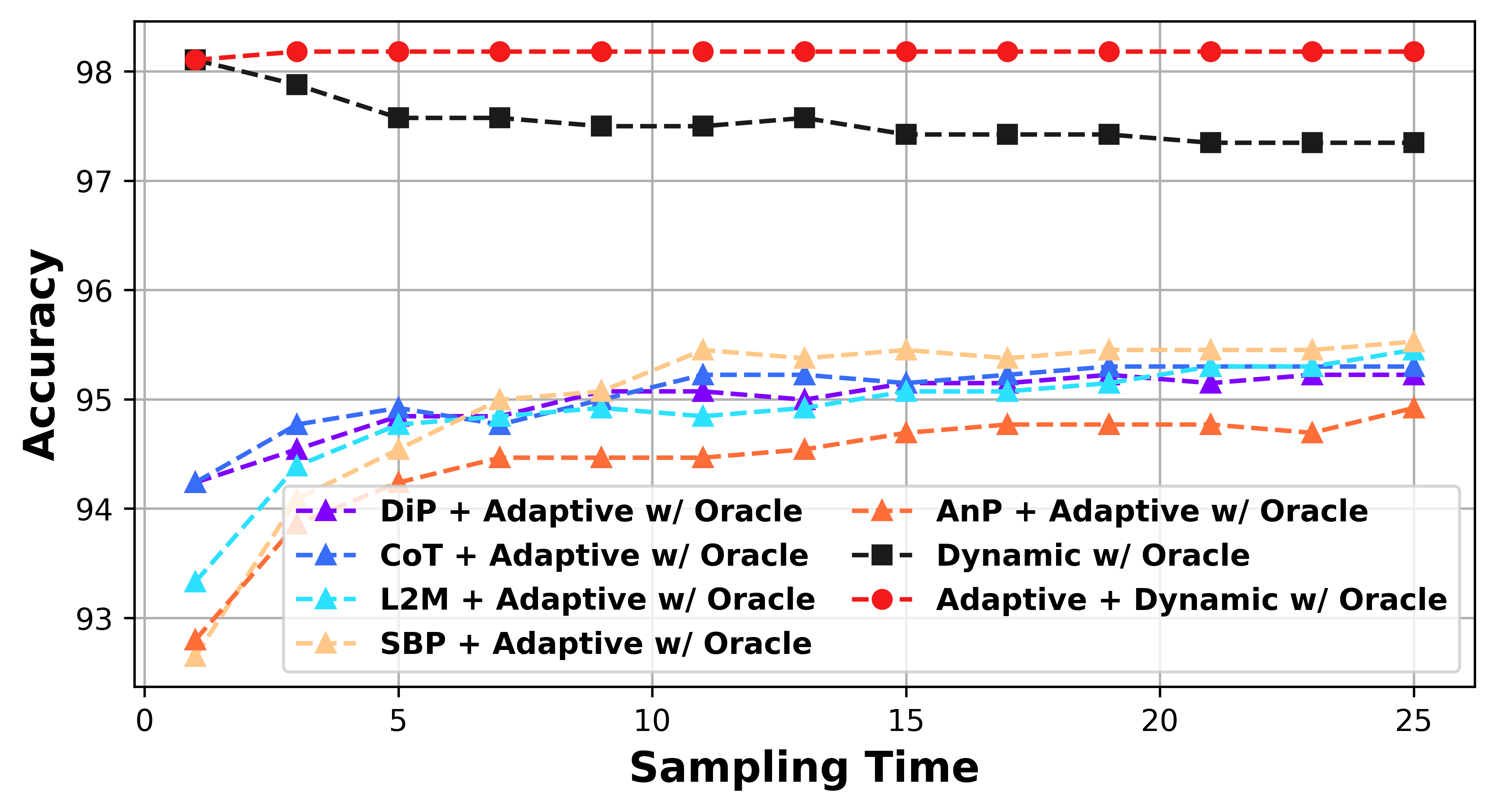}
    \caption{Results of combining adaptively scaling and dynamically choosing the optimal $\prompt{i}$ on Gemini-1.5-Flash on GSM8K.}
    \label{combine gsm8k gemini}
\end{figure}

\begin{figure}[!h]
    \centering
    \includegraphics[width=1\linewidth]{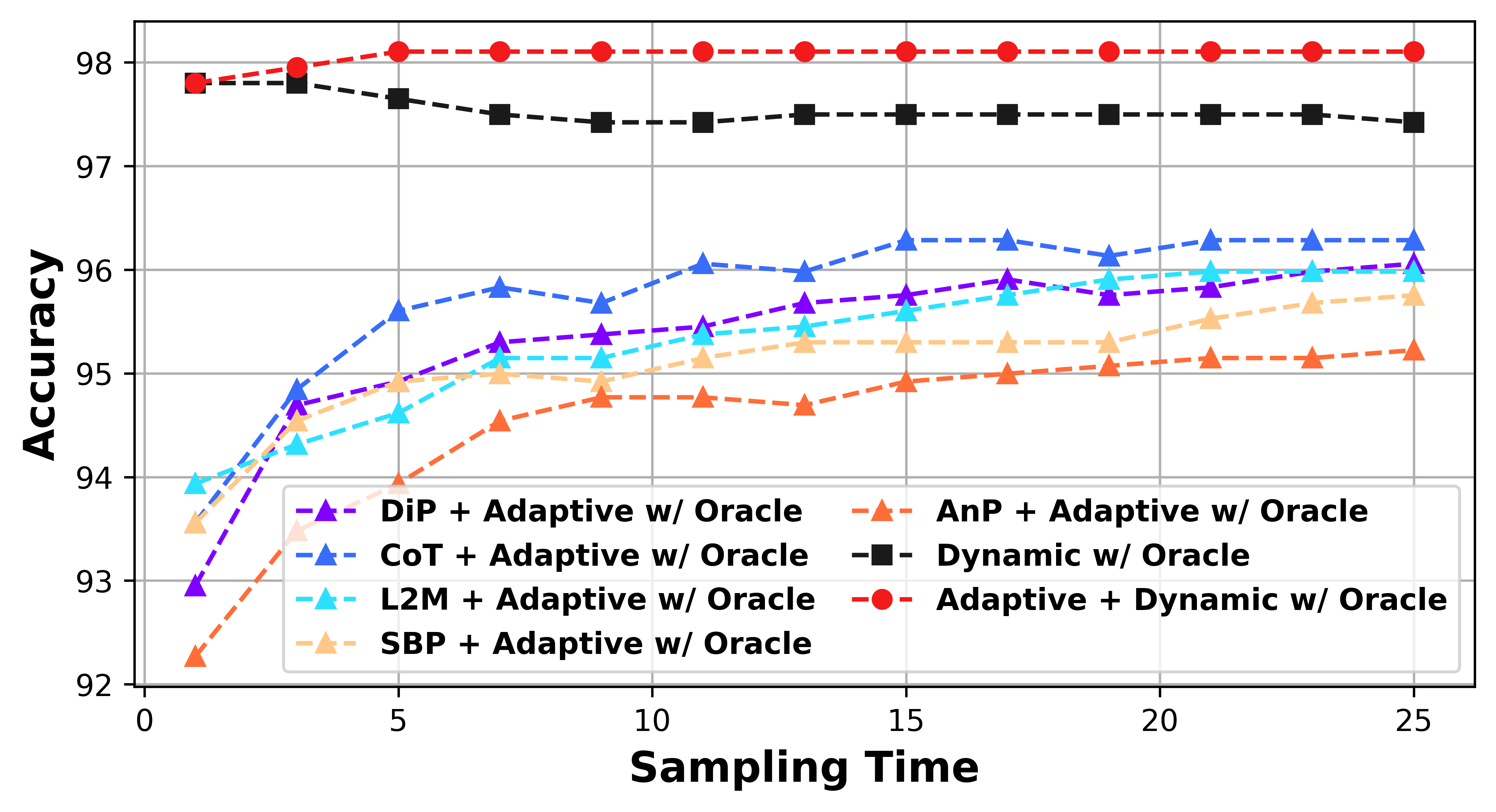}
    \caption{Results of combining adaptively scaling and dynamically choosing the optimal $\prompt{i}$ on GPT-4o-mini on GSM8K.}
    \label{combine gsm8k gpt}
\end{figure}

\begin{figure}[!h]
    \centering
    \includegraphics[width=1\linewidth]{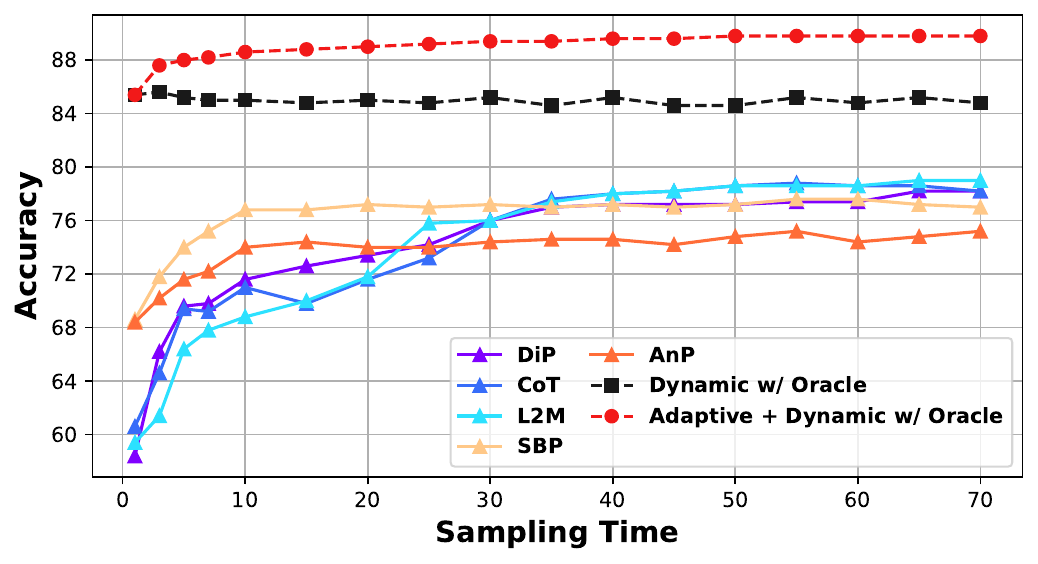}
    \caption{Results of combining adaptively scaling and dynamically choosing the optimal $\prompt{i}$ on Qwen2.5-7B-Instruct on MATH.}
    \label{combine math qwen}
\end{figure}

\begin{figure}[!h]
    \centering
    \includegraphics[width=1\linewidth]{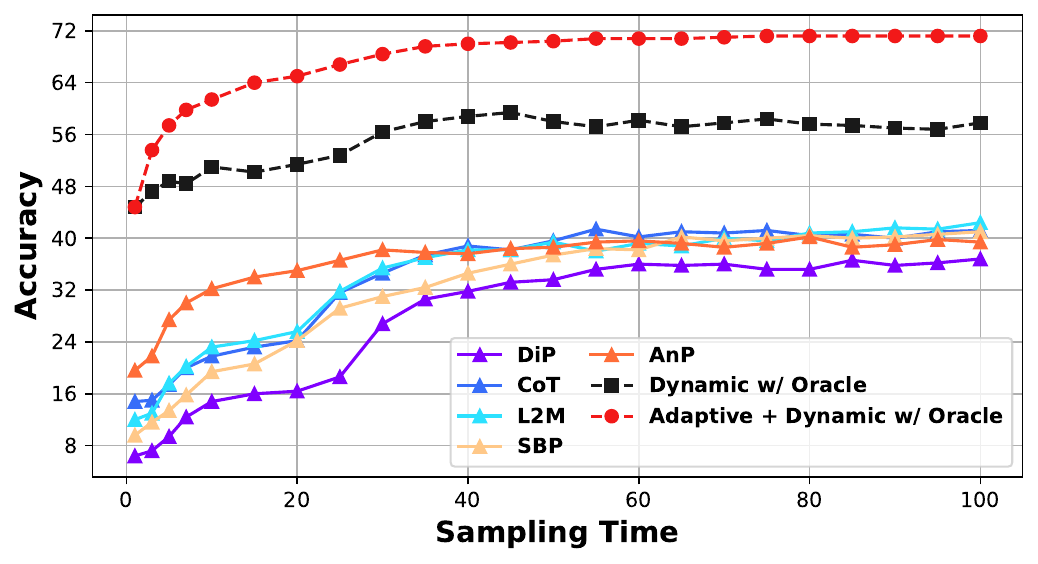}
    \caption{Results of combining adaptively scaling and dynamically choosing the optimal $\prompt{i}$ on LLaMA-3-8B-Instruct on MATH.}
    \label{combine math llama}
\end{figure}

\begin{figure}[!h]
    \centering
    \includegraphics[width=1\linewidth]{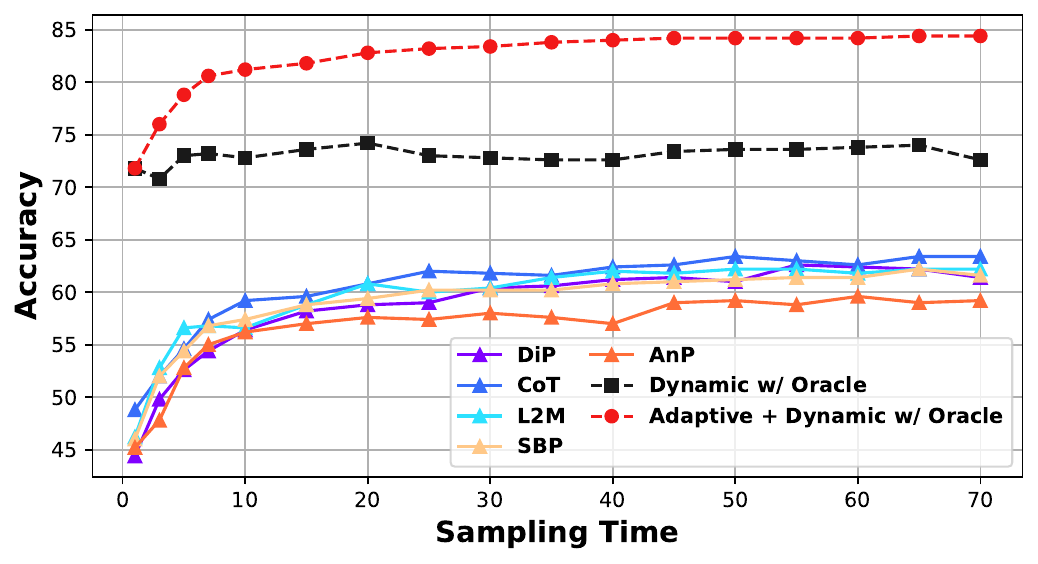}
    \caption{Results of combining adaptively scaling and dynamically choosing the optimal $\prompt{i}$ on GLM-4-9B-Instruct on MATH.}
    \label{combine math glm}
\end{figure}

\begin{figure}[!h]
    \centering
    \includegraphics[width=1\linewidth]{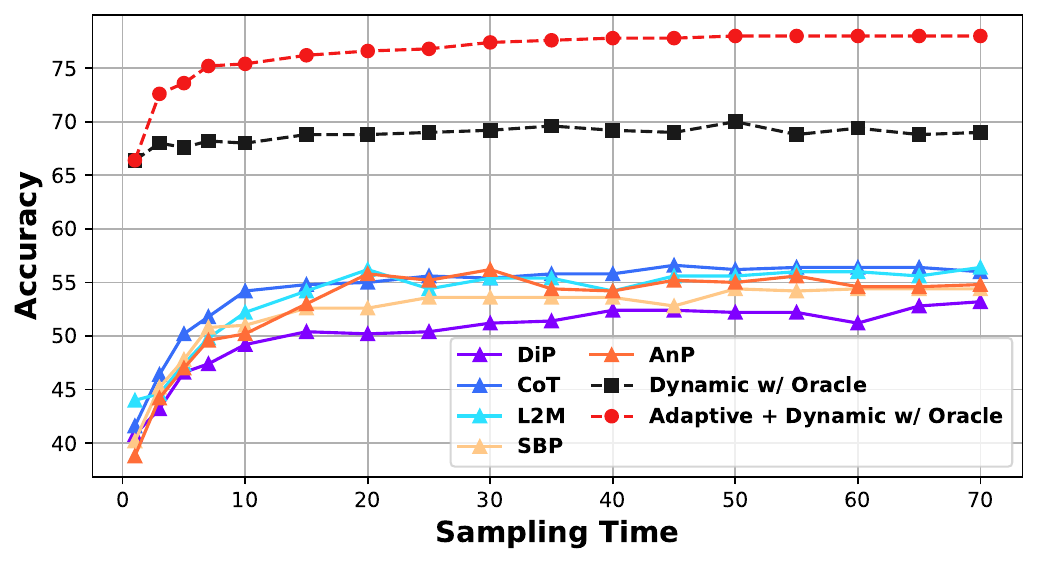}
    \caption{Results of combining adaptively scaling and dynamically choosing the optimal $\prompt{i}$ on Phi-3.5-mini-Instruct on MATH.}
    \label{combine math phi}
\end{figure}

\begin{figure}[!h]
    \centering
    \includegraphics[width=1\linewidth]{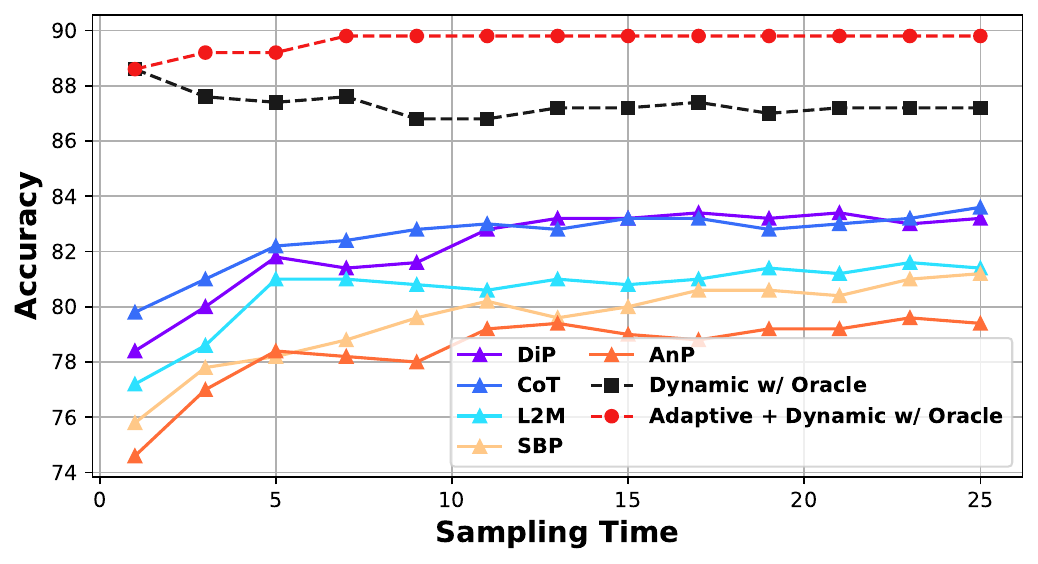}
    \caption{Results of combining adaptively scaling and dynamically choosing the optimal $\prompt{i}$ on Gemini-1.5-Flash on MATH.}
    \label{combine math gemini}
\end{figure}

\begin{figure}[!h]
    \centering
    \includegraphics[width=1\linewidth]{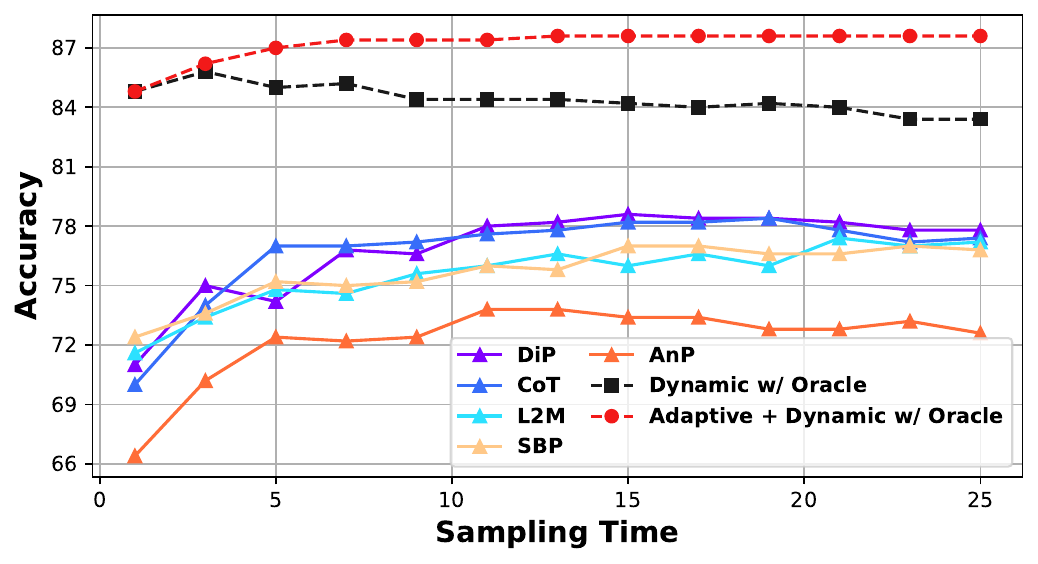}
    \caption{Results of combining adaptively scaling and dynamically choosing the optimal $\prompt{i}$ on GPT-4o-mini on MATH.}
    \label{combine math gpt}
\end{figure}

Figures \ref{combine gsm8k llama} to \ref{combine gsm8k gpt} display the results of combining adaptively scaling and dynamically choosing the optimal $\prompt{i}$ on GSM8K on LLaMA-3-8B-Instruct, GLM-4-9B-Chat, Phi-3.5-mini-Instruct, Gemini-1.5-Flash, and GPT-4o-mini, respectively. Figures \ref{combine math qwen} to \ref{combine math gpt} show the results on each LLM on MATH, respectively. Extensive experiments demonstrate the general effectiveness and superiority of this method, which has an extremely high upper bound.

\clearpage
\section{Extended Experiments}
\label{appendix extended experiments}
\begin{figure*}[!h]
    \centering
    \includegraphics[width=1\linewidth]{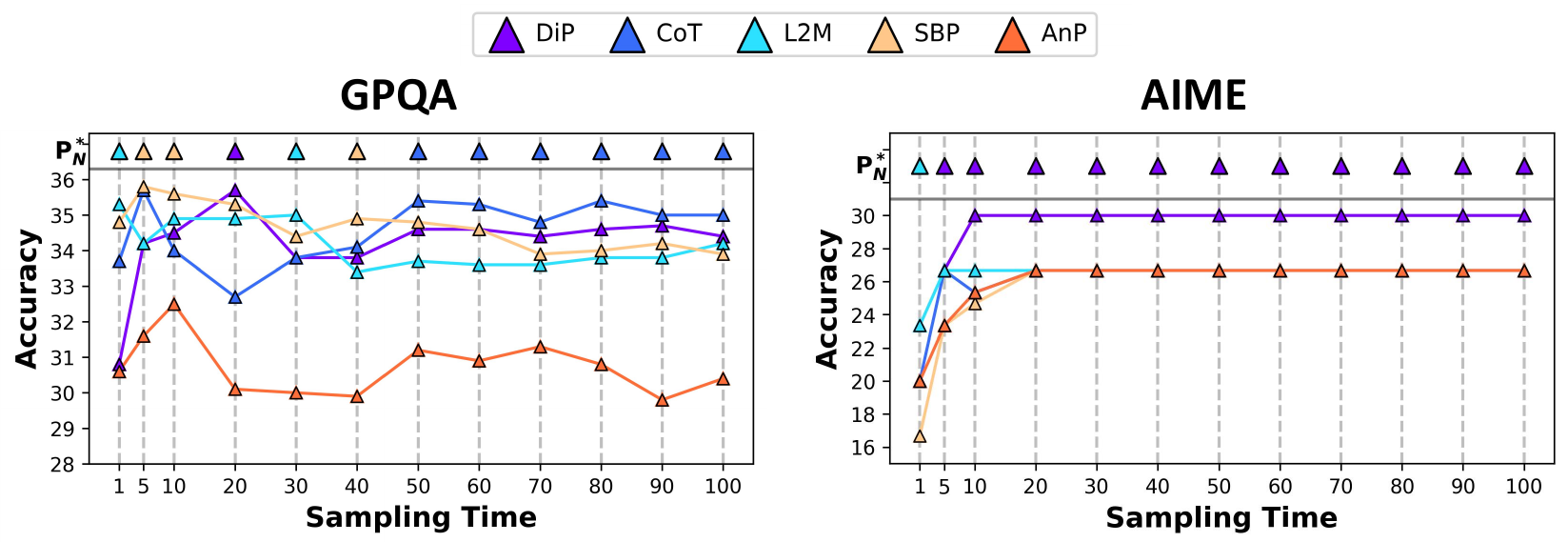}
    \caption{Accuracy of each $\prompt{i}$ with Qwen2.5-7B-Instruct on GQPA and AIME2024.}
    \label{gpqa aime result}
\end{figure*}

\begin{figure}[!h]
    \centering
    \includegraphics[width=1\linewidth]{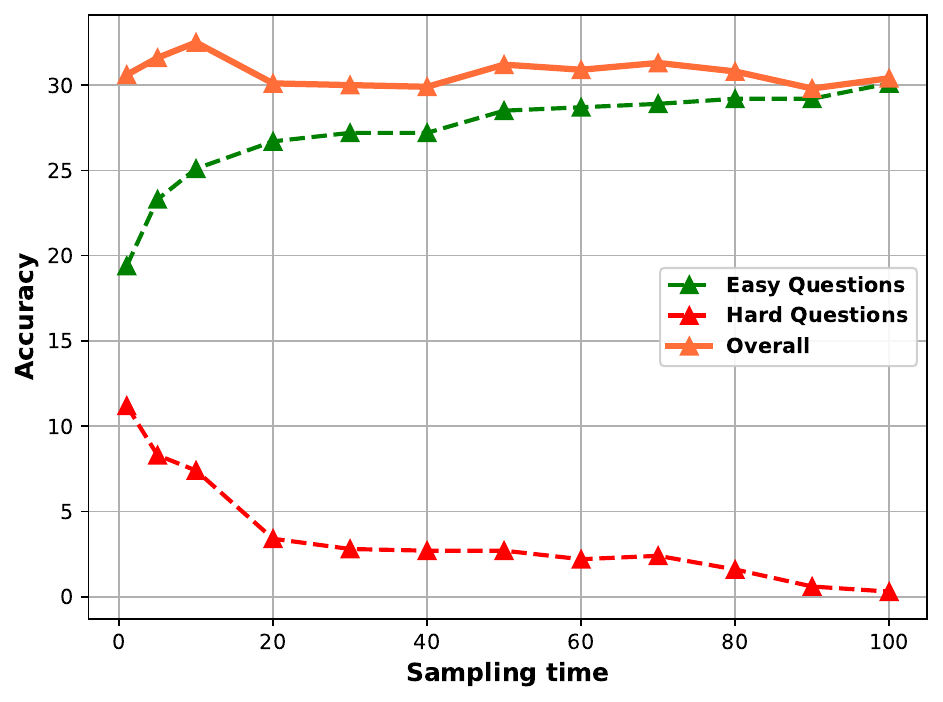}
    \caption{Scaling performance of AnP on the easy/hard question subsets of GPQA. The accuracy on easy questions is non-decreasing with the sampling time, while it exhibits a general declining trend on hard questions. This also holds for all other prompting strategies.}
    \label{Easy vs Hard}
\end{figure}

We extend experiments with Qwen2.5-7B-Instruct on two more challenging reasoning benchmarks, GPQA \citep{GPQA} and AIME2024 \citep{MAA_AIME2024}, further verifying the generality of our findings and methods. 
Figure \ref{gpqa aime result} displays the accuracy of each $\prompt{i}$ with Qwen2.5-7B-Instruct on GQPA and AIME. Simple CoT and DiP gradually dominate as scaling sampling times on GPQA and AIME, respectively. On GPQA, we can see that the performance fluctuates as scaling, rather than constantly increasing. This can be attributed to the ratio of easy/hard questions and their accuracies in the entire dataset. Given AnP as an example, Figure \ref{Easy vs Hard} reports its scaling performance on the easy/hard question subsets of GPQA. Performance on easy questions monotonically improves with increasing sampling times, while accuracy on hard questions exhibits a corresponding decline. This fundamental trade-off induces characteristic oscillation in the aggregate accuracy, with consistent replication across all tested prompting strategies, substantiating our theoretical framework.

\begin{figure*}[!h]
    \centering
    \includegraphics[width=1\linewidth]{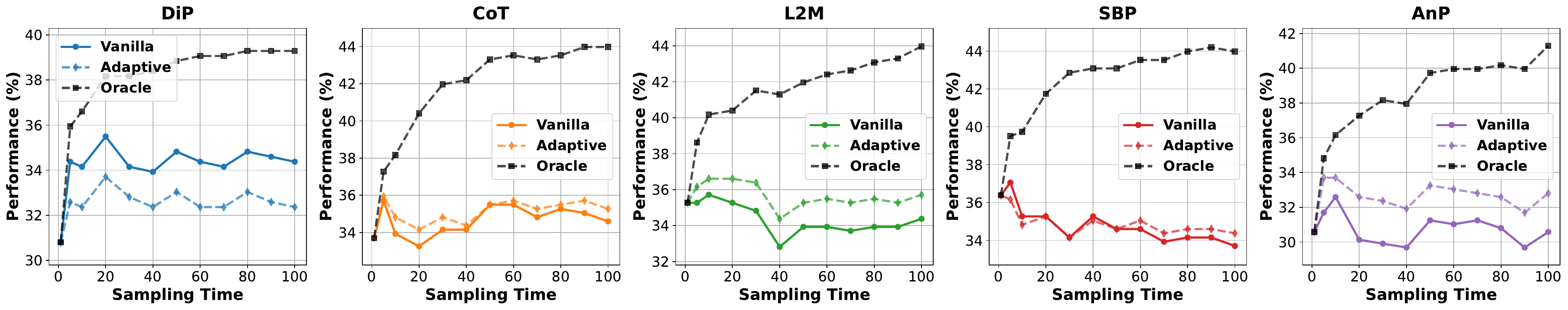}
    \caption{Results of adaptively scaling based on the question difficulty on Qwen2.5-7B-Instruct on GPQA.}
    \label{adaptive qwen gpqa}
\end{figure*}

Figure \ref{adaptive qwen gpqa} shows the results of adaptively scaling on each $\prompt{i}$. ``Adaptive'' approach demonstrates evident performance gains over ``Vanilla'' baselines across CoT, L2M, SBP, and AnP, indicating the model's intrinsic capability to assess question difficulty. Under ``Oracle'' conditions, it can achieve further performance amplification. 

\begin{figure}[!h]
    \centering
    \includegraphics[width=1\linewidth]{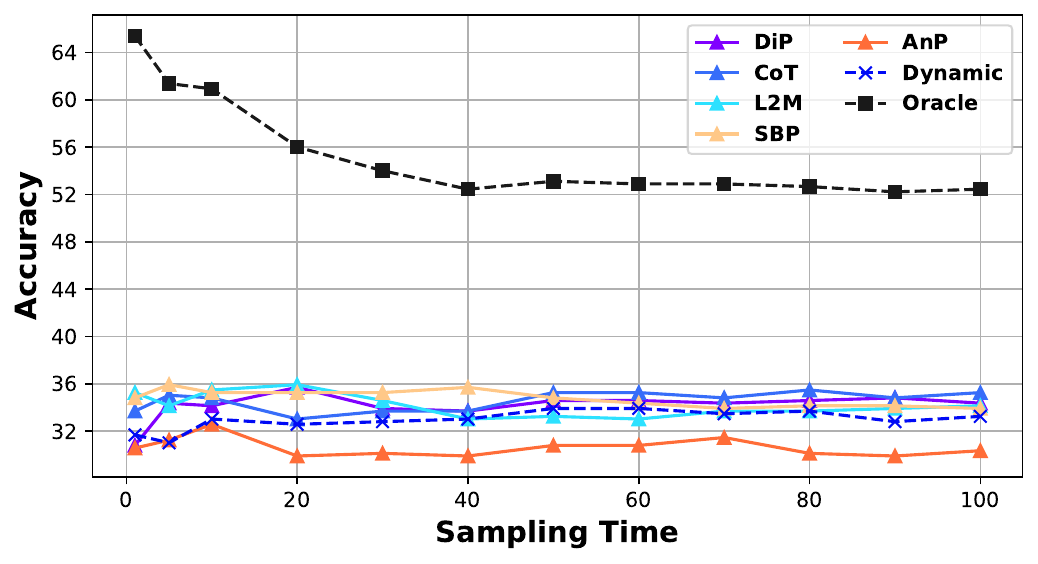}
    \caption{Results of dynamically choosing the optimal $\prompt{i}$ on Qwen2.5-7B-Instruct on GPQA.}
    \label{dynamic qwen gpqa}
\end{figure}

Figure \ref{dynamic qwen gpqa} reports the results of dynamically choosing the optimal $\prompt{i}$ on Qwen2.5-7B-Instruct on GPQA. ``Dynamic'' approach achieves median-range performance across all tested $\prompt{i}$, quantitatively confirming the model's suboptimal strategy selection capacity. Strikingly, ``Oracle'' intervention enables dramatic performance elevation, with 65.4\% accuracy at $\numbersample=1$.

\begin{figure}[!h]
    \centering
    \includegraphics[width=1\linewidth]{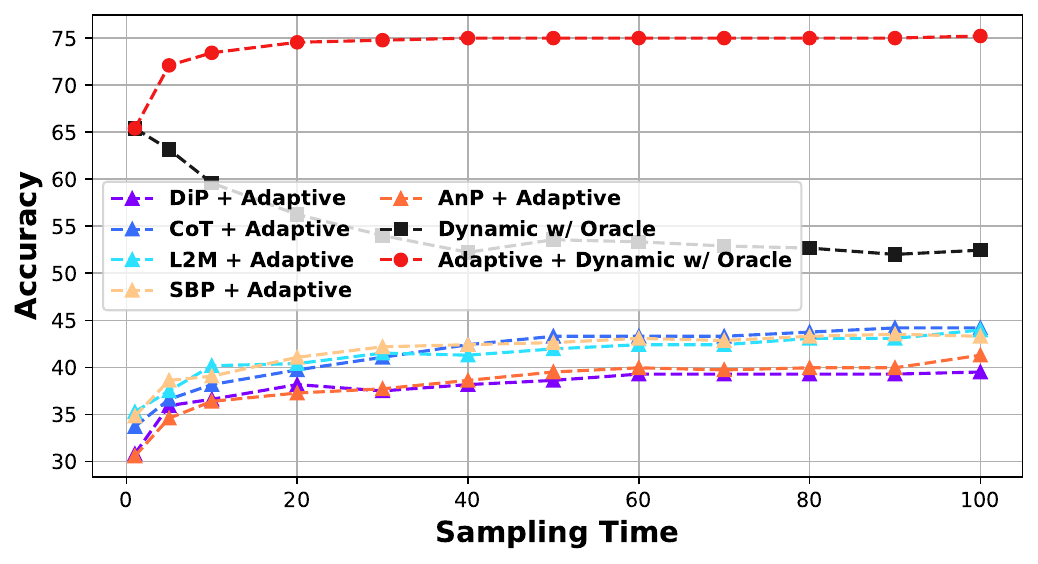}
    \caption{Results of combining adaptively scaling and dynamically choosing the optimal $\prompt{i}$ on Qwen2.5-7B-Instrcut on GPQA.}
    \label{combine qwen gpqa}
\end{figure}

Figure \ref{combine qwen gpqa} summarizes the results of combining adaptively scaling and dynamically choosing the optimal $\prompt{i}$ on Qwen2.5-7B-Instrcut on GPQA, which further enormously enhances the scaling performance, with 75.7\% accuracy at $\numbersample=100$.

\newpage
\section{Implementation Details and Prompts}
\label{implementation details and prompts}
We use vllm \citep{kwon2023efficient} to deploy open-sourced LLMs, with top-p = 0.9 and temperature = 0.7. For closed-sourced LLMs, we use their APIs with default settings. We set the content safety detection threshold of Gemini-1.5-Flash to zero to prevent erroneous judgments that may result in null outputs. 

\begin{figure*}[h]
    \centering
    \includegraphics[width=0.9\linewidth]{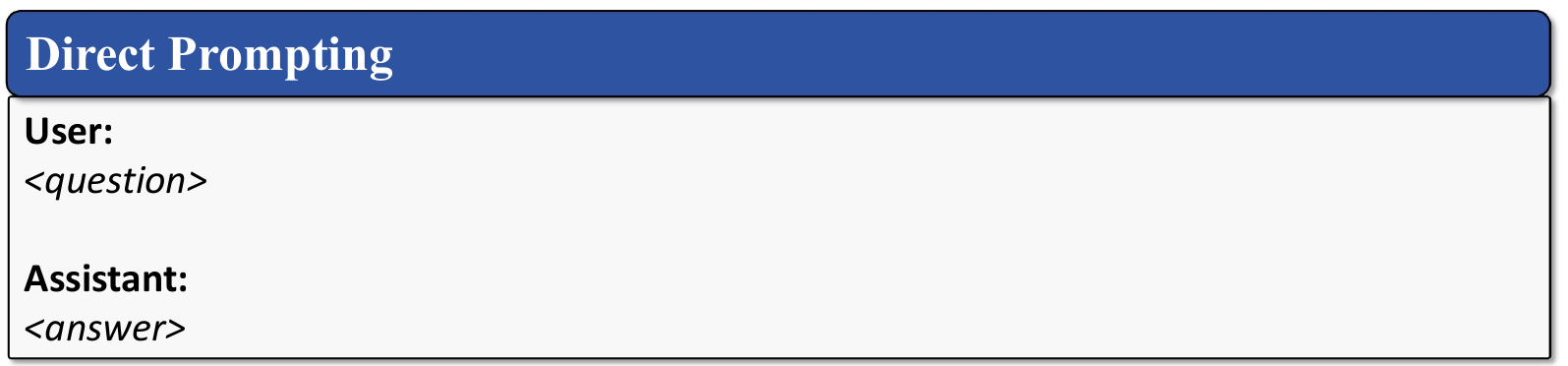}
    \caption{Prompt of DiP.}
    \label{DiP prompt}
\end{figure*}

\begin{figure*}
    \centering
    \includegraphics[width=0.9\linewidth]{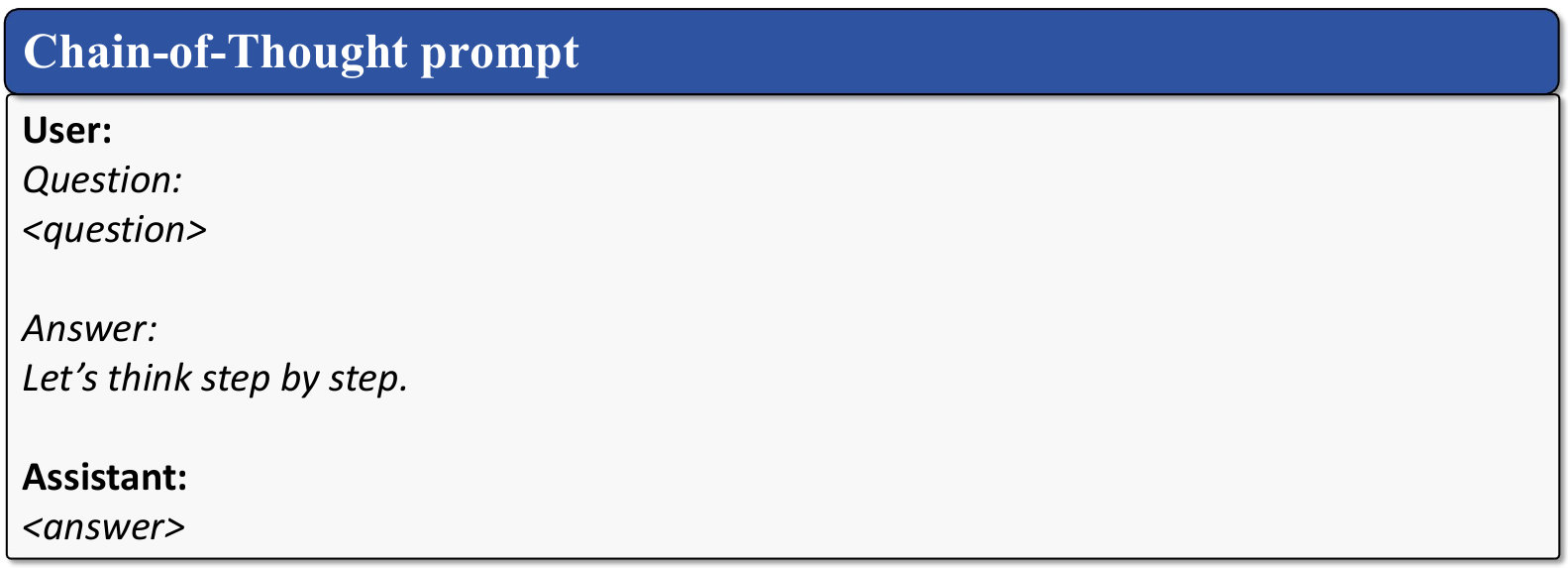}
    \caption{Prompt of CoT.}
    \label{CoT prompt}
\end{figure*}

\begin{figure*}
    \centering
    \includegraphics[width=0.9\linewidth]{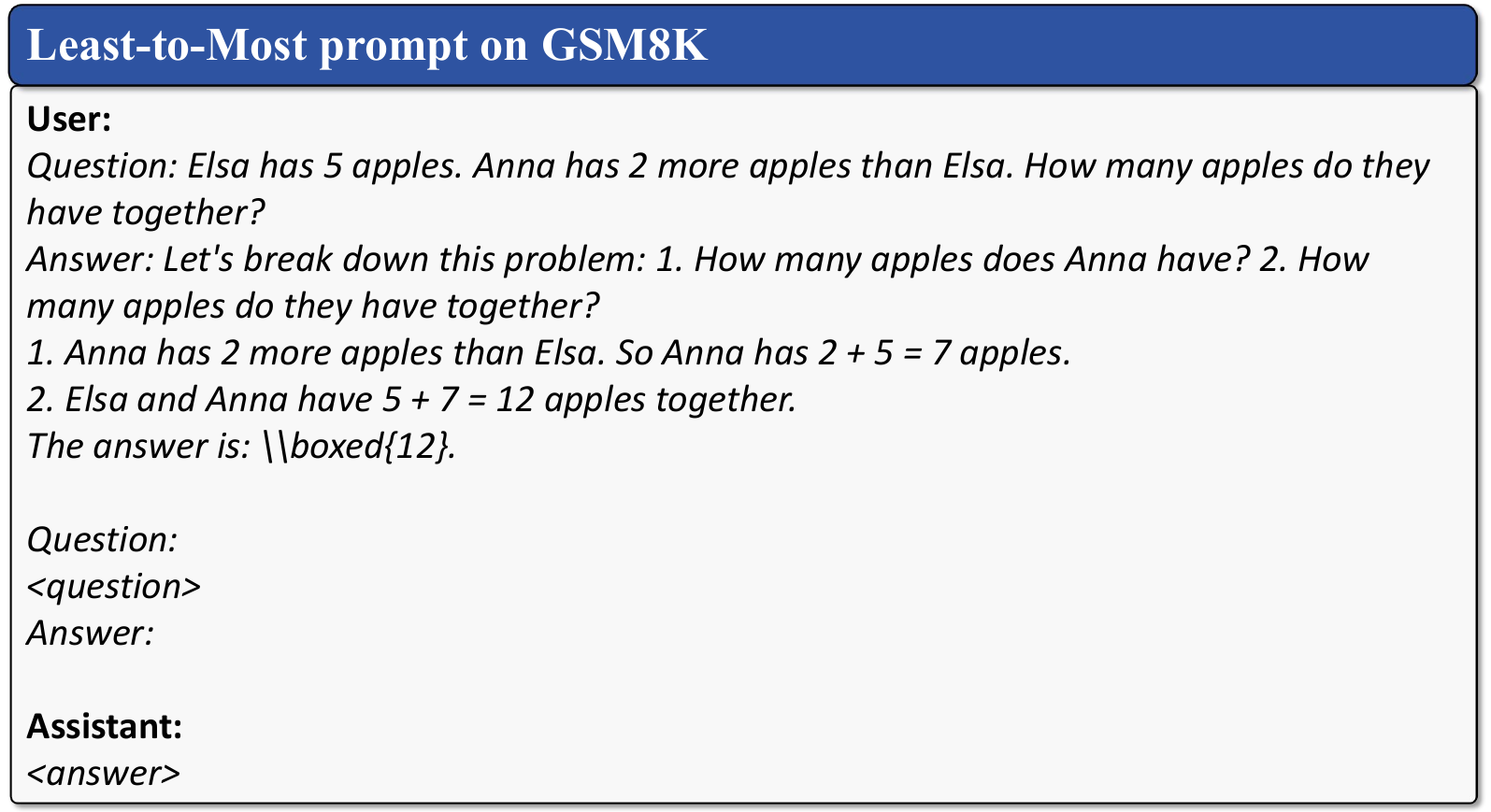}
    \caption{Prompt of L2M on GSM8K.}
    \label{L2M prompt GSM8K}
\end{figure*}

\begin{figure*}
    \centering
    \includegraphics[width=0.9\linewidth]{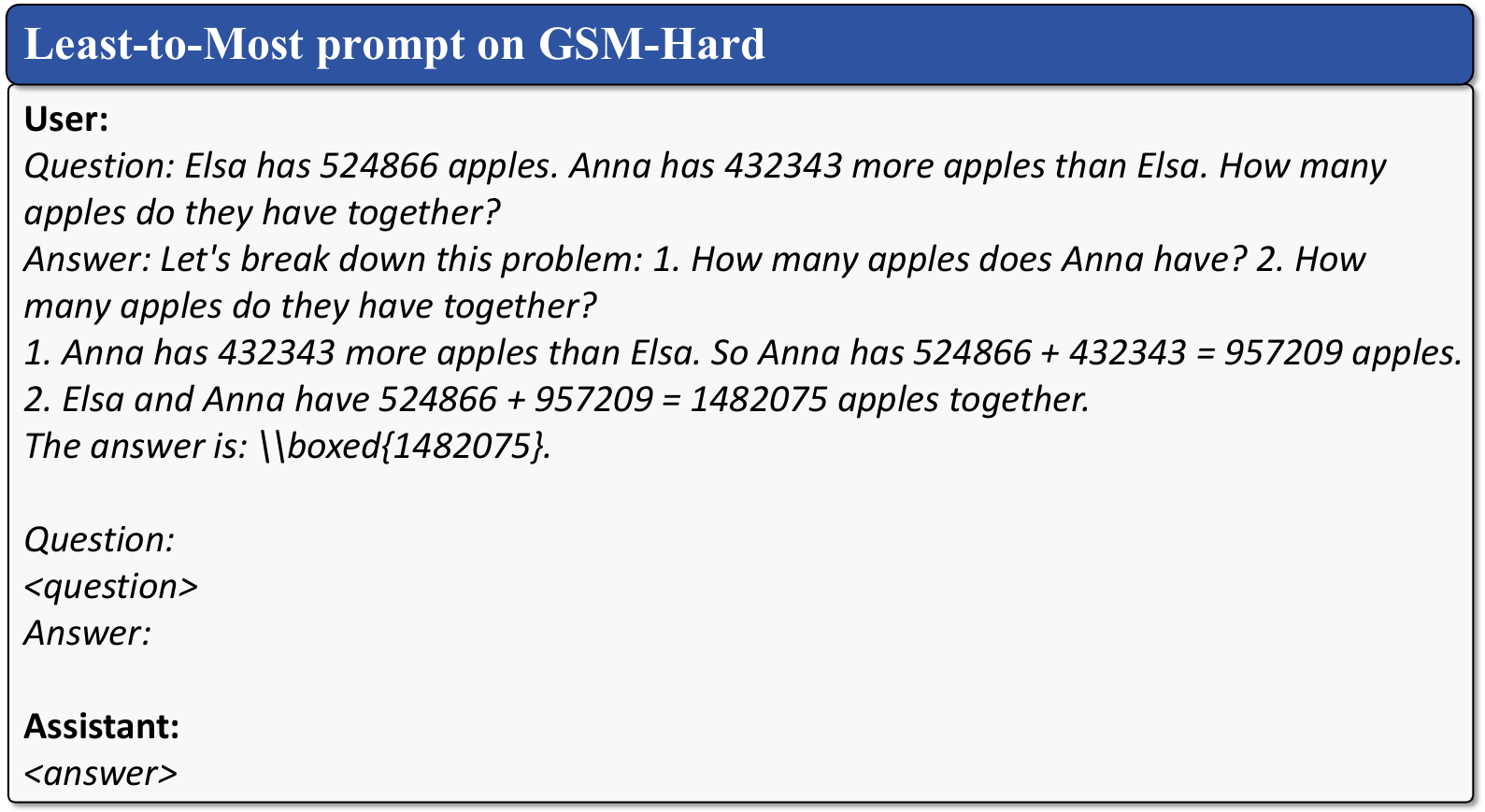}
    \caption{Prompt of L2M on GSM-Hard.}
    \label{L2M prompt GSM-Hard}
\end{figure*}

\begin{figure*}
    \centering
    \includegraphics[width=0.9\linewidth]{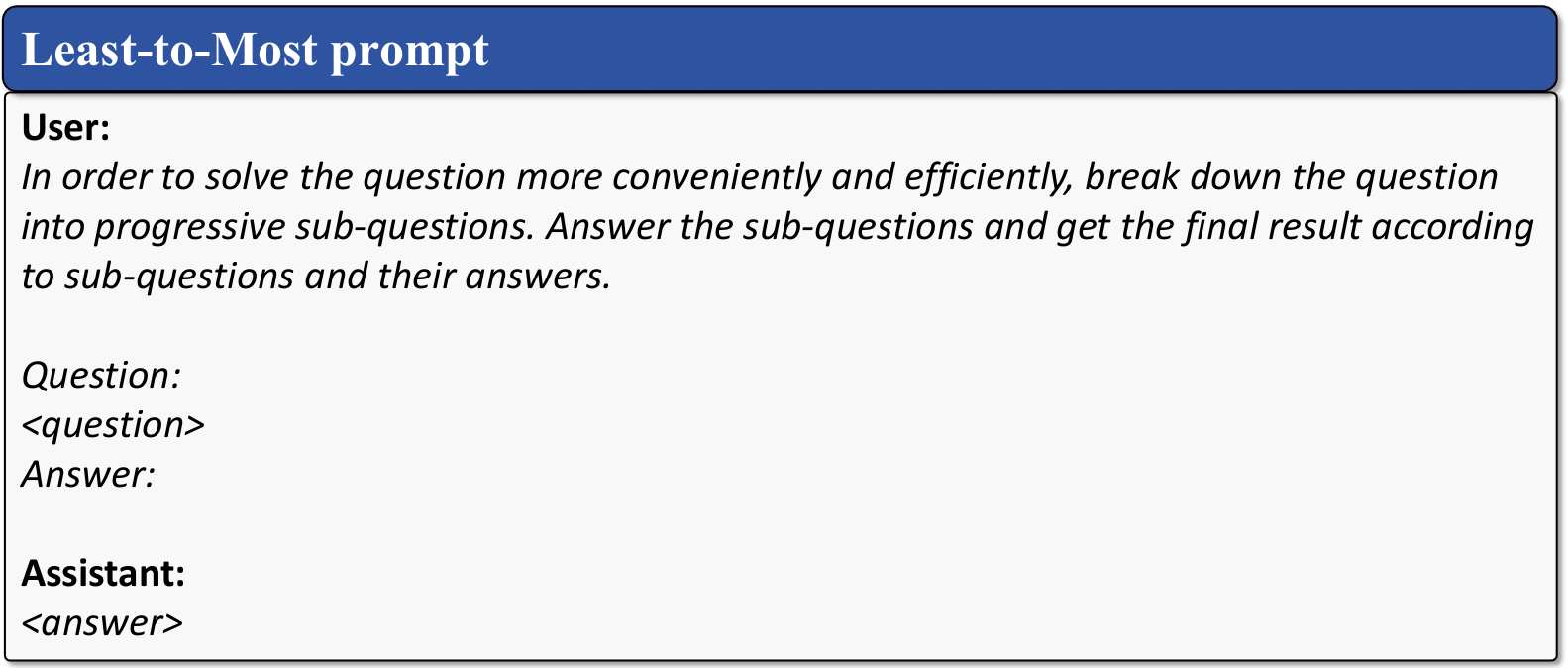}
    \caption{Prompt of L2M on MATH and MMLU.}
    \label{L2M prompt}
\end{figure*}

\begin{figure*}
    \centering
    \includegraphics[width=0.9\linewidth]{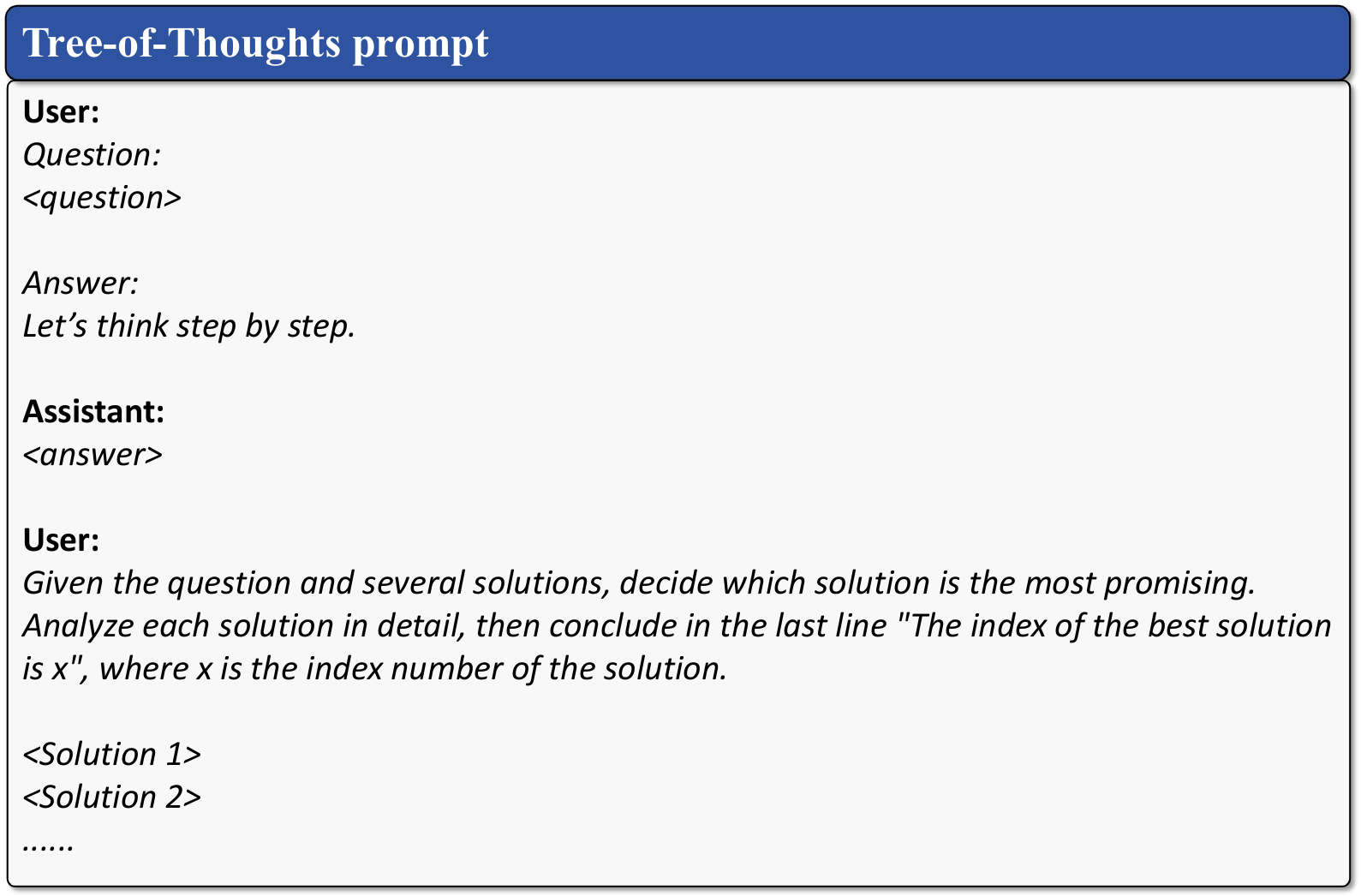}
    \caption{Prompt of ToT.}
    \label{ToT prompt}
\end{figure*}

\begin{figure*}
    \centering
    \includegraphics[width=0.9\linewidth]{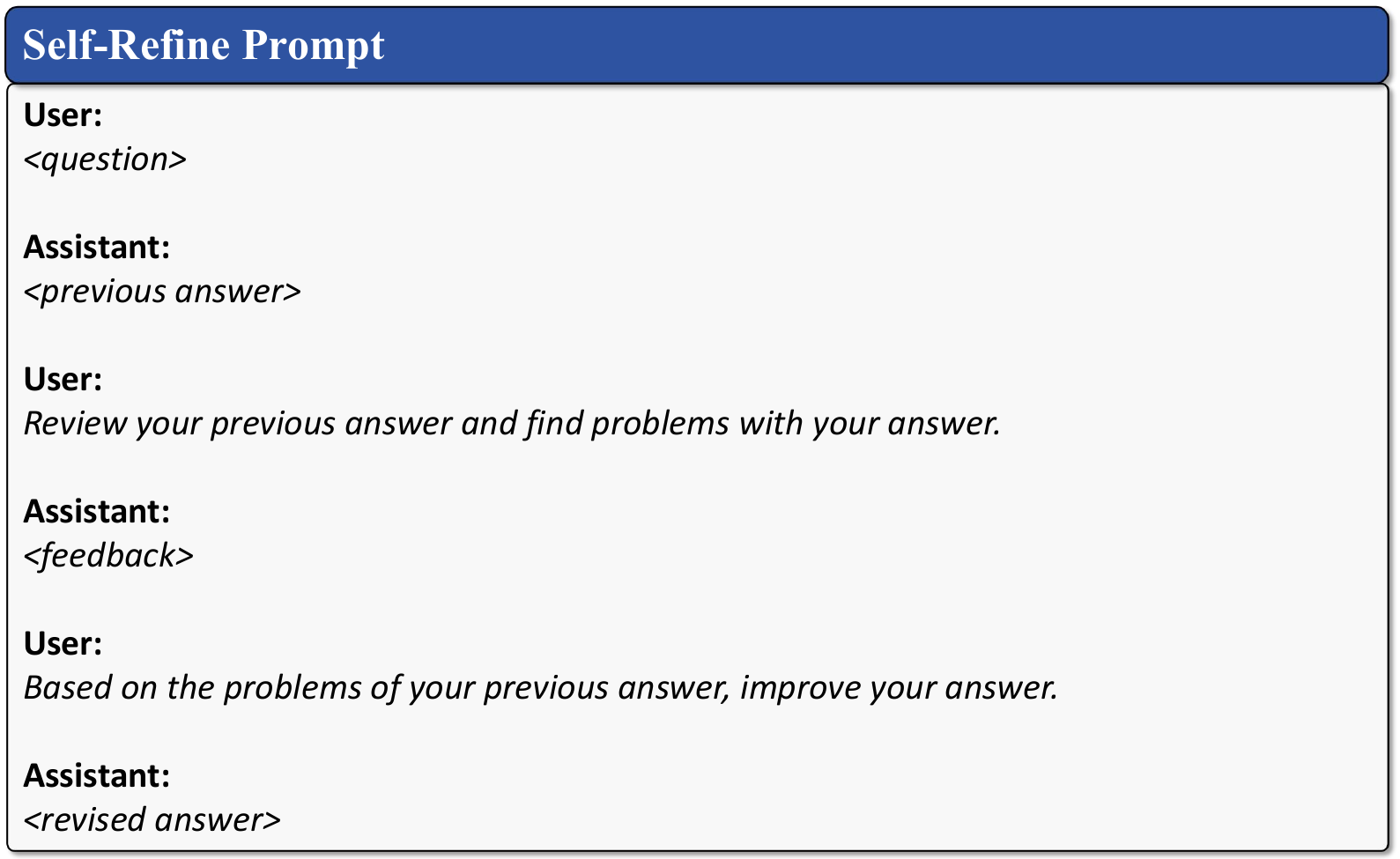}
    \caption{Prompt of S-RF.}
    \label{S-RF prompt}
\end{figure*}

\begin{figure*}
    \centering
    \includegraphics[width=0.9\linewidth]{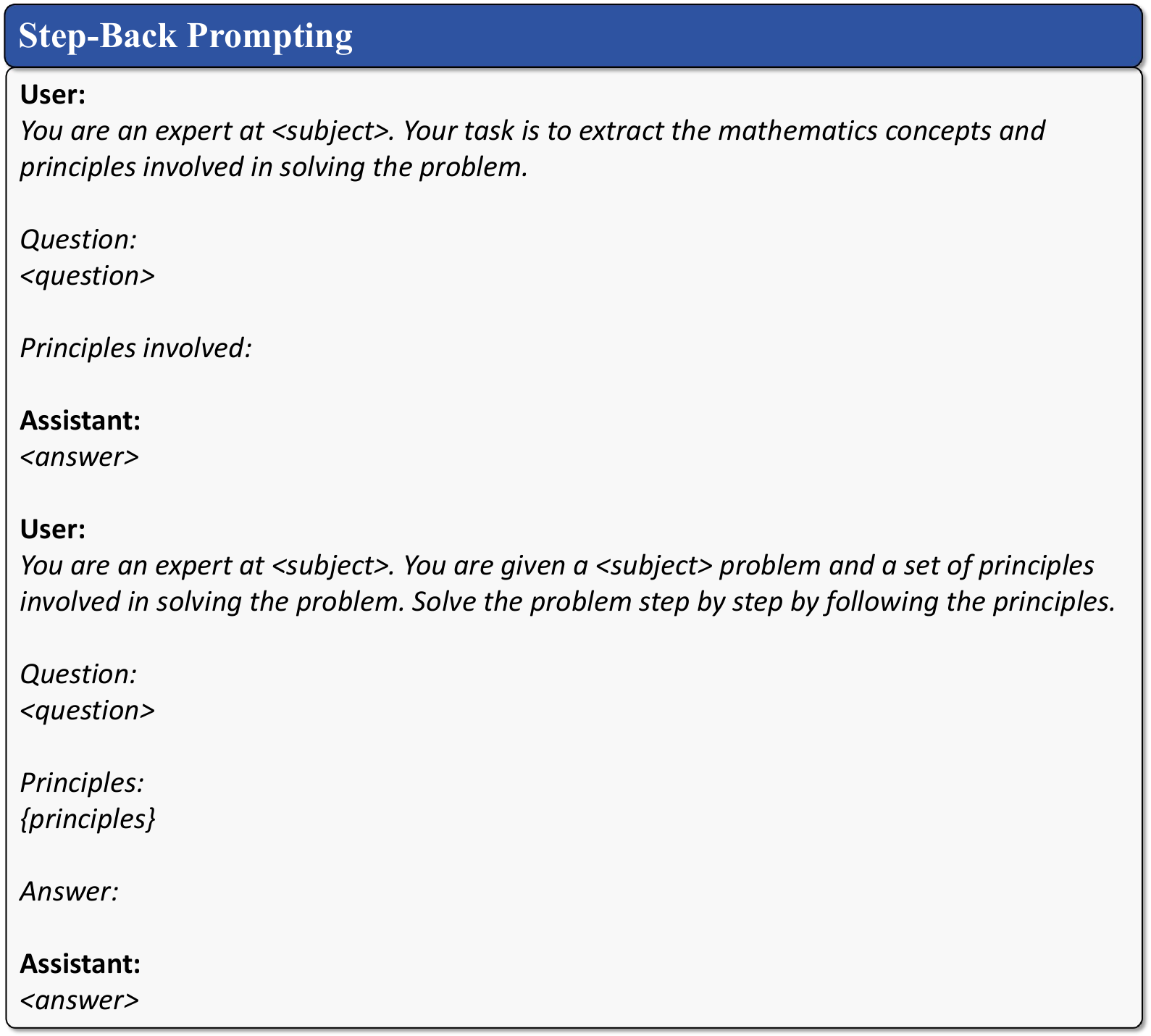}
    \caption{Prompt of SBP.}
    \label{SBP prompt}
\end{figure*}

\begin{figure*}
    \centering
    \includegraphics[width=0.9\linewidth]{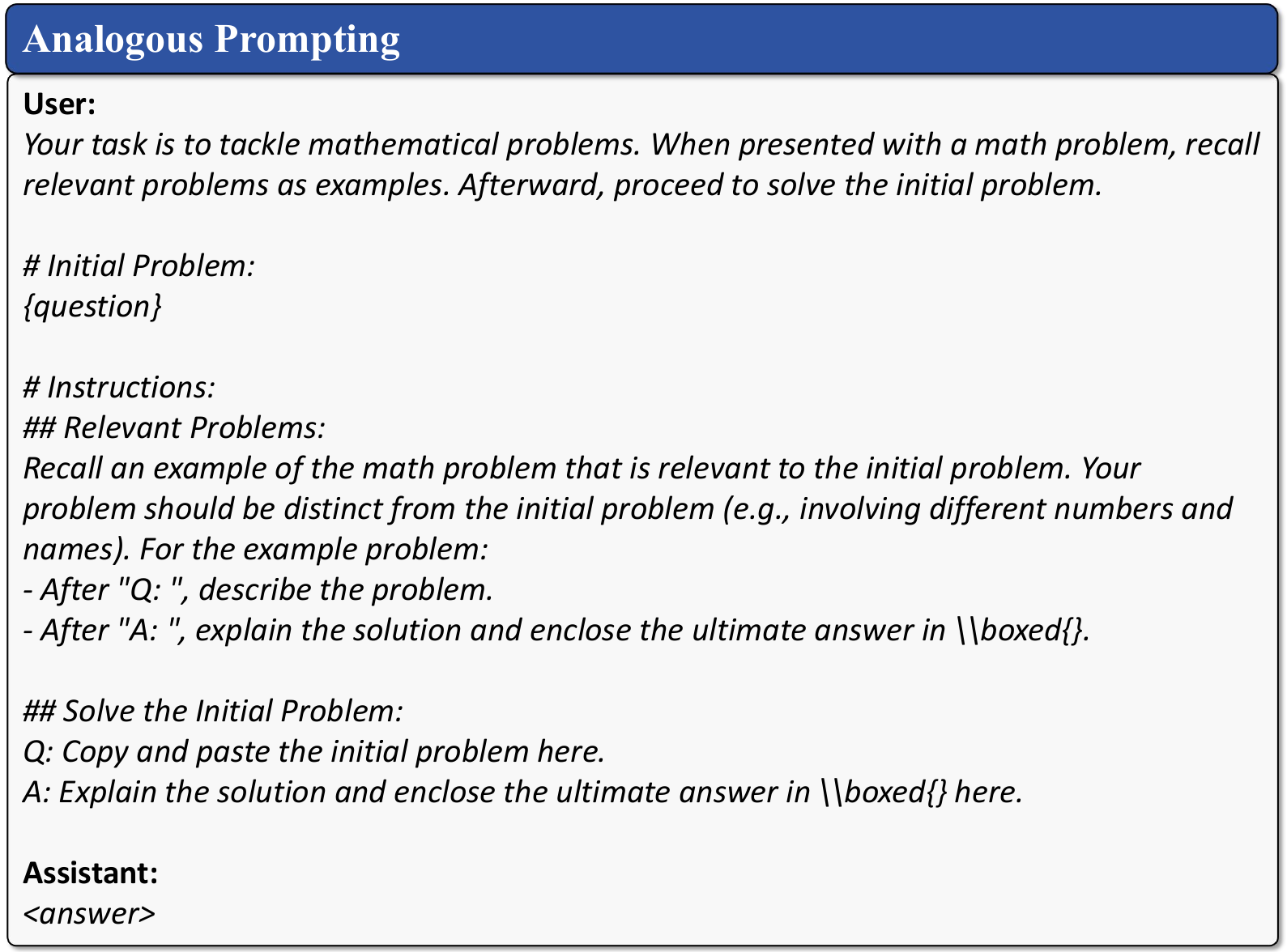}
    \caption{Prompt of AnP.}
    \label{AnP prompt}
\end{figure*}

\begin{figure*}
    \centering
    \includegraphics[width=0.9\linewidth]{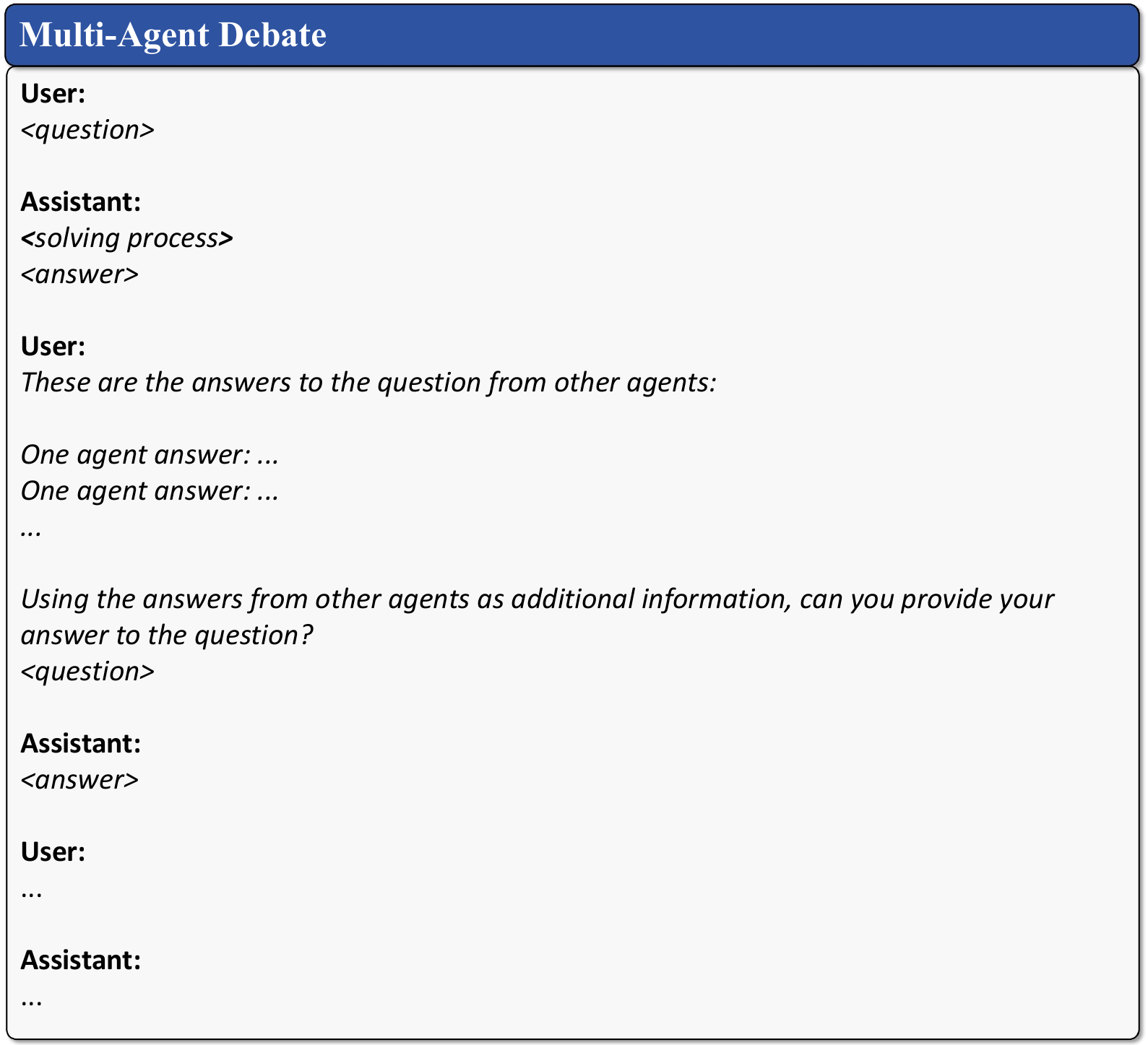}
    \caption{Prompt of MAD.}
    \label{MAD prompt}
\end{figure*}

Following \citep{wang2024math, lightmanlet, qi2024mutual}, we use MATH-500, a subset of representative problems from the MATH dataset to speed up the evaluation. We use the test split of each dataset. The license for all datasets is CC-BY 4.0 or others for open academic research. The number of samples on each dataset is shown in Table \ref{sample number dataset}. We ensure our use of existing artifacts is aligned with their intended purposes. All of them are public English datasets for academic research. On GSM8K and GSM-Hard, we use the same 1-shot prompt in the original paper of Least-to-Most \citep{least-to-most} shown in Figure \ref{L2M prompt GSM8K} and Figure \ref{L2M prompt GSM-Hard}. On other datasets, we use the 0-shot prompt shown in Figure \ref{L2M prompt}. We use the same prompt in Analogous Prompting \citep{AnP}, and guide the LLM to recall one analogous problem. We use the same 1-shot prompt in Step-Back Prompting \citep{SBP} on MMLU, and apply their prompt designed for reasoning tasks on other datasets. We use the same prompt in 0-shot Chain-of-Thought \cite{cot-0shot}, Multi-Agent Debate \cite{MAD} and Self-Refine \cite{huanglarge} on all datasets. The prompts are shown in Figures \ref{DiP prompt} to \ref{MAD prompt}.

\begin{table}[h]
    \centering
     \caption{The number of samples in each dataset.}
    \begin{tabular}{c|c}
    \toprule
         Dataset & Samples\\  \midrule
         GSM8K & 1318\\
         GSM-Hard &1318\\
         MATH-500  &500\\
         MMLU-Biology & 310\\
         MMLU-Chemistry &203\\
         MMLU-Physics  &151 \\
    \bottomrule
    \end{tabular}
    \label{sample number dataset}
\end{table}

\end{document}